\documentclass[fleqn,11pt]{wlscirep}
\usepackage[utf8]{inputenc}
\usepackage[T1]{fontenc}
\usepackage{lineno}

\usepackage{times}

  \DeclareFontEncoding{FML}{}{}
  \DeclareFontSubstitution{FML}{fncmi}{m}{it}
  \DeclareSymbolFont{fourierletters}{FML}{fncmi}{m}{it}
  \DeclareMathSymbol{\freq}{\mathalpha}{fourierletters}{`f}

\DeclareMathOperator{\arccosh}{arccosh}
\DeclareMathOperator{\arcsinh}{arcsinh}
\DeclareMathOperator{\arctanh}{arctanh} 
\usepackage[symbol]{footmisc}

\usepackage{booktabs} %
\usepackage{graphicx}

\usepackage{amsthm}
\usepackage{multirow}
\usepackage{multicol}
\usepackage{pgfplots}
\usepackage{mathtools}

\usepackage{multicol}
\usepackage{xcolor}
\usepackage{caption}
\usepackage{subcaption}

\newcommand{\new}[1]{\textcolor{black}{#1}}

\newcommand\keymaera{\textrm{KeYmaera} } %
\newcommand\keymaeraX{\textrm{KeYmaeraX} } %

 \usepackage{array,multirow,graphicx}
 \usepackage{float}
\usepackage{listings}
\usepackage{url}
\usepackage{amsmath}   %
\usepackage{amsfonts}   %

\definecolor{airforceblue}{rgb}{0.25, 0.42, 0.70}
\usepackage{longtable}

\makeatletter
\newcommand{\dashrule}[1][black]{%
  \color{#1}\rule[\dimexpr.5ex-.2pt]{4pt}{.4pt}\xleaders\hbox{\rule{4pt}{0pt}\rule[\dimexpr.5ex-.2pt]{4pt}{.4pt}}\hfill\kern0pt%
}
\makeatother

\usepackage{algorithm}
\usepackage[noend]{algpseudocode}
\makeatletter
\usepackage{caption}
\usepackage{xcolor}
\usepackage{float}
\definecolor{dkblue}{rgb}{0,0,.6}

\usepackage[obeyFinal, textsize=tiny]{todonotes} %

\renewcommand\fs@ruled{\def\@fs@cfont{\bfseries}\let\@fs@capt\floatc@ruled
  \def\@fs@pre{{\color{black}\hrule height.8pt depth0pt \kern2pt}}%
  \def\@fs@post{{\color{black}\kern2pt\hrule\relax}}%
  \def\@fs@mid{{\kern2pt\color{black}\hrule\kern2pt}}%
  \let\@fs@iftopcapt\iftrue}
\makeatother

\title{\bf AI Descartes: Combining Data and Theory for Derivable Scientific Discovery}

\author[1,2]{Cristina Cornelio}
\author[1]{Sanjeeb Dash}
\author[1]{Vernon Austel}
\author[3,4]{Tyler R. Josephson}
\author[1]{Joao Goncalves}
\author[1]{Kenneth Clarkson} 
\author[1]{Nimrod Megiddo}
\author[1]{Bachir El Khadir}
\author[1,*]{Lior Horesh}
\affil[1]{IBM Research}
\affil[2]{Samsung AI - Cambridge}
\affil[3]{Department of Chemical, Biochemical, and Environmental Engineering, University of Maryland, Baltimore County}
\affil[4]{Department of Chemistry and Chemical Theory Center, University of Minnesota}

\affil[*]{corresponding author: lhoresh@us.ibm.com}

\date{}

\begin{abstract}
Scientists have long aimed to discover meaningful formulae which accurately describe experimental data.
A common approach is to manually create mathematical models of natural phenomena using domain knowledge, and then fit these models to data.
In contrast, machine-learning algorithms automate the construction of accurate data-driven models while consuming large amounts of data. 
\new{The problem of incorporating prior knowledge in the form of constraints on the functional form of a learned model (e.g., nonnegativity) has been explored in the literature~\cite{scott2021lgml,ashok2020logic,kubalik2020symbolic}. However, finding models that are consistent with prior knowledge expressed in the form of general logical axioms (e.g., conservation of energy) is an open problem.}
\new{We develop a method to enable principled derivations of models of natural phenomena from axiomatic knowledge and experimental data by combining logical reasoning with symbolic regression.}
We demonstrate these concepts for Kepler’s third law of planetary motion, Einstein’s relativistic time-dilation law, and Langmuir’s theory of adsorption, automatically connecting experimental data with background theory in each case.
We show that laws can be discovered from few data points when using formal logical reasoning to distinguish the correct formula from a set of plausible formulas that have similar error on the data.
The combination of reasoning with machine learning provides generalizeable insights into key aspects of natural phenomena.
We envision that this combination will enable derivable discovery of fundamental laws of science
and believe that our work is \new{an important step} towards automating the scientific method.

\end{abstract}

\begin{document}

\flushbottom
\maketitle

\thispagestyle{empty}

\section*{Introduction}

Artificial neural networks (NN) and statistical regression are commonly used to automate the discovery of patterns and relations in data. 
NNs return ``black-box" models, where the underlying functions 
are typically used for prediction only. 
In  %
standard regression, the functional form is determined in advance, so model discovery amounts to parameter fitting. %
In symbolic regression (SR) \cite{koza,koza2}, the functional form is {\em not} determined in advance, but is instead composed from operators in a given list, 
e.g., $+$, $-$, $\times$, and $\div$, %
and %
calculated from the data. %
SR models are typically more ``interpretable'' than NN models, and require less data.
Thus, for discovering laws of nature in symbolic form from experimental data, SR may work better than NNs or fixed-form regression   \cite{schmidt2009distil}; integration of NNs  with SR has been a topic of recent research in neuro-symbolic AI \cite{Extrapolation_learning_equations,NN_discovery, AI-Feynman}.
A major challenge in SR is to identify,
out of many models that fit the data,
those that are {\em scientifically meaningful}. Schmidt and Lipson \cite{schmidt2009distil} %
identify meaningful functions as those that balance accuracy and complexity. However many such expressions exist for a given dataset, and not all are consistent with the known background theory.

Another approach would be to start from the known background theory, but there are no existing practical reasoning tools that generate theorems consistent with experimental data from a set of known axioms. %
Automated Theorem Provers (ATPs), the most widely-used reasoning tools, instead solve the task of proving a conjecture for a given logical theory. Computational complexity is a major challenge for ATPs; for certain types of logic, proving a conjecture is undecidable. %
Moreover, deriving models from a logical theory using formal reasoning tools is especially difficult when arithmetic and calculus operators are involved (e.g., see \cite{Complexity_Semialgebraic_Proofs} for the case of inequalities).
Machine-learning techniques have been used to improve the performance of ATPs, e.g., by using reinforcement learning to guide the search process \cite{TRAIL}.
This research area has received much attention recently \cite{Structured_Semidefinite_Programs,Sum-of-Squares,Learning_dynamic_polynomial_proofs}.

Models that are derivable, and not merely empirically accurate, are appealing because they are arguably correct, predictive, and insightful. %
We attempt to obtain such models by combining a novel mathematical-optimization-based SR method with a reasoning system.
This yields an end-to-end discovery system, which extracts formulas from data via SR, and then furnishes either a formal proof of derivability of the formula from a set of axioms, or a proof of inconsistency.
We present novel measures that indicate
how close a formula is to a derivable formula,
when the model is provably non-derivable, and we calculate the values of these measures using our reasoning system.
\new{In earlier work combining machine learning with reasoning,}
Marra et al.~\cite{visual_learning_logic} use a logic-based description to constrain the output of a GAN neural architecture for generating images. 
Scott et al.~\cite{scott2021lgml} and Ashok et al.~\cite{ashok2020logic} combine machine-learning tools and reasoning engines 
to search for functional forms that satisfy prespecified constraints. They augment the initial dataset with new points in order to improve the efficiency of learning methods and the accuracy of the final model. %
\new{Kubalik et. al. \cite{kubalik2020symbolic} also exploit prior knowledge to create additional data points.} However, these papers only consider constraints on the functional form to be learned, and do not incorporate general background-theory axioms (logic constraints that describe the other laws and unmeasured variables that are involved in the phenomenon).

\section*{Methodology}
Our automated scientific discovery method aims to discover an unknown \emph{symbolic model}
$y = f^*(\textbf{x})$ where
$\textbf{x}$ is the vector\footnote{Bold letters indicate vectors.} $(x_1, \ldots, x_n)$ of independent variables, and $y$ is the dependent variable.
The discovered model $f$ (an approximation of $f^*$) should fit a collection of $m$ data points,  
($(\mathbf{X}^1,Y^1), \cdots , (\mathbf{X}^m,Y^m)$), be derivable from a background theory, have low complexity and bounded prediction error. 
More specifically, the inputs to our system are 4-tuples
$\langle \mathcal B, \mathcal C,  \mathcal D, \mathcal M\rangle$ as follows.

\begin{figure}[t]
\centering
\includegraphics[width=0.8\textwidth]{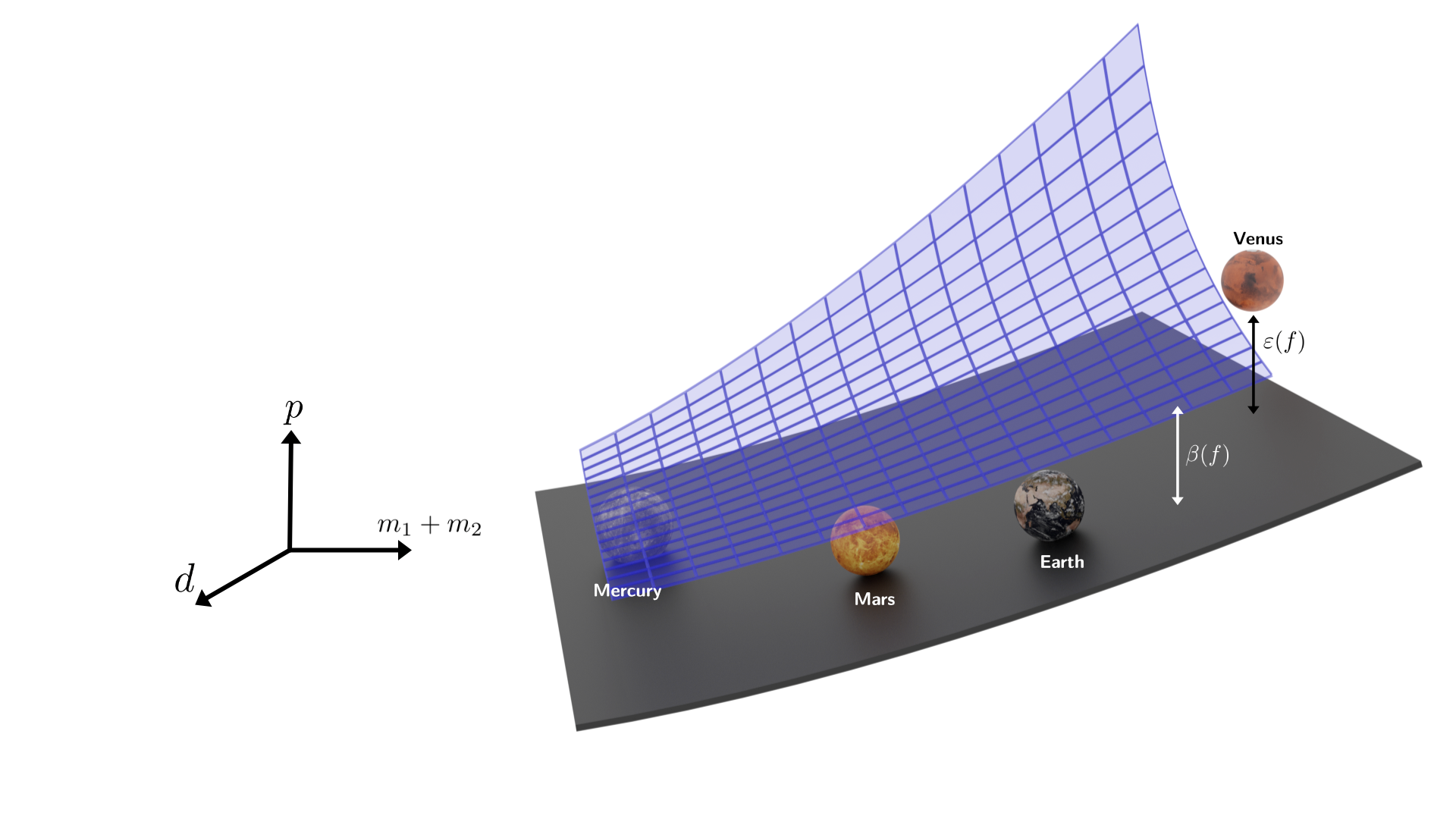}
\caption{Depiction of the numerical data, background theory, and a discovered model 
for Kepler's third law of planetary motion giving the orbital period of a planet in the solar system. 
The data consists of measurements $(m_1, m_2, d, p)$ of the mass of the sun $m_1$, the orbital period $p$ and mass $m_2$ for each planet and its distance $d$ from the sun. 
The background theory amounts to Newton's laws of motion, i.e., the formulae for centrifugal force, gravitational force, and equilibrium conditions. 
The 4-tuples $(m_1, m_2, d, p)$ are projected into $(m_1+m_2, d, p)$. 
The blue manifold represents solutions of $f_{\mathcal{B}}$ which is the function derivable from the background-theory axioms that represents the variable of interest.
The grey manifold represents solutions of the discovered model $f$.
The double arrows indicate the distances 
$\beta(f)$ and $\varepsilon(f)$.%
}
\label{fig:kepler3d} 
\end{figure}

\begin{itemize}%
\item {\bf Background Knowledge} \,$\mathcal B$: a set  of domain-specific axioms expressed as logic formulae\footnote{In our implementation we focus on first-order-logic formulae with equality, inequality and basic arithmetic operators.}. 
They involve $\mathbf{x}$, and $y$, and possibly more variables that %
are necessary to formulate the background theory.  
We assume that the background theory $\mathcal B$ is {\em complete}, that is, it contains all the axioms necessary to comprehensively explain the phenomena under consideration, and {\em consistent}, that is, the axioms do not contradict one another.
These two assumptions guarantee that there exists a unique derivable function\footnote{
Note that although the derivable function is unique, there may exist different functional forms that are equivalent on the domain of interest. 
Considering the domain $\{0,1\}$ for a variable $x$, the two functional forms $f(x)=x$ and $f(x)=x^2$ both define the same function.
}
$f_{\mathcal{B}}$ that logically represents the variable of interest $y$. 
\item {\bf A Hypothesis Class} \,$\mathcal C$: a set of admissible symbolic models defined by a grammar, a set of invariance constraints to avoid redundant expressions (e.g., $A+B$ is equivalent to $B+A$) and constraints on the functional form (e.g., monotonicity).

\item {\bf Data} \,$\mathcal D$: a set of $m$ examples, each providing certain values for $x_1, \ldots, x_n$, and $y$. 
\item {\bf Modeler Preferences} \,$\mathcal M$: a set of numerical parameters, e.g., error bounds on accuracy.
\end{itemize}

\parindent=18pt
In general, there may not exist a function $f \in \mathcal C$ that fits the data exactly and is derivable from $\mathcal B$.
This could happen because
the symbolic model generating the data might not belong to $\mathcal C$,
the sensors used to collect the data might give noisy measurements, 
or the background knowledge might be inaccurate or incomplete. 
To quantify the compatibility of a symbolic model with data and background theory, we introduce
the notion of \emph{distance} between a model $f$ and $\mathcal B$. 
Roughly, it reflects the error between the predictions of $f$ and the predictions of a formula $f_{\mathcal{B}}$ derivable from $\mathcal B$ (thus, the distance equals zero when $f$ is derivable from $\mathcal B$).
Figure~\ref{fig:kepler3d} provides a visualization of these two notions of distance for the problem of learning Kepler's third law of planetary motion from solar-system data and background theory.

Our system consists mainly of an SR-module and a reasoning module.
The SR-module returns multiple candidate symbolic models (or formulae) expressing $y$ as a function of $x_1, \ldots, x_n$ and that fit the data. 
For each of these models, the system outputs the distance $\varepsilon(f)$ between $f$ and $\mathcal D$ and the distance $\beta(f)$ between $f$ and $\mathcal B$. We will also be referring to $\varepsilon(f)$ and $\beta(f)$
as \emph{errors}.

\begin{figure}[t]
\centering
\includegraphics[width=0.8\textwidth]{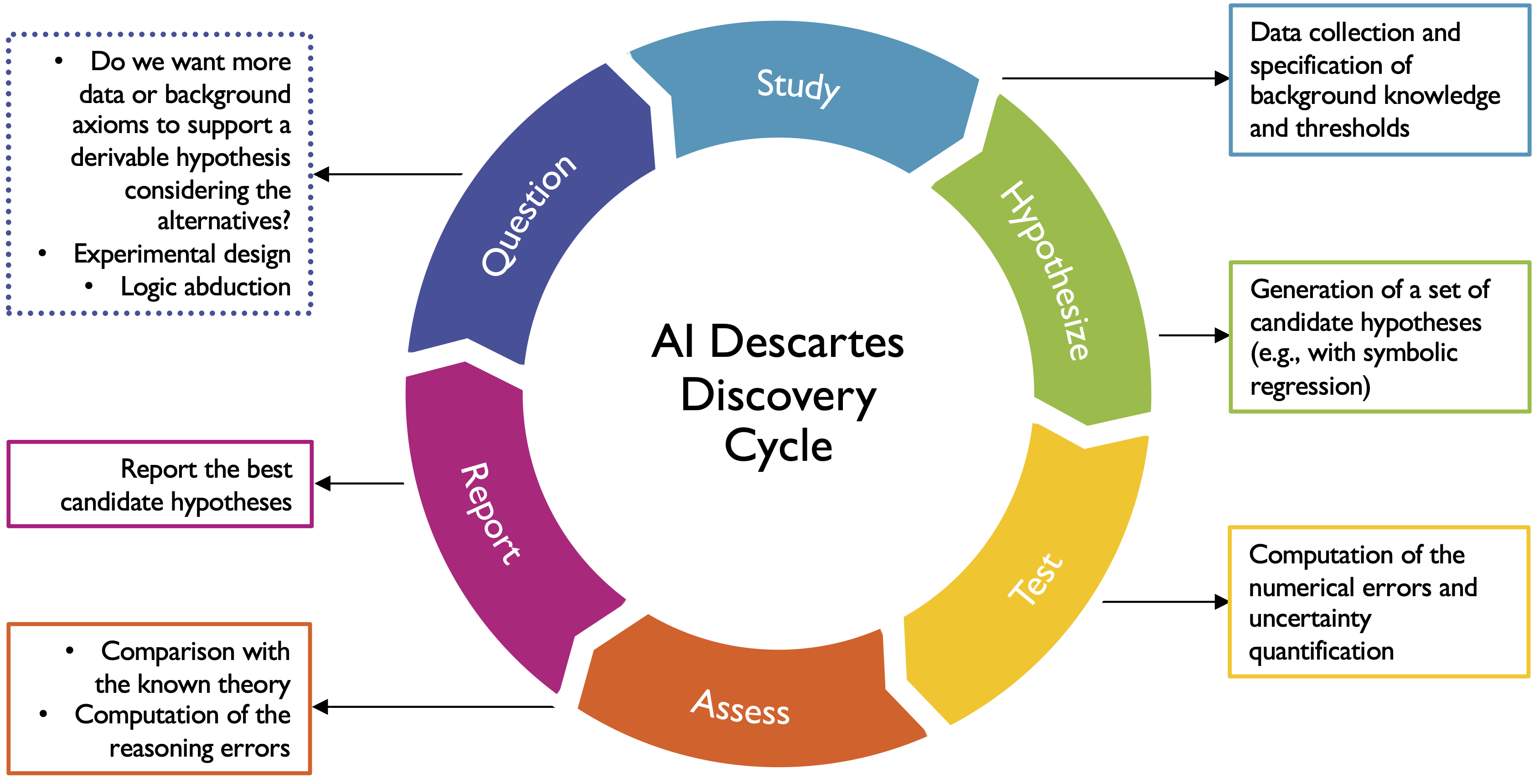}
\caption{An interpretation of the scientific method as implemented by our system. The colors match the respective components of the system in Fig~\ref{fig:system_overview}. The box with dotted line boundaries consists of techniques not yet integrated in our system.}
\label{fig:discovery_cycle}
\end{figure}

These functions are also tested to see if they satisfy the specified constraints on the functional form (in $\mathcal C$) and the modeler-specified level of accuracy and complexity (in $\mathcal M$).
When the models are passed to the reasoning module (along with the background theory $\mathcal B$), they are tested for derivability.
If a model is found to be derivable from $\mathcal B$, it is returned as the chosen model for prediction; otherwise, 
if the reasoning module concludes that no candidate model is derivable, it is necessary to either collect additional data or add constraints.
In this case, the reasoning module will return a quality assessment of the input set of candidate hypotheses based on the distance $\beta$, removing models that do not satisfy the modeler-specified bounds on $\beta$. 
The distance (or error) $\beta$ is computed between a function (or formula) $f$, derived from numerical data, and the derivable function $f_{\mathcal{B}}$ which is implicitly defined by the set of axioms in $\mathcal B$ and is logically represented by the variable of interest $y$.
The distance between the function $f_{\mathcal{B}}$ and any other formula $f$ depends only on the background theory and the formula $f$ and not on any particular functional form of $f_{\mathcal{B}}$. 
Moreover, the reasoning module can prove that a model is not derivable by returning counterexample points that satisfy $\mathcal B$ but do not fit the model.

SR is typically solved with genetic programming (GP)~\cite{koza, koza2,augusto2000symbolic, schmidt2009distil}, however methods based on mixed-integer nonlinear programming (MINLP) have recently been proposed~\cite{Austel2017, cozadthesis, cozsah}. In this work, we develop a new MINLP-based SR solver (described in the supplementary material).
The input consists of a subset of the operators $\{+, -, \times, \div, \sqrt{\ }, \log, \exp\}$, an upper bound on expression complexity, and an upper bound on the number of constants used that do not equal $1$.
Given a dataset, the system formulates multiple MINLP instances to find an expression that minimizes the least-square error. 
Each instance is solved approximately, subject to a time limit.
Both linear and nonlinear constraints can be imposed.
In particular, dimensional consistency is imposed when physical dimensions of variables are available.

We use KeYmaera X~\cite{keymaera} as a reasoning tool; it is an ATP for hybrid systems and combines different types of reasoning: deductive, real-algebraic, and computer-algebraic reasoning.
We also use Mathematica~\cite{mathematica} for certain types of analysis of symbolic expressions.
While a formula found by any grammar-based system (such as a SR system) is syntactically correct, it may contradict the axioms of the theory or not be derivable from them. 
In some cases, a formula may not be derivable as the theory may not have enough axioms;  the formula may be provable under an extended axiom set (possibly discovered by abduction) or an alternative one (e.g., using a relativistic set of axioms rather than a ``Newtonian'' one).

An overview of our system seen as a discovery cycle is shown in Figure~\ref{fig:discovery_cycle}.
Our discovery cycle is inspired by Descartes who advanced the scientific method and emphasized the role that logical deduction, and not empirical evidence alone, plays in forming and validating scientific discoveries.
Our present approach differs from implementations of the scientific method that obtain hypotheses from theory and then check them against data; instead we obtain hypotheses from data and assess them against theory.

\begin{figure}[t]
\includegraphics[width=\linewidth]{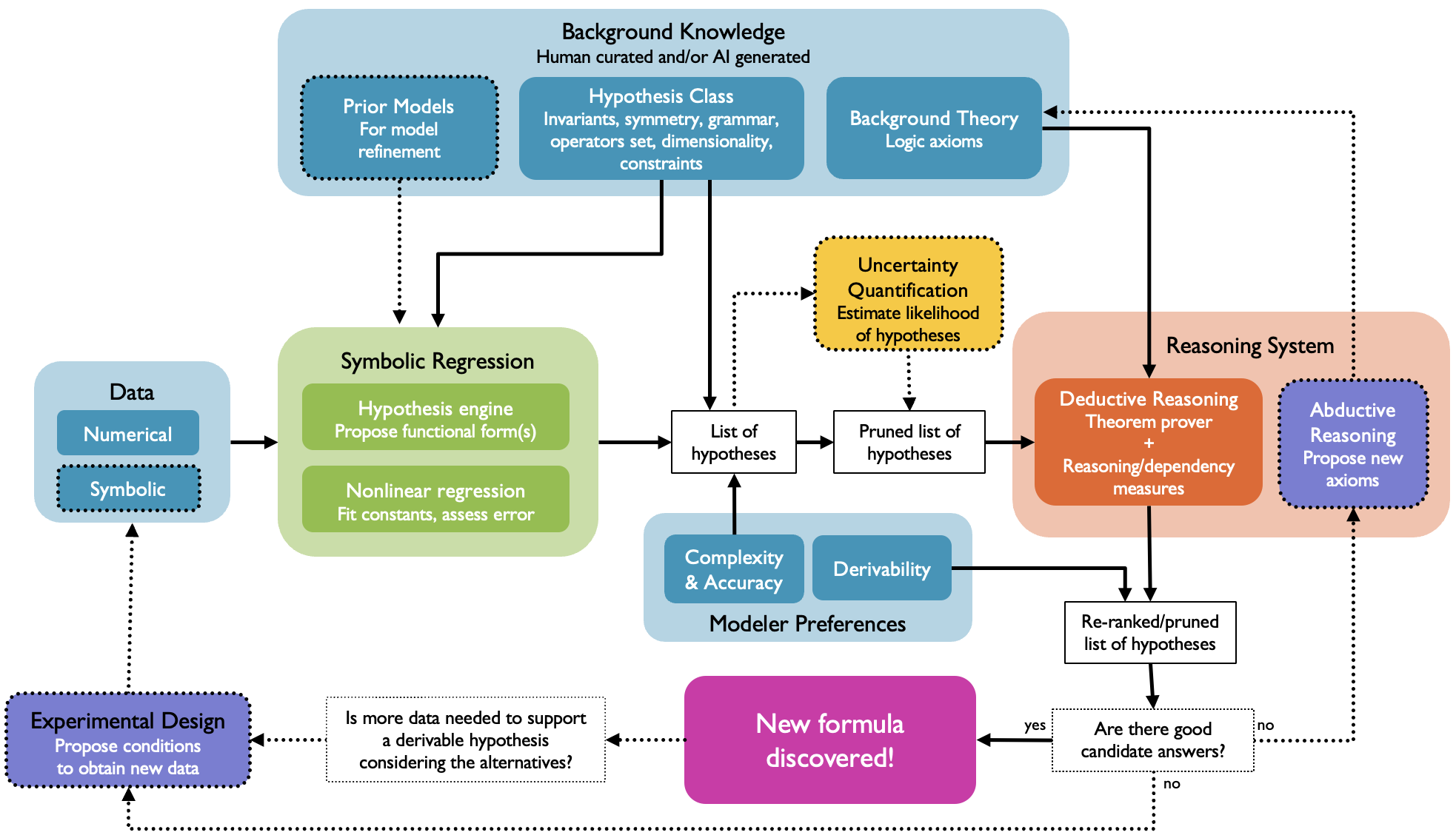}
\captionof{figure}{System overview. Solid lines and boundaries correspond to our system, and dotted lines and boundaries indicate standard techniques for scientific discovery (human-driven or artificial) that have not been integrated into the current system. The present system generates hypotheses from data using symbolic regression, which are posed as conjectures to an automated deductive reasoning system, which proves or disproves them based on background theory or provides reasoning-based quality measures.}
\label{fig:system_overview}
\end{figure}

A more detailed schematic of the system is depicted in Figure~\ref{fig:system_overview}, where the bold lines and boundaries correspond to the system we present in this work, and the dashed lines and boundaries refer to standard techniques for scientific discovery that we have not yet integrated into our current implementation.

\section*{Results}
We tested the different capabilities of our system on three problems (more details in Methods).
First, we considered the problem of deriving Kepler's third law of planetary motion, providing reasoning-based measures to analyze the quality and generalizablity of the generated formulae.
Extracting this law from experimental data is challenging, especially when the masses involved are of very different magnitudes. This is the case for the solar system, where the solar mass is much larger than the planetary masses. 
The reasoning module helps in choosing between different candidate formulae and identifying the one that generalizes well: using our data and theory integration we were able to re-discover Kepler's third law.
We then considered Einstein's time-dilation formula. Although we did not recover this formula from data, we used the reasoning module to identify the formula that generalizes best. Moreover, analyzing the reasoning errors with two different sets of axioms (one with ``Newtonian” assumptions and one relativistic), we were able to identify the theory that better explains the phenomenon.
Finally, we considered Langmuir's adsorption equation,
whose background theory contains material-dependent coefficients. By relating these coefficients to the ones in the SR-generated models via existential quantification, we were able to logically prove one of the extracted formulae.

To conclude, we have demonstrated the value of combining logical reasoning with symbolic regression in obtaining meaningful symbolic models of physical phenomena, in the sense that they are consistent with background theory and generalize well in a domain that is significantly larger than the experimental data. The synthesis of regression and reasoning yields better models than can be obtained by SR or logical reasoning alone.

Improvements or replacements of individual system components and introduction of new modules such as abductive reasoning or experimental design (not described in this work for the sake of brevity) would extend the capabilities of the overall system.

A deeper integration of reasoning and regression can help synthesize models that are both data driven and based on first principles, and lead to a revolution in the scientific discovery process. 
The discovery of models that are consistent with prior knowledge will accelerate scientific discovery, and enable going beyond existing discovery paradigms.

\section*{Reproducibility}
The code used for this work along with the datasets described below can be found, freely available, at:
\\
\noindent\href{https://github.com/IBM/AI-Descartes}{https://github.com/IBM/AI-Descartes}.

\clearpage
\noindent
{\huge \bf Methods}

\section*{Kepler's third law of planetary motion.}

Kepler's law relates the distance $d$ between two bodies, e.g., the sun and a planet in the solar system, and their orbital periods. 
It can be expressed as
\begin{equation}
p  = \sqrt{ \frac{4\pi^2 d^3 }{G (m_1 + m_2)} }~~,
\end{equation}
where 
$p$ is the period, 
$G$ is the gravitational constant, 
and $m_1$ and $m_2$ are the two masses. 
It can be derived using the following axioms of the background theory $\mathcal{B}$, describing the center of mass (axiom K1), the distance between bodies (axiom K2), the gravitational force (axiom K3), the centrifugal force (axiom K4), the force balance (axiom K5), and the period (axiom K6):

{\small
\begin{flalign*}
&\begin{aligned}
\hspace{1.3cm} \text{K1. } & m_1 * d_1 = m_2 * d_2         \\
\text{K2. }  & d = d_1 + d_2    \\[4pt]     
\text{K3. }  & F_g = \frac{G m_1 m_2}{d^2} \\[4pt]
\text{K4.  }  & F_c = m_2 d_2 w^2 \\[4pt]
\text{K5. }    & F_g = F_c   \\[4pt]
\text{K6.  }   & p = \frac{2\pi}{w} \\[4pt]
\text{K7.  }   & m_1 > 0, ~ m_2 > 0, ~ p > 0, ~ d_1 > 0, ~ d_2 > 0~~.
\end{aligned}&
\end{flalign*}
}
We considered three real-world datasets: 
planets of the solar system,%
\footnote{https://nssdc.gsfc.nasa.gov/planetary/factsheet/} 
the solar-system planets along with exoplanets from Trappist-1 and the GJ 667 system,%
\footnote{NASA exoplanet archive https://exoplanetarchive.ipac.caltech.edu/}
and binary stars \cite{binary_star}. 
These datasets contain measurements of pairs of masses, a sun and a planet for the first two, and two suns for the third, the distance between them, and the orbital period of the planet around its sun in the first two datasets or the orbital period around the common center of mass in the third dataset.
The data we use is given in the supplementary material. Note that the dataset does not contain measurements for a number of variables in the axiom system, such as $d_1, d_2, F_g,$ etc.
The goal is to recover Kepler's third law from the data, i.e., to obtain $p$ as the above-stated function of $d$, $m_1$ and $m_2$. 
The SR-module takes as input the set
$\{+,-,\times,\div,\sqrt{\ }\}$ and outputs a set of candidate formulas.

None of the formulae obtained via SR are derivable, though some are close approximations to derivable formulae.
We evaluate the quality of these formulae 
by writing a logic program for calculating the error  $\beta$ of a formula with respect to a derivable formula. 

We use three measures, defined below, to assess the correctness of a data-driven formula from a reasoning viewpoint: the {\em pointwise reasoning error},
the {\em generalization reasoning error},
and {\em variable dependence}. 

\paragraph{\it Pointwise reasoning error:}
The key idea is to compute a distance between a formula generated from the numerical data and some derivable formula that is implicitly defined by the axiom set. 
The distance is measured by the $\ell_2$ or $\ell_{\infty}$ norm applied to the differences between the values of the numerically-derived formula and a derivable formula at the points in the dataset. 

These results can be extended to other norms.

We compute the relative error of numerically derived formula $f(\mathbf{x})$ applied to the $m$ data points $\mathbf{X}^i~(i=1, \ldots, m)$
with respect to $f_{\mathcal{B}}(\mathbf{x})$, derivable from the axioms via these expressions\footnote{$f_{\mathcal{B}}(\mathbf{X}^i)$ denotes a derivable formula for the variable of interest $y$ evaluated at the data point $\mathbf{X}^i$.}:
\begin{equation}\label{eq:relerror}
    \beta_2^r  = \sqrt{\sum_{i=1}^m \left(\frac{f(\mathbf{X}^i) -  f_{\mathcal{B}}(\mathbf{X}^i)}{f_{\mathcal{B}}(\mathbf{X}^i)}\right)^2} ~~~  
    \text{ and } ~~~ 
     \beta_\infty^r  = \max_{1\le i \le m} \left\{\frac{|f(\mathbf{X}^i) -  f_{\mathcal{B}}(\mathbf{X}^i)|}{|f_{\mathcal{B}}(\mathbf{X}^i)|}\right\}~.
\end{equation}

The \keymaera formulation of these two measures for the first formula of Table~\ref{tab:kepler} can be found in the supplementary material. 
Absolute-error variants of the first and second expressions in (\ref{eq:relerror}) are denoted by $\beta^a_2, \beta^a_\infty$, respectively. The numerical (data) error measures $\varepsilon_2^r$ and $\varepsilon_\infty^r$ are defined by replacing $f_{\mathcal{B}}(\mathbf{X}^i)$ by $Y_i$ in (\ref{eq:relerror}). Analogous to $\beta^a_2$ and $\beta^a_\infty$, we also define absolute-numerical-error measures $\varepsilon^a_2$ and $\varepsilon^a_\infty$. 

Table~\ref{tab:kepler} 
reports in columns 5 and 6 the values of $\beta_2^r$ and $\beta_{\infty}^r$, respectively. 
It also reports the relative numerical errors $\varepsilon_2^r$ and $\varepsilon_{\infty}^r$ in columns 3 and 4, measured by the $\ell_2$ and $\ell_\infty$ norms, respectively, for the candidate expressions given in column 2 when evaluated on the points in the dataset.\footnote{We minimize the absolute $\ell_2$ error $\varepsilon_2^a$ (and not the relative error $\varepsilon_2^r$), when obtaining candidate expressions via symbolic regression.}

The pointwise reasoning errors $\beta_2$ and $\beta_{\infty}$ are not very informative if SR yields a low-error candidate expression (measured with respect to the data), 
and the data itself satisfies the background theory up to a small error, 
which indeed is the case with the data we use; the reasoning errors and numerical errors are very similar.

\paragraph{\it Generalization reasoning error:}
Even when one can find a function that fits given data points well, it is challenging to obtain a function that {\em generalizes} well, i.e., one which yields good results at points of the domain not equal to the data points.
Let $\beta^r_{\infty,S}$ be calculated for a candidate formula $f(\mathbf{x})$ over a domain $S$ that is not equal to the original set of data points as follows:
\begin{equation}
    \beta_{\infty,S}^r  = \max_{\mathbf{x} \in S} \left\{\frac{|f(\mathbf{x}) - f_{\mathcal{B}}(\mathbf{x}) |}{|f_{\mathcal{B}}(\mathbf{x})|}\right\}~,
\end{equation}
where we consider the relative error and, as before, the function $f_{\mathcal{B}}(\mathbf{x})$ is not known, but is implicitly defined by the axioms in the background theory.
We call this measure the {\em relative generalization reasoning error}.
If we do not divide by $f_{\mathcal{B}}(\mathbf{x})$ in the above expression, we get the \emph{absolute} version $\beta_{\infty,S}^a$ of this error metric. For the Kepler dataset, we let $S$ be the smallest multi-dimensional interval (or Cartesian product of intervals on the real line) containing all data points.
In column 7 of Table~\ref{tab:kepler}, 
we show the relative generalization reasoning error $\beta_{\infty,S}^r$ on the Kepler datasets
with $S$ defined as above. If this error is roughly the same as $\beta_{\infty}^r$ the pointwise relative reasoning error for $\ell_{\infty}$, e.g., for the solar system dataset, then the formula extracted from the numerical data is as accurate at points in $S$ as it is at the data points.
\paragraph
{\it Variable dependence:}
In order to check if the functional dependence of a candidate formula on a specific variable is accurate, we compute the generalization error over a domain $S$ where the domain of this variable is extended by an order of magnitude beyond the smallest interval containing the values of the variable in the dataset. 
Thus we can check whether there exist special conditions under which the formula does not hold. 
We modify the endpoints of an interval by one order of magnitude, one variable at a time.
If we notice an increase in the generalization reasoning error while modifying intervals for one variable, we deem the candidate formula as missing a dependency on that variable.
A missing dependency might occur, for example, because the exponent for a variable is incorrect,
or that variable is not considered at all when it should be. 
One can get further insight into the type of dependency by analyzing how the error varies, e.g., linearly or exponentially.
Table~\ref{tab:kepler} 
provides, in columns 8--10, results regarding the candidate formulae for Kepler's third law. 
For each formula, the dependencies on $m_1$, $m_2$, and $d$
are indicated by $1$ or $0$ (for correct or incorrect dependency).
For example, the candidate formula 
$p = \sqrt{0.1319 \cdot d^3 }$ for the solar system
does not depend on either mass, and the dependency analysis suggests that the formula approximates well the phenomenon in the solar system, but not for  larger masses.

The best formula for the binary-star dataset, $\sqrt{d^3/(0.9967m_1+m_2)}$, has no missing dependency (all ones in columns 8--10), i.e., it generalizes well;
increasing the domain along any variable does not increase the generalized reasoning error.

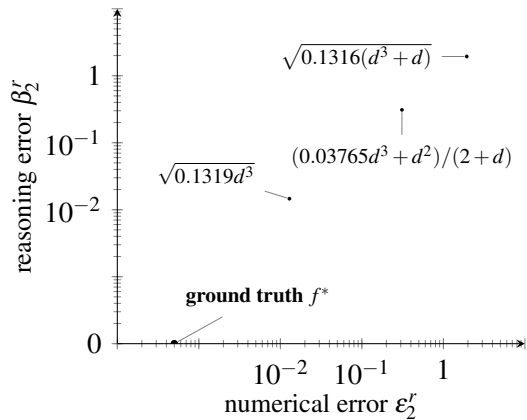
\begin{figure}[H]
\centering
\begin{tikzpicture}
    \begin{axis}[
    axis lines = left,
    xmode=log,
            ymode=log,
            xmin=1e-4,
            xmax=10,
            ymin=1e-4,
            ymax=10,
            xlabel={\small numerical error $\varepsilon^r_2$},
            ylabel={\small reasoning error $\beta^r_2$},
            xtick pos=left,
            ytick pos=left,
            xticklabels={$ $, $ $, $10^{-2}$, $10^{-1}$, $1$, $ $},xtick={0.0001, 0.001, 0.01, 0.1, 1, 10},
            yticklabels={ $0$,  $ $, $10^{-2}$, $10^{-1}$, $1$, $ $},ytick={0.0001, 0.001, 0.01, 0.1, 1, 10},
            width=7cm]
        \node[pin={[pin distance=.3cm]87:{\scriptsize \bf %
        ground truth $f^*$}},circle, fill, inner sep=1pt] at (axis cs:.0005,.0001) {};
        \node[pin={[pin distance=.3cm]-190:{\scriptsize $\sqrt{0.1319 d^3 }$}},circle,fill,inner sep=.5pt] at (axis cs:.01291,.0146) {};
        \node[pin={[pin distance=.3cm]180:{\scriptsize $ \sqrt{0.1316(d^3 + d)}$}},circle,fill,inner sep=.5pt] at (axis cs:1.9348,1.9385) {};
        \node[pin={[pin distance=.3cm]-90:{ \scriptsize $(0.03765 d^3 + d^2)/ (2+d)$}},circle,fill,inner sep=.5pt] at (axis cs:.3102,.3095) {};
    \end{axis}
\end{tikzpicture}
\caption{
Depiction of symbolic models for Kepler's third law of planetary motion giving the orbital period of a planet in the solar system. The model are 
produced by our SR system by points $(\varepsilon,\beta)$, where $\varepsilon$ represents distance to data, and $\beta$ represents distance to background theory. Both distances are computed with an appropriate norm on the scaled data.
}
\label{fig:keplerpareto}
\end{figure}

Figure~\ref{fig:keplerpareto} provides a visualization of the two errors $\varepsilon^r_2$ and $\beta^r_2$ for the first three functions of Table~\ref{tab:kepler} (solar-system dataset) and the ground truth $f^*$.

\begin{table*}[ht]
\centering
\resizebox{\linewidth}{!}{
\begin{tabular}{llccccccccccc}  
~~~~~1 &  \multicolumn{1}{c}{2} & 3 & 4 & & 5 & 6 & & 7 &  & 8 & 9 & 10 \\
\toprule
 & Candidate formula &\multicolumn{2}{c}{numerical error}  & & \multicolumn{2}{c}{point. reas. err.}  &  & gen. reas. & & \multicolumn{3}{c}{dependencies} \\
 Dataset &  $p=$ & $\varepsilon_2^r$ & $\varepsilon^r_{\infty}$ & & $\beta_2^r$ & $\beta^r_{\infty}$ & & error $\beta^r_{\infty,S}$ &  & $m_1$ & $m_2$ & $d$ \\
\midrule
\multirow{4}{*}{solar}  
    & $ \sqrt{0.1319 d^3 }$  & .01291 & .006412 && .0146  & .0052  && .0052  && 0 & 0 & 1\\[5pt]
    & $ \sqrt{0.1316 (d^3 + d)}$   & 1.9348 & 1.7498  && 1.9385 & 1.7533 && 1.7559 && 0 & 0 & 0\\[5pt]
    & $ (0.03765 d^3 + d^2)/(2+d)$ & .3102  & .2766   && .3095  & .2758  && .2758  && 0 & 0 & 0\\
\midrule
\multirow{4}{*}{exoplanet}  
    & $ \sqrt{0.1319 d^3/m_1}$                  & .08446 & .08192 && .02310 & .0052 && .0052 && 0 & 0 & 1 \\[5pt]
    & $ \sqrt{m_1^2 m_2^3/d + 0.1319\,d^3/m_1}$ & .1988  & .1636  && .1320  & .1097 && $ > 550$ && 0 & 0 & 0\\[5pt]
    & $ \sqrt{(1-.7362 m_1) d^3/2}$             & 1.2246 & .4697  && 1.2418 & .4686 && .4686 && 0 & 0 & 1 \\
\midrule
\multirow{5}{*}{binary stars}  
& $ 1/(d^2 m_1^2)+ 1/(d m_2^2)-m_1^3 m_2^2~+ $ 
    &  \multirow{2}{*}{.002291} & \multirow{2}{*}{.001467} && \multirow{2}{*}{.0059 } & \multirow{2}{*}{.0050} && \multirow{2}{*}{timeout} && \multirow{2}{*}{0} & \multirow{2}{*}{0} & \multirow{2}{*}{0}\\
    & \hspace{0.2cm} + $\sqrt{.4787d^3/m_2+d^2 m_2^2}$ \\[5pt]
    & $ (\sqrt{d^3}+ m_1^3 m_2/\sqrt{d})/\sqrt{m_1+m_2}$ & .003221 & .003071 &&.0038  & .0031 && timeout && 0 & 0 & 0 \\     [5pt]          
    & $ \sqrt{d^3/(0.9967m_1+m_2)}$ & .005815 & .005337 && .0014 & .0008 && .0020 && 1 & 1 & 1 \\
\bottomrule 
\end{tabular}
}
\caption{
Numerical error values, pointwise reasoning error values,  and generalization error values for candidate solutions for the Kepler dataset. We also give an analysis of the variable dependence of candidate solutions.  
For simplicity of notation, in the table we use the variables $d$, $m_1$, $m_2$ and $p$, while referring to the scaled counterparts. We assume that all the errors are relative.
}
\label{tab:kepler}
\end{table*}

\section*{Relativistic time dilation.}

Einstein's theory of relativity postulates that the speed of light is constant, and implies that two observers in relative motion to each other will experience time differently and observe different clock frequencies. The frequency $\freq$ for a clock moving at speed $v$ is related to the frequency $\freq_0$ of a stationary clock by the formula
\begin{equation}\label{eq:timedilation} \frac{\freq - \freq_0}{\freq_0} = \sqrt{1-\frac{v^2}{c^2}}-1
\end{equation}
where $c$ is the speed of light. This formula was recently confirmed experimentally by Chou et al. \cite{chou2010} using high precision atomic clocks. We test our system on the experimental data of \cite{chou2010} which consists of measurements of $v$ and associated values of $(\freq-\freq_0)/\freq_0$, reproduced in the supplementary material.
We take the axioms for derivation of the time dilation formula from ~\cite{behroozi, gssmith}. These are also listed in the 
supplementary material and involve variables that are not present in the experimental data.

In Table~\ref{tab:timedilationfunctions} we give some functions obtained by our SR module (using $\{+, -, \times, \div, \sqrt{\ }\, \}$ as the set of input operators) along with the numerical errors of the associated functions and generalization reasoning  errors. The sixth column gives the set $S$ as an interval for $v$ for which our reasoning module can verify that the absolute generalization reasoning error of the function in the first column is at most 1. The last column gives the interval for $v$ for which we can verify a relative generalization reasoning error of at most 2\%. Even though the last function has low relative error according to this metric, it can be ruled out as a reasonable candidate if one assumes the target function should be continuous (it has a singularity at $v = 1$). Thus, even though we cannot obtain the original function, we obtain another which generalizes well, as it yields excellent predictions for a very large range of velocities.

\begin{table*}[ht]
\centering
\resizebox{\linewidth}{!}{
\begin{tabular}{lcccccc}  
\toprule
             Candidate     & \multicolumn{2}{c}{Numerical Error}  & \multicolumn{2}{c}{Numerical Error} & $S$ s.t. Absolute & $S$ s.t. Relative \\
 formula & \multicolumn{2}{c}{Absolute}         & \multicolumn{2}{c}{Relative}        & Gen. Reas. Error & Gen. Reas. Error \\
$y =$ & $\varepsilon_2^a$ & $\varepsilon^a_{\infty}$ & $\varepsilon_2^r$ & $\varepsilon^r_{\infty}$ & $\beta^a_{\infty,S} \leq 1$ & $\beta^r_{\infty,S} \leq 0.02$ \\
\midrule
$-0.00563 v^2$                                & .3822 & .3067 & 1.081 & .001824  & $37 \leq v \leq 115$  & $37 \leq v \leq 10^8$\\[2pt]
$\frac{v}{1+0.00689v} - v$                    & .3152 & .2097 & 1.012 & .006927  & $37 \leq v \leq 49$ & $37 \leq v \leq 38$ \\[2pt]
$-0.00537\frac{v^2\sqrt{v + v^2}}{(v-1)}$     & .3027 & .2299 & 1.254 & .002147  & $37 \leq v \leq 98$ & $37 \leq v \leq 109$ \\[2pt]
$-0.00545\frac{v^4}{\sqrt{v^2 +v^{-2}}(v-1)}$ & .3238 & .2531 & 1.131 & .0009792 & $37 \leq v \leq 126$ & $37 \leq v \leq 10^7$\\
\bottomrule
\end{tabular}
}
\caption{Candidate functions derived from time dilation data, and associated error values. The values of $v$ are defined in $m/s$.
}
\label{tab:timedilationfunctions}
\end{table*}

In this case, our system can also help rule out alternative axioms. Consider replacing the axiom that the speed of light is a constant value $c$ by a ``Newtonian'' assumption that light behaves like other mechanical objects: if emitted from an object with velocity $v$ in a direction perpendicular to the direction of motion of the object, it has velocity $\sqrt{v^2 + c^2}$. Replacing $c$ by $\sqrt{v^2+c^2}$ (in axiom R2 in the supplementary material to obtain R2') produces a self-consistent axiom system (as confirmed by the theorem prover), albeit one leading to no time dilation. Our reasoning module concludes that none of the functions in Table~\ref{tab:timedilationfunctions} is compatible
with this updated axiom system: the absolute generalization reasoning error is greater than 1 even on the dataset domain, as well as the pointwise reasoning error.
Consequently, the data is used indirectly to discriminate between axiom systems relevant for the phenomenon under study; SR poses only accurate formulae as conjectures. 

\section*{Langmuir's adsorption equation.}
The Langmuir adsorption equation describes a chemical process in which gas molecules contact a surface, and relates the loading $q$ on the surface to the pressure $p$ of the gas%
\footnote{Langmuir was awarded the Nobel Prize in Chemistry 1932 for this work.}
\cite{Langmuir1918}:
\begin{equation}
    q = \frac{q_{\mathrm{max}} K_{\mathrm{a}} \cdot p}{1+K_{\mathrm{a}} \cdot p}.
    \label{eq:Langmuir}
\end{equation}
The constants $q_{\mathrm{max}}$ and $K_a$
characterize the maximum loading and the adsorption strength, respectively.
A similar model for a material with two types of adsorption sites yields:
\begin{equation}
    q = \frac{q_{\mathrm{max,1}} K_{\mathrm{a,1}} \cdot p}{1+K_{\mathrm{a,1}} \cdot p} + \frac{q_{\mathrm{max,2}} K_{\mathrm{a,2}} \cdot p}{1+K_{\mathrm{a,2}} \cdot p}~,
    \label{eq:Langmuir2site}
\end{equation}
with parameters for maximum loading and adsorption strength on each type of site. 
The parameters in (\ref{eq:Langmuir}) and (\ref{eq:Langmuir2site}) fit experimental data using linear or nonlinear regression, and depend on the material, gas, and temperature.

We used data from \cite{Langmuir1918} %
for methane adsorption on mica at a temperature of 90 K, and also data from \cite[Table 1]{sunetal1998} for isobutane adsorption on silicalite at a temperature of 277 K. 
In both cases, observed values of $q$ are given for specific values of $p$; %
the goal is to express $q$ as a function of $p$.  
We give the SR-module the operators $\{+, -, \times, \div\}$, 
and obtain the best fitting functions with two and four constants.
The code ran for 20 minutes on 45 cores, and seven of these functions are displayed for each dataset. 

To encode the background theory, %
following Langmuir’s original theory \cite{Langmuir1918}
we elicited the following set $\mathcal{A}$ of axioms:\\

\parbox{2.5in}{
\begin{tabular}{llll}
& L1. & Site balance: &
 $S_0 = S+S_{\mathrm{a}} $            \\[4pt]
& L2. & Adsorption rate model: &
    $r_{\mathrm{ads}} = k_{\mathrm{ads}} \cdot p \cdot S $    \\[4pt]     
& L3. & Desorption rate model: &
    $r_{\mathrm{des}} = k_{\mathrm{des}} \cdot S_{\mathrm{a}} $ \\[4pt]
& L4.  & Equilibrium assumption: &
    $r_{\mathrm{ads}} = r_{\mathrm{des}}$ \\[4pt]
& L5.   & Mass balance on $q$ & 
    $q = S_{\mathrm{a}}$  ~~. \\
\end{tabular}
}
~\\

Here, $S_0$ is the total number of sites, of which $S$ are unoccupied and $S_a$ are occupied (L1). 
The adsorption rate $r_{\mathrm{ads}}$ is proportional to the pressure $p$ and the number of unoccupied sites (L2). 
The desorption rate $r_{\mathrm{des}}$ is proportional to the number of occupied sites (L3). 
At equilibrium, $r_{\mathrm{ads}} = r_{\mathrm{des}}$ (L4), and the total amount adsorbed, $q$, is the number of occupied sites (L5) because the model assumes each site adsorbs at most one molecule. 
Langmuir solved these equations to obtain:
\begin{equation}
    q = \frac{S_0 * (k_{\mathrm{ads}}/k_{\mathrm{des}}) * p}{1 + (k_{\mathrm{ads}}/k_{\mathrm{des}})*p}.
    \label{eq:LangmuirFull}
\end{equation}
which corresponds to Eq.~\ref{eq:Langmuir}, where $q_{\mathrm{max}} = S_0$ and $K_{\mathrm{a}} = k_{\mathrm{ads}}/k_{\mathrm{des}}$. An axiomatic formulation for the multi-site Langmuir expression is described in the supplementary material.
Additionally, constants and variables are constrained to be positive, e.g., $S_0>0$, $S>0$, and $S_{\mathrm{a}}>0$, or non-negative, e.g., $q \geq 0$. 
 
The logic formulation to prove is:
\begin{equation}\label{formulation:langmuir}
 (\mathcal{C} \wedge \mathcal{A}) \rightarrow f~,
\end{equation}
where $\mathcal{C}$ is the conjunction of the non-negativity constraints, %
$\mathcal{A}$ is a conjunction of the axioms, the union of $\mathcal{C}$ and $\mathcal{A}$ constitutes the background theory $\mathcal{B}$,
and $f$ is the formula we wish to prove, 
e.g., (\ref{eq:LangmuirFull}). 

SR can only generate numerical expressions involving the 
(dependent and independent)
variables occurring in the input data, with certain values for constants;
for example, the expression $f= {p}/{(0.709 \cdot p + 0.157)}$. The expressions built from variables and constants from the background theory, such as \eqref{eq:LangmuirFull}, involve the constants (in their symbolic form) explicitly: e.g., $k_{\mathrm{ads}}$ and $k_{\mathrm{des}}$ appear explicitly in \eqref{eq:LangmuirFull} while SR only generates a numerical instance of the ratio of these constants. 
Thus, we cannot use (\ref{formulation:langmuir}) directly to prove formulae generated from SR. 
Instead, we replace each numerical constant of the formula by a logic variable $c_i$\,; for example, the formula $f= p/(0.709\cdot p+0.157)$ is replaced by $f'=p/(c_1\cdot p + c_2)$, introducing two new variables $c_1$ and $c_2$.
We then quantify the new variables existentially, and define a new set of non-negativity constraints $\mathcal{C}'$.
In the example above we will have $\mathcal{C}'= c_1>0~\wedge~c_2>0$. 

The final formulation is:%
\begin{equation}\label{formulation:langmuir2}
 \exists c_1 \cdots \exists c_n.~ (\mathcal{C} \wedge \mathcal{A})  \rightarrow (f' \wedge \mathcal{C}')~.
\end{equation}

For example, $f'=p/(c_1\cdot p + c_2)$ is proved true if the reasoner can prove that there exist values of $c_1$ and $c_2$  such that $f'$ satisfies the background theory $\mathcal{A}$ and the constraints $\mathcal{C}$. Here $c_1$ and $c_2$ can be functions of constants $k_{\mathrm{ads}}$, $k_{\mathrm{des}}$, $S_0$, and/or real numbers, but not the variables $q$ and $p$.

We also consider background knowledge in the form of a list of desired properties of the relation between $p$ and $q$, which helps trim the set of candidate formulae. 
Thus, we define a collection $\mathcal K$ of constraints on $f$, where $q = f(p)$, enforcing monotonicity or certain types of limiting behavior (see supplementary material).
We use  Mathematica~\cite{mathematica} to verify that a candidate function satisfies the constraints in $\mathcal{K}$.

\begin{table*}[t]
\centering
\footnotesize
\resizebox{\linewidth}{!}{
\begin{tabular}{cclcccc} 
~~~1 & ~~~~2 & \multicolumn{1}{c}{3} & 4 & 5  & 6 & 7 \\
\toprule
 & & \multicolumn{1}{c}{Candidate formula}  & \multicolumn{2}{c}{Numerical Error}  & & $\mathcal{K}$\\
 Data & Condition & \multicolumn{1}{c}{$q=$} & $\varepsilon_2^r$  &  $\varepsilon_{\infty}^r$ & provability &  constr. \\
\midrule
\parbox[t]{2mm}{\multirow{7}{*}{\rotatebox[origin=c]{90}{Langmuir \cite[Table IX]{Langmuir1918} ~~}}}
& \multirow{3}{*}{2 const.}  
& $f_1:$ ~~$(p^2+2p-1)/(.00888p^2+.118p)$ & .06312 & .04865 & timeout  & 2/5\\[5pt]
    & & $f_2:$ ~~$p/(.00927p+.0759)$ *  &  .1799 & .1258 & Yes & 5/5\\[5pt]
\cmidrule{2-7}
& \multirow{4}{*}{4 const.} 
    & $f_3:$ ~~${(p^2-10.5p-15.)}/{(.00892p^2-1.23)}$  &  .04432 & .02951 & timeout  & 2/5\\[5pt]
    & & $f_4:$ ~~${(8.86p+13.9)}/{(.0787p+1)}$  &  .06578 & .04654 & No  & 4/5\\[5pt]
    & & $f_5:$ ~~${p^2}/{(.00895p^2+.0934p-.0860)}$  &  .07589 & .04959 & No & 2/5\\[5pt]
\cmidrule{2-7}
& \multirow{2}{*}{4 const.} 
    & $f_6:$ ~~${(p^2+p)}/{(.00890p^2+.106p-.0311)}$  &  .06833 & .04705 & timeout  & 2/5\\[5pt]
& \multirow{1}{*}{extra-point} & $f_7:$ ~~${(112p^2-p)}/{(p^2+10.4p-9.66)}$  &  .07708 & .05324 & timeout  & 3/5\\[5pt]
\midrule
\parbox[t]{2mm}{\multirow{7}{*}{\rotatebox[origin=c]{90}{Sun et al. \cite[Table 1]{sunetal1998} ~~}}}
& \multirow{3}{*}{2 const.}
     & $g_1:$ ~~${(p + 3)}/{(.584p + 4.01)}$ & .1625 & .1007 & No & 4/5 \\[5pt] 
    & & $g_2:$ ~~${p}/{(.709p + .157)}$ & .9680 & .5120 & Yes & 5/5\\[5pt]
     \cmidrule{2-7}
    & \multirow{3}{*}{4 const.}
     & $g_3:$ ~~${(.0298p^2 + 1)}/{(.0185p^2 + 1.16)} - {.000905}/{p^2}$ & .1053 & .05383 & timeout & 2/5 \\[5pt] 
     && $g_4:$ ~~${1}/{(p^2+1)} + {(2.53p - 1)}/{(1.54p+2.77)}$ & .1300 & .07247 & timeout & 3/5 \\[5pt]
     \cmidrule{2-7}
    & \multirow{3}{*}{4 constants} & $g_5:$ ~~${(1.74p^2 + 7.61p)}/{(p^2 + 9.29p + 0.129)}$ & .1119 & .0996 & timeout & 5/5\\[5pt] 
    & \multirow{2}{*}{extra-point} & $g_6:$ ~~${(.226p^2 + .762p - 7.62 \cdot 10^{-4})}/{(.131p^2+p)}$ & .1540 & .09348 & timeout & 2/5\\[5pt]
    & & $g_7:$ ~~${(4.78p^2+26.6p)}/{(2.71p^2+30.4p+ 1.)}$ & .1239 & .1364 & timeout & 5/5 \\[5pt]
\bottomrule 
\end{tabular}
}
\caption{
Results on two datasets for the Langmuir problem. 
(*This expression is also generated when using four constants, and also when an extra point approximating  $(0,0)$ is added).
}
\label{tab:langmuir}
\end{table*}
\begin{figure}[H]
\centering
\includegraphics[width=0.8\columnwidth]{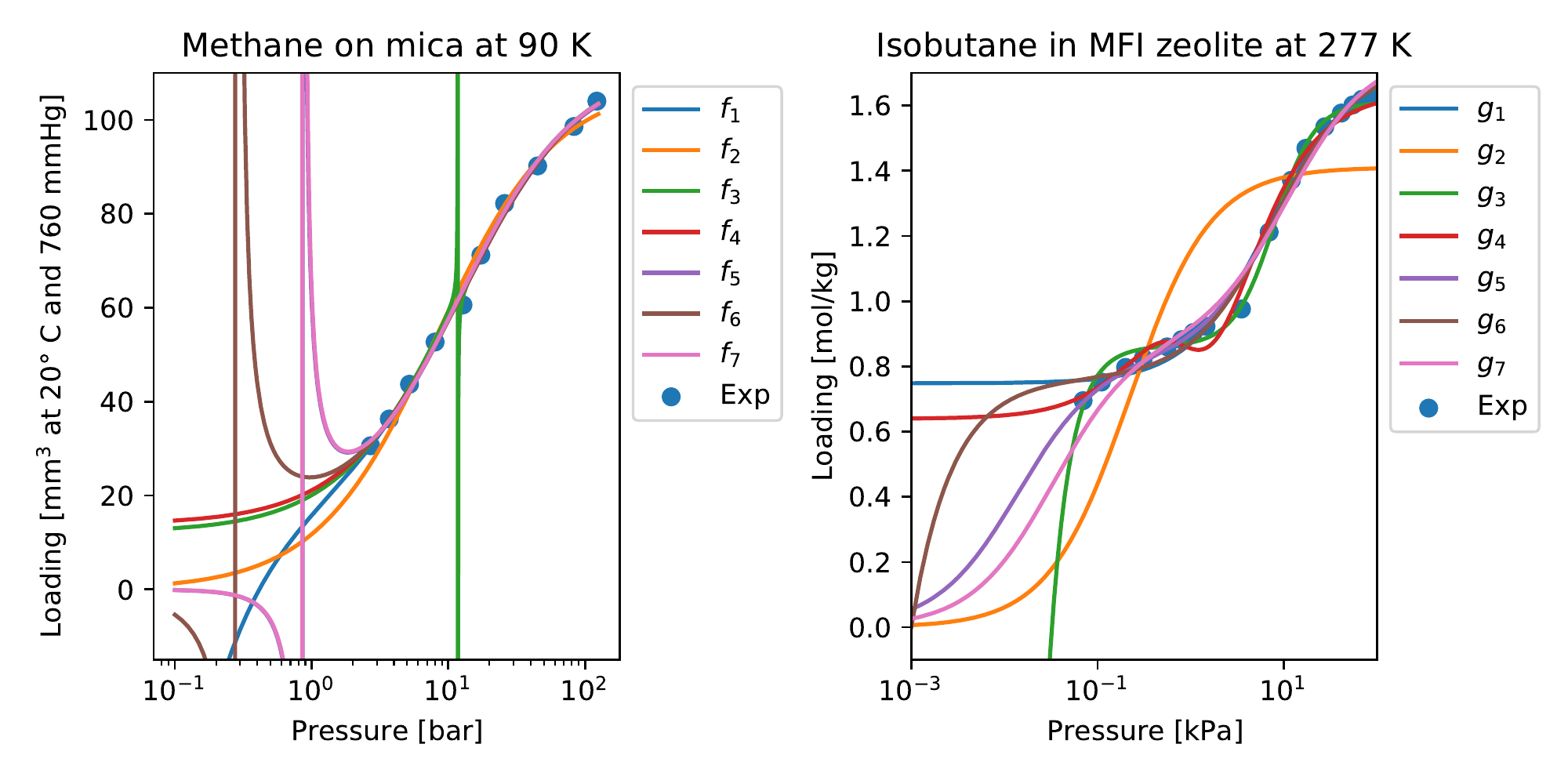}
\caption{Symbolic regression solutions to two adsorption datasets. $f_2$ and $g_2$ are equivalent to the single-site Langmuir equation; $g_5$ and $g_7$ are equivalent to the two-site Langmuir equation.}
\label{fig:circle}
\end{figure}

In Table~\ref{tab:langmuir} 
column~1 gives the data source, 
and column 2 gives the ``hyperparameters" used in our SR experiments: we allow either two or four constants in the derived expressions. 
Furthermore, as the first constraint C1 from $\mathcal K$ %
can be modeled by simply adding the data point $p=q = 0$, %
we also experiment with an ``extra point.''\footnote{We used the approximation $p = q = 0.001$ as our SR solver deals with expressions of the form $p^t$ which is not always defined for $p = 0, t<0$.}
Column 3 displays a derived expression, 
while the columns~4 and 5 give, respectively, the relative numerical errors $\varepsilon_2^r$ and $\varepsilon_{\infty}^r$. 
If the expression can be derived from our background theory, then we indicate that in the column~6. 
These results are visualized in Figure~\ref{fig:circle}.
Column~7 indicates the number of constraints from %
 $\mathcal K$ that each expression satisfies, verified by Mathematica. 
Among the top two-constant expressions,
$f_1$ fits the data better than $f_2$, which is derivable from the background theory, whereas $f_1$ is not. 

When we search for four-constant expressions, we get much smaller errors than (\ref{eq:Langmuir}) or even (\ref{eq:Langmuir2site}) \cite{Langmuir1918}, %
but we do not obtain the two-site formula (\ref{eq:Langmuir2site}) as a candidate expression. 
For the dataset from Sun et al. \cite{sunetal1998}, $g_2$ has a form equivalent to Langmuir's one-site formula, and $g_5$ and $g_7$ have forms equivalent to Langmuir's two-site formula\footnote{This equivalence was not verified by KeYmaera within the time limit, as shown in column 6 in Table~\ref{tab:langmuir}. Further discussion is in the supplementary material.}, with appropriate values of $q_{\mathrm{max},i}$ and $K_{a,i}$ for $i=1, 2$.

\section*{System limitations and future improvements}
Our results on three problems and associated data are %
encouraging and provide the foundations of a new approach to automated scientific discovery.
However our work is only a first, although crucial, step towards completing the missing links in automating the scientific method.

One limitation of the reasoning component is the assumption of correctness and completeness of the background theory. The incompleteness could be partially solved by the introduction of abductive reasoning~\cite{abductive_reasoning} (as depicted in Figure 3 of the main paper). 
Abduction is a logic technique that aims to find explanations of an (or a set of) observation, given a logical theory. The explanation axioms are produced in a way that satisfy the following: 1) the explanation axioms are consistent with the original logical theory and 2) the observation can be deduced by the new enhanced theory (the original logical theory combined with the explanation axioms). In our context the logical theory corresponds to the set of background knowledge axioms that describe a scientific phenomenon, the observation is one of the formulas extracted from the numerical data and the explanations are the missing axioms in the incomplete background theory.

However the availability of background theory axioms in machine readable format for physics and other natural sciences is currently limited. Acquiring axioms could potentially be automated (or partially automated) using knowledge extraction techniques. Extraction from technical books or articles that describe a natural science phenomenon can be done by, for example, deep learning methods (e.g. the work of Pfahler and Morik~\cite{Semantic_Search_Millions_Equations}, Alexeeva et al.~\cite{MathAlign}, Wang and Liu~\cite{Wang2021TranslatingMF}) both from NL plain text or semi-structured text such as LateX or HTML. Despite the recent advancements in this research field, the quality of the existing tools remains quite inadequate with respect to the scope of our system.

Another limitation of our system, that heavily depends on the tools used, is the scaling behavior.
Excessive computational complexity is a major challenge for automated theorem provers (ATPs): for certain types of logic (including the one that we use), proving a conjecture is undecidable. Deriving models from a logical theory using formal reasoning tools is even more difficult when using complex arithmetic and calculus operators. Moreover, the run-time variance of a theorem prover is very large: the system can at times solve some ``large'' problems while having difficulties with some smaller problems.
Recent developments in the neuro-symbolic area use deep-learning techniques to enhance standard theorem provers (e.g., see Crouse et al.~\cite{TRAIL}). We are still at the early stages of this research and there is a lot to be done. We envision that the performance and capability (in terms of speed and expressivity) of theorem provers will improve with time.
Symbolic regression tools, including the one based on solving mixed-integer nonlinear programs (MINLP) that we developed, often take an excessive amount of time to explore the space of possible symbolic expressions and find one that has low error and expression complexity, especially with noisy data. In practice, the worst-case solution time for MINLP solvers (including BARON) grows exponentially with input data encoding size. See the supplementary material for additional details. However, MINLP solver performance and genetic programming based symbolic regression solvers are active areas of research.

Our proposed system could benefit from other improvements in individual components (especially in the functionality available).
For example, Keymaera only supports differential equations in time and not in other variables and does not support higher order logic; BARON cannot handle differential equations.

Beyond improving individual components, our system can be improved by introducing techniques such as experimental design (not described in this work but envisioned in Figure 3 of the main paper\cite{ken_experimental}).
A fundamental question in the holistic view of the discovery process is what data should be collected to give us maximum information regarding the underlying model? The goal of optimal experimental design (OED) is to find an optimal sequence of data acquisition steps such that the uncertainty associated with the inferred parameters, or some predicted quantity derived from them, is minimized with respect to a statistical or information theoretic criterion. In many realistic settings, experimentation may be restricted or costly, providing limited support for any given hypothesis as to the underlying functional form. It is therefore  critical at times to incorporate an effective OED framework. 
In the context of model discovery, a large body of work addresses the question of experimental design for predetermined functional forms, and another body of research addresses the selection of a model (functional form) out of a set of candidates. A framework that can deal with both the functional form and the continuous set of parameters that define the model behavior is obviously desirable \cite{horesh2021experimental}; one that consistently accounts for logical derivability or knowledge-oriented considerations \cite{haber2008numerical} would be even better.

\clearpage
\noindent
{\huge \bf Supplementary Material}

\section*{Datasets}

\subsection*{Kepler's third law of planetary motion}
\begin{table*}[ht]
\centering
\begin{tabular}{lcclll}  
\toprule
  &  & & \multicolumn{3}{c}{normalization factors} \\
\cmidrule{4-6}
 & Original  & & Solar & Exoplanet & Binary Stars \\
\midrule
$p$     & [$s$]      & & $ 1000 \cdot 24 \cdot 60 \cdot 60$   & $ 1000 \cdot 24 \cdot 60 \cdot 60$& $ 365 \cdot 24 \cdot 60 \cdot 60$ [y]\\

$m_1$   & [$kg$]     & & $ 1.9885 \cdot 10^{30}$    & $ 1.9885 \cdot 10^{30}$  & $1.9885 \cdot 10^{30}$\\

$m_2$   & [$kg$]     & & $ 5.972 \cdot 10^{24}$    & $1.898 \cdot 10^{27}$ & $1.9885 \cdot 10^{30}$\\

$d$     & [$m$]     & & $ 1.496 \cdot 10^{11}$  [$au$] & $ 1.496 \cdot 10^{11}$ [$au$] & $1.496 \cdot 10^{11}$ [$au$]\\
\bottomrule
\end{tabular}
\caption{Units of measurement and normalization factors for Kepler data}
\label{tab:refactoring}
\end{table*}

We use three different datasets, provided in Table~\ref{tab:keplerdata}. 
The first has eight data points corresponding to eight planets of the solar system. 
The second has 20 data points consisting of all eight data points from the first dataset, and, in addition, data points corresponding to some exoplanets in the Trappist-1 and the GJ 667 systems.
The third consists of data for five binary stars \cite{binary_star}.
All three datasets consist of real measurements of four variables: 
the distance $d$ between two bodies, a star and an orbiting planet in the first two datasets and binary stars in the third dataset,
the masses $m_1$ and $m_2$ of the two bodies, 
and the orbital period $p$, which is our target variable.
We normalized the data (e.g., masses of planets and stars) to reduce errors that can arise due to processing large numbers in our system.
Table~\ref{tab:refactoring} gives the original unit of measurement and the normalization factors for each dataset. 
Each star mass is given as a multiple of the mass of the sun, hence the sun mass equals 1.
In the first dataset, each planetary mass is given as a multiple of Earth's mass, whereas in the second dataset each planetary mass is given relative to Jupiter's mass.
The distance $d$ is given in astronomical units [au].
The period $p$ is given as days$/1000$ or years.

\begin{table*}[ht]
\centering
\resizebox{\textwidth}{!}{%
\begin{tabular}{cccccccccccccc}  
\toprule
\multicolumn{4}{c}{Solar} && \multicolumn{4}{c}{Exoplanet} && \multicolumn{4}{c}{Exoplanets (contd.)} \\
\cmidrule{1-4} \cmidrule{6-9} \cmidrule{11-14}
$m_1$ & $m_2$ & d & t && $m_1$ & $m_2$ & d & t && $m_1$ & $m_2$ & d & t \\
\cmidrule{1-4} \cmidrule{6-9} \cmidrule{11-14}
1.0 &   0.0553 &  0.3870 &  0.0880    && 1.0  & 0.000174 &  0.3870 &  0.0880    && 0.08 & 0.0043 &  0.0152 &  0.0024218  \\
1.0 &   0.815  &  0.7233 &  0.2247    && 1.0  & 0.00256  &  0.7233 &  0.2247    && 0.08 & 0.0013 &  0.0214 &  0.0040496  \\
1.0 &   1.0    &  1.0    &  0.3652    && 1.0  & 0.00315  &  1.0    &  0.3652    && 0.08 & 0.002  &  0.0282 &  0.0060996  \\
1.0 &   0.107  &  1.5234 &  0.6870    && 1.0  & 0.000338 &  1.5234 &  0.6870    && 0.08 & 0.0021 &  0.0371 &  0.0092067  \\
1.0 & 317.83   &  5.2045 &  4.331     && 1.0  & 1.0      &  5.2045 &  4.331     && 0.08 & 0.0042 &  0.0451 &  0.0123529  \\
1.0 &  95.16   &  9.5822 & 10.747     && 1.0  & 0.299    &  9.5822 & 10.747     && 0.08 & 0.086  &  0.063  &  0.018767   \\
1.0 &  14.54   & 19.2012 & 30.589     && 1.0  & 0.0457   & 19.2012 & 30.589     &&      &      &         &        \\
1.0 &  17.15   & 30.0475 & 59.800     && 1.0  & 0.0540   & 30.0475 & 59.800     &&      &      &         &        \\
    &         &         &             && 0.33 & 0.018  &  0.0505 &  0.0072004 &&      &      &         &        \\
\multicolumn{4}{c}{Binary stars}      && 0.33 & 0.012  &  0.125  &  0.02814   &&      &      &         &        \\
\cmidrule{1-4}
0.54 & 0.50 & 107.270 & 1089.0        && 0.33 & 0.008  &  0.213  &  0.06224   &&      &      &         &        \\
1.33 & 1.41 &  38.235 &  143.1        && 0.33 & 0.008  &  0.156  &  0.039026  &&      &      &         &        \\
0.88 & 0.82 & 113.769 &  930.0        && 0.33 & 0.014  &  0.549  &  0.2562    &&      &      &         &        \\
3.06 & 1.97 & 131.352 &  675.5        && 0.08 & 0.0027 &  0.0111 &  0.0015109 &&      &      &         &        \\
\bottomrule
\end{tabular}
}
\caption{Kepler data}
\label{tab:keplerdata}
\end{table*}

\subsection*{Relativistic time dilation}

\begin{table*}[ht]
\centering
\begin{tabular}{rr}  
\toprule
Velocity (m/s)  & Time dilation ($10^{-15}$)\\
\midrule
0.55 & -0.018 \\
4.10 & -0.21 \\
8.60 &  -0.43 \\
14.84 & -1.54 \\
22.18 & -2.92 \\
29.65 & -4.82 \\
36.22 & -7.36 \\
\bottomrule
\end{tabular}
\caption{Time dilation data}
\label{tab:timedilationdata}
\end{table*}

We report in Table~\ref{tab:timedilationdata} the data used in \cite[Figure 2]{chou2010}. The first column gives the velocity of a moving clock relative to another stationary clock, and the second column gives the relative change in clock rates (i.e., it gives the clock rate (or frequency) of the moving clock minus the clock rate of the stationary clock divided by the clock rate of the stationary clock) scaled by $10^{15}$.

\begin{figure}
\centering
\includegraphics[width=5in]{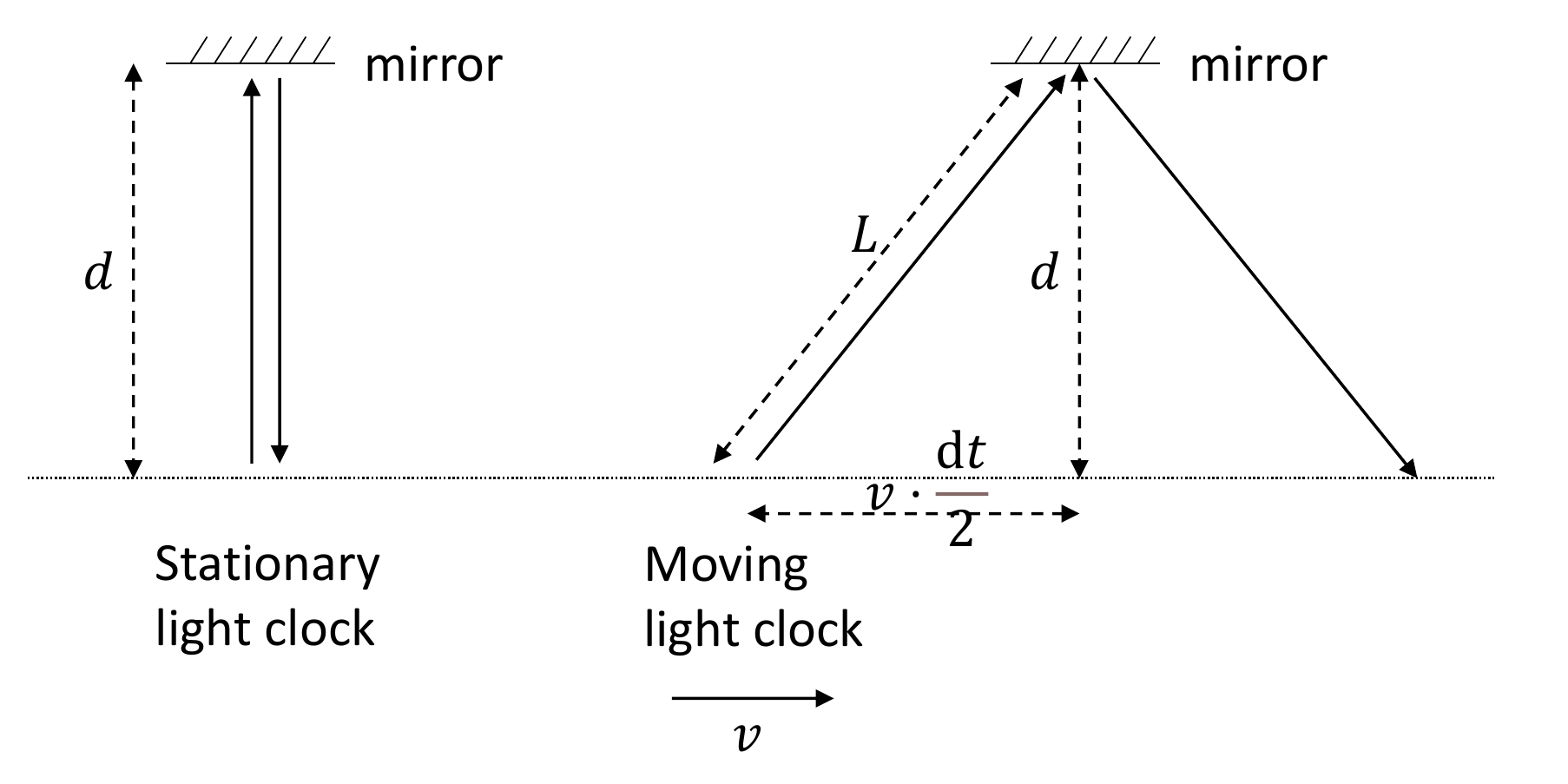}
\caption{Depiction of moving light clock}\label{fig:light-clock}
\end{figure}

We next list and describe the axioms given as input to the reasoning module for this problem (see Table~\ref{tab:relativity}). The period $dt_0$ of a ``light clock'' is defined as the time for light (at velocity $c$) to travel between two stationary mirrors separated by distance $d$ (axiom R1 below). The period $dt$ of a similar pair of mirrors moving with velocity $v$ (axiom R2), is the time taken for light to bounce between the two mirrors, but in this case, while traveling the distance $L$ (calculated via the Pythagorean theorem in axiom R3). The observed change in clock frequency due to motion, $d\freq = \freq - {\freq}_0$, (axiom R6) is related to periods $dt_0$ and $dt$ using definitions of frequency (axioms A4 and R5). The second column of the previous table gives values for $d\freq/{\freq}_0$ (after scaling by $10^{15}$). Here all variables are positive, and the speed of light is taken to be $3 \times 10^8$ meters per second (axioms R7-R8). In Figure~\ref{fig:light-clock}, the dashed lines represent lengths, the solid lines represent the direction of travel for light (if vertical or diagonal), or the direction of motion of the light source (horizontal).\\

\begin{table}
\begin{tabular}{llllllll}
& R1. & $dt_0 = 2\cdot d/c$ &~~&&  & R1'. & $dt_0 = 2\cdot d/c$  \\[4pt]
& R2. & $dt = 2\cdot L/c$   &~~&&  & R2'. & $dt = 2\cdot L/\sqrt{v^2+c^2}$ \\[4pt]     
& R3. & $L^2 = d^2 + (v\cdot dt/2)^2$ &~~&& & R3. & $L^2 = d^2 + (v\cdot dt/2)^2$ \\[4pt]
& R4. & $\freq_0 = 1/dt_0$      &~~&&   & R4'. & $\freq_0 = 1/dt_0$ \\[4pt]
& R5. & $\freq = 1/dt$          &~~&&   & R5'. & $\freq = 1/dt$\\[4pt]
& R6. & $d\freq = \freq - \freq_0$      &~~&&   & R6'. & $d\freq = \freq - \freq_0$ \\[4pt]
& R7. & $d > 0, v > 0$      &~~&&   & R7'. & $d > 0, v > 0$\\[4pt]
& R8. & $c = 3\times 10^8~$ &~~&&   & R8'. & $c = 3\times 10^8~$\\
\end{tabular}
\caption{Axioms for relativistic time dilation are given in the first two columns, and an alternate set of axioms for ``Newtonian behavior'' in columns three and four.}
\label{tab:relativity}
\end{table}

\subsection*{Langmuir's adsorption equation}

\subsubsection*{Data}

In Table~\ref{tab:langmuirdata} we provide two datasets, one taken from Langmuir's original paper \cite[Table IX]{Langmuir1918}, and the other from \cite[Table 1]{sunetal1998}. Each dataset gives the measured loading $q$ at different values of  pressure $p$ at a fixed temperature. 
We note that different scales for pressure and loading are used in these two datasets. 

\begin{table*}[ht]
\centering
\begin{tabular}{ccccccc}  
\toprule
\multicolumn{2}{c}{Langmuir \cite[Table IX]{Langmuir1918}} && \multicolumn{4}{c}{Sun et al. \cite[Table 1]{sunetal1998}} \\
\cmidrule{1-2}  \cmidrule{4-7}
$p$ & $q$  && $p$ & $q$ & $p$ & $q$\\
\cmidrule{1-2}  \cmidrule{4-7}
  2.7 & 30.6 &&  0.07 & 0.695 & 12.06 & 1.371\\
  3.7 & 36.3 &&  0.11 & 0.752 & 17.26 & 1.469 \\
  5.2 & 43.7 &&  0.20 & 0.797 & 27.56 & 1.535 \\
  8.0 & 52.7 &&  0.31 & 0.825 & 41.42 & 1.577 \\
 12.8 & 60.6 &&  0.56 & 0.860 & 55.20 & 1.602 \\
 17.3 & 71.2 &&  0.80 & 0.882 & 68.95 & 1.619 \\
 25.8 & 82.2 &&  1.07 & 0.904 & 86.17 & 1.632 \\
 45.0 & 90.2 &&  1.46 & 0.923 \\
 83.0 & 98.6 &&  3.51 & 0.976 \\
122.0 &104.0 &&  6.96 & 1.212 \\
\bottomrule
\end{tabular}
\caption{Langmuir data}
\label{tab:langmuirdata}
\end{table*}

\subsubsection*{Thermodynamic constraints}

A different form of background knowledge is also available for adsorption thermodynamics; axioms for single-component adsorption are more plausible when they satisfy certain thermodynamic constraints $\mathcal{K}$:

\parbox{2.5in}{
\begin{tabular}{llll}
& C1. & $f(0)=0$            \\[4pt]
& C2. & $(\forall p>0 )~(f(p)>0)$    \\[4pt]     
& C3. & $(\forall p>0 )~(f'(p)\geq 0 )$ \\[4pt]
& C4. & $ 0 < \lim_{p\rightarrow 0} f'(p) < \infty$ \\[4pt]
& C5. & $ 0 < \lim_{p\rightarrow \infty} f(p) < \infty$   \\
\end{tabular}
}

C1 requires that at zero pressure, zero molecules may be adsorbed, and C2 requires that only positive loadings are feasible. C3 requires the isotherm increase monotonically with pressure, which holds for all single-component adsorption systems. C4 requires the slope of the adsorption isotherm in the limit of zero pressure (the adsorption second virial coefficient) to be positive and finite \cite{thermodynamic_treatment}. C5 requires the loading to be finite in the limit of infinite pressure, thus imposing a saturation capacity for the material. These five constraints are satisfied by Langmuir and multi-site Langmuir models, but some popular models in the literature violate these to various degrees. For example, Freundlich and Sips formulae violate constraint C4 -- a well-known issue critiqued by \cite{thermodynamic_treatment}. The BET isotherm violates C2 and C3 because of its singularity at the vapor pressure of the fluid $p_{\mathrm{vap}}$; this could be resolved by instead enforcing positivity and monotonicity from $0 < p < p_{\mathrm{vap}}$.

\section*{Additional related work}

Symbolic discovery of formulae is a well studied research field, and it is a recognised challenge for the entire Artificial Intelligence community \cite{Kitano_2016}.

We addressed the most relevant literature in the introduction to the paper.
Some other works worth mentioning are at the intersection of symbolic discovery and deep learning. Several attempts have been made to use deep neural network for learning or discovering %
symbolic formulae \cite{Extrapolation_learning_equations,deep-symbol-19,univ-approx,symb-and-neural,tensor-net,NN_discovery,dicovery_symb_bias}. 
For example, in the work by Iten et al.~\cite{NN_discovery}
the authors focused on modelling a neural network architecture on the human physical reasoning process. 
In the work by Arabshahi et al.~\cite{symb-and-neural} the authors use tree LSTMs to incorporate the structure of the symbolic expression trees, and combine symbolic reasoning and function evaluation for solving the tasks of formula verification and formula completion.
Symbolic regression has been used~\cite{dicovery_symb_bias} to extract explicit physical relations from components of the learned model of a Graph Neural Network (GNN).
In a work by Derner et al.~\cite{symb-reg-analytic}, the authors employ symbolic regression to construct parsimonious process models described by analytic formulae for real-time RL control when dealing with unknown or time-varying dynamics.
In AI-Feynman~\cite{AI-Feynman} (and the subsequent paper AI-Feynman 2.0 \cite{AI-Feynman2.0}), the authors introduce a recursive multidimensional symbolic regression algorithm that combines neural network fitting with a suite of physics-inspired techniques
to find an formula that matches data from an unknown function.
In the work by Jin et al.~\cite{bayesianSR}, symbolic regression is combined with Bayesian models.
Moreover, neuro-symbolic systems have received a lot of attention in the past years: see the papers on combining mathematical reasoning with deep neural networks~\cite{math-latent,ai-coper}. Bayesian program synthesis has also been successful for learning a library of formulae~\cite{DreamCoder}.
Other types of search have been investigated as well, e.g. Bayesian Markov chain Monte Carlo search \cite{bayesian_machine}, graph-based search \cite{reis}, or methods for identifying linear combinations of nonlinear descriptors for dynamic systems (SINDy) \cite{Brunton3932} or material property prediction (SISSO) \cite{SISSO}.
However, to the best of our knowledge, these methods have never been combined with logical reasoning. 

The use of logic constraints to help with the discovery of formulas from numerical data is a well known technique. 

\new{
The LGML tool of Scott et al. 2021~\cite{scott2021lgml}, 
is a pipeline involving a learning component (either a symbolic regression tool or a deep neural network) and a consistency checking component. The consistency checking is performed against a set of equities or inequalities describing the functional form that they are trying to learn: ``logical equation in terms of the feature space and the unknown function''.
The LGGA tool of Ashok et al.~\cite{ashok2020logic} is an extension of LGML and is a genetic algorithm enhanced by auxiliary truth in the form of ``mathematical expressions that capture domain specific knowledge or simple properties of an unknown function''.
The main goal of the consistency checking component of these two works is to understand if a set of constraints are satisfied by a given function $f'$ (the approximation of the correct function $f$). They thus check if $f’\models C$.
Our work differentiate from these two methods as follows: 
1) the type of constraints used is fundamentally different. We use general scientific laws describing the environment, often not including any of the variables present in the data, while in LGML and LGGA they only consider simple constraints on the functional form
2) the reasoning task we are trying to solve is different. As mentioned above LGGA and LGML solve the logic task of $f’\models C$ where $f'$ is an approximation of the function $f$ and $C$ is a single constraint on $f$ functional form. AI-Descartes solve the task of $B\models f’$, where $B$ is a collection of axioms and $f'$ is an approximation of the function $f$. Moreover we solve other reasoning tasks such as the computation of the reasoning errors etc.
A similar, however earlier, approach is the one by Bladek and Krawiec \cite{bladek19} 
where the authors formalize the task of symbolic regression with formal constraints by the generation of counterexamples.
Another work that combines SR and prior knowledge is the work of Kubalik et al.~\cite{kubalik2020symbolic,kubalik2021multi}, where the authors consider a set of nonlinear inequality and equality constraints on the functional form of the function to discover. They add this prior knowledge in their multi-objective symbolic regression approach as a set of discrete additional data samples on which candidate models are exactly checked. In this way consistency check can be incorporated into the fitness evaluation or as an additional optimization objective.
Finally, the work of Engle and Sahinidis~\cite{engle2021deterministic} introduce a novel deterministic mixed-integer nonlinear programming formulation for symbolic regression that uses derivative constraints through auxiliary expression trees.
}

\new{
Using logic constraints as part of ML tools increased in popularity in recent years in the Neuro-Symbolic field: e.g. using the violation of logic constraints as part of the loss of an NN\cite{xu2018semantic,wang2020integrating}, or in other techniques~\cite{li2019augmenting,daniele2020neural,xie2019embedding,li2019logic,survey_vonRueden,survey_dash,survey_eleonora}.
However all these methods are based only on constraints describing the functional form that has to be learnt, and do not incorporate background-theory axioms (logic constraints that describe the other laws and variables that are involved in the phenomenon).
Another Neuro-Symbolic research area that is relevant to our work is the extraction of logical rules from data (also called ILP or rule induction). Some examples are the work of Sen et al.~\cite{Riegel_lnn} or Evans and Grefenstette~\cite{evans2018learning} among many others~\cite{NEURIPS2019_0c72cb7e,law2018inductive}.
A related topic with a lot of attention is program synthesis. Recently there have been a lot of discussion on this topic in the context of Neuro-symbolic systems. Some example are the works of Nye et al.~\cite{NEURIPS2020_7a685d9e}, Parisotto et al.~\cite{parisotto2016neuro}, Valkov et al.~\cite{valkov2018houdini} or Yang et al.~\cite{yang2017differentiable}.
}

\section*{Symbolic regression}
\begin{figure}
\begin{center}
\includegraphics[width=0.6\columnwidth]{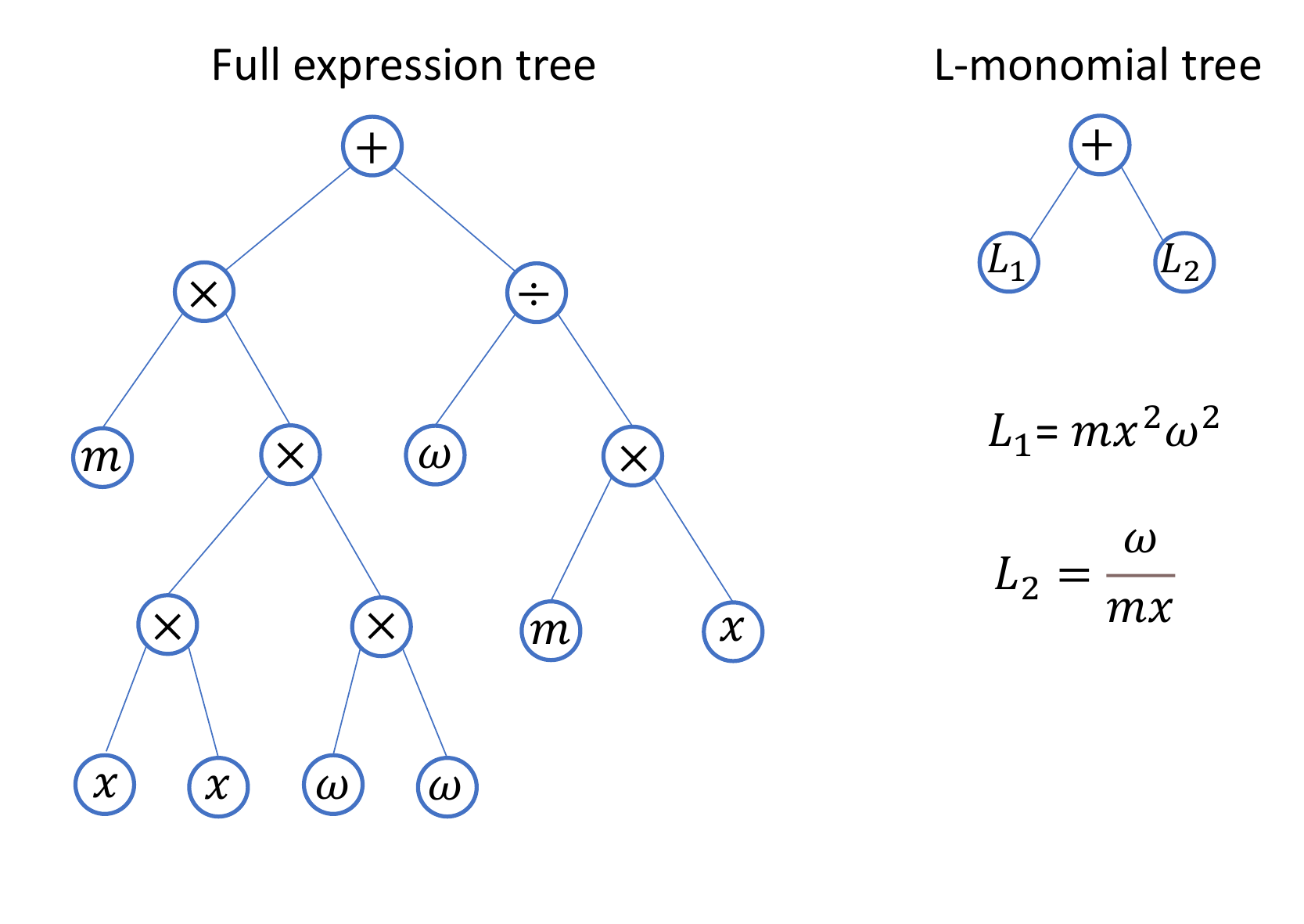}
\caption{Expression tree for $mx^2\omega^2 + \frac{\omega}{mx}$, expressed with full arithmetic notation, and as a more compact tree of L-monomials.}
\label{fig:expression_tree}
\end{center}
\end{figure}

A symbolic regression scheme consists of a space of valid mathematical expressions composable from a basic list of operators (we assume these to be unary or binary), and a mechanism for exploring the space.  
Each valid mathematical expression can be represented by an expression tree, i.e., a rooted binary tree where each non-leaf node has an associated binary or unary operator ($+$, $-$, $\times$, $\sqrt{}$, $\log$, etc.), and each leaf node has an associated constant or independent variable. 
An example of an expression tree for the expression
$$mx^2\omega^2 + \frac{\omega}{mx}$$
is presented in Figure~\ref{fig:expression_tree} (full expression tree).

Symbolic regression is often solved with genetic programming (GP). 
The Eureqa package \cite{schmidt14}, based on the article \cite{schmidt2009distil}, is a state-of-the-art GP-based solver. 
Another popular solver is {\tt gplearn} \cite{gplearn}.
Such solvers search the space of expressions using genetic algorithms.
Models generated
by GP often suffer from poor accuracy \cite{korns} and lengthy descriptions.

\newcommand{\Z}{\mathbb{Z}}
\newcommand{\R}{\mathbb{R}}

The symbolic regression problem can be formulated in various ways as a Mixed-Integer Nonlinear-Programming (MINLP) problem, see \cite{cozadthesis,horesh2015globally, Austel2017,cozsah}. 
The MINLP problem is solved to global optimality using an off-the-shelf MINLP solver such as BARON \cite{baron96}, COUENNE  or SCIP \cite{MaherFischerGallyetal2017}.
These solvers solve problems of the form
\begin{eqnarray}
\mathop{\rm Minimize}  & f(\mathbf{x},\mathbf{y}) \\
\mbox{\rm s.t.} & g(\mathbf{x},\mathbf{y}) \leq \mathbf{b} \\
& \mathbf{x} \in \R^m, \mathbf{y} \in \Z^m
\end{eqnarray}
where $\mathbf{x}$ is a vector of $m$ continuous variables, $\mathbf{y}$ is a vector of $n$ discrete variables, $f(\mathbf{x},\mathbf{y})$ is a real-valued function that can be composed using a finite list of operators such as $+, -, \times, /, exp(\cdot)$ etc. (this list will vary with each solver), and $g(\mathbf{x},\mathbf{y})$ is a vector-valued function created using the same operators.
These solvers
use various convex-relaxation schemes to obtain lower bounds on the objective-function value, and utilize these bounds in branch-and-bound schemes to obtain globally optimal solutions.

This approach produces a globally-optimal mathematical expression along with a certificate of optimality, while avoiding an exhaustive search of the solution space. 
Another advantage is that it directly produces correct, within a tolerance, real-valued constants; most other methods use specialized algorithms to refine constants \cite{dgp} and cannot guarantee global optimality.

In these MINLPs, the set of valid binary expression trees is specified by a set of constraints over discrete and continuous variables. The discrete variables of the formulation are used to define the structure of the expression tree including the assignment of operators to non-leaf nodes and whether a leaf node is assigned a constant or a specific independent variable. The continuous variables are used for the undetermined constants, and also to evaluate the resulting symbolic expression for specific numerical values associated with individual data points.
The objective functions vary from {\em accuracy} (measured as sum of squared deviations) to {\em model complexity}.
The MINLP formulations broadly have the following form:
\begin{eqnarray*}
\mathop{\rm Minimize}
~~~~~~ &  \mathcal{C}(f_{\theta}) & \mbox{complexity}\\ 
\mbox{s.t.}~~~~~~~ & \theta \in \mathcal{T} &  \mbox{grammar}\\ 
& \mathbf{v}_i = f_{\theta}(\mathbf{X}^{(i)}), ~~ \forall i \in I  & \mbox{prediction}\\
& \mathcal{D}(f_{\theta}(\mathbf{X}),\mathbf{Y}) \leq \varepsilon & \mbox{error}
\end{eqnarray*}
where $f_{\theta}$ represents an expression tree defined by the structural and continuous decision variables collectively designated as $\theta$; $\mathcal{T}$ is the universe of valid expression trees; $\mathcal{C}$ measures the description complexity; $\mathcal{D}$ measures error of the predicted values $\mathbf{v}$; $\mathbf{Y}$ is the vector of observations for input vectors $\{\mathbf{X}^{(i)}\}_{i \in I}$.

\subsection*{System Description}
We implemented a symbolic regression system based on a novel mixed-integer nonlinear programming formulation.
We first describe the basics of our system (without dimensional analysis), and then later explain how we incorporate dimensional analysis.

We take as input a list of operators (for this discussion, assume the input operators are $+, -, \times,/$, and $\sqrt{}$), an upper bound $d$ on the tree depth, an upper bound $k$ on the number of constants, and a domain for the constants $[-\Omega , \Omega]$. In Figure~\ref{fig:expression_tree}, there is a single constant (distinct from 1) with value $1/4$ in the expression tree and in the L-monomial tree.
In the prior work on globally optimal symbolic regression, the MINLP formulations typically have discrete variables that (1) determine the placement of the operators in nodes of the expression tree and (2) the mapping of independent variables to leaf nodes, and (3) determine whether to map an independent variable or a constant to a leaf node.

In our formulation, we simply do not have discrete variables for (1). We explicitly enumerate all possible assignments of the operators in expression trees up to depth $d$. More precisely, if a ``partial'' expression tree is one where the leaf nodes are undetermined, but the operator assignments to non-leaf nodes are determined, then we enumerate all possible partial expression trees up to a certain depth.
Our assignment of variables/constants to leaf nodes is also different from prior work. Let $x_1, \ldots, x_n$ be all the independent variables. Instead of assuming that each leaf node is either a constant $h$ or one of $x_1, \ldots, x_n$ as in the prior work, we assume each leaf node is a one-term multivariable Laurent polynomial of the form
\begin{equation}\label{lmonomial} hx_1^{a_1}x_2^{a_2}\cdots x_n^{a_n}, \end{equation}
where $a_1, \ldots, a_n$ are (undetermined) integers (for computational efficiency, we limit these integers to lie in the range $[-\delta, \delta]$ for some input constant $\delta$), and $h$ represents an (undetermined) constant in the final expression.
We refer to an expression of the type (\ref{lmonomial}) as an {\it L-monomial}.
In other words, rather than assigning a single variable or constant to a leaf node, we potentially assign both variables and constants, and also multiple variables (with positive or negative powers to a leaf node).  We call the resulting trees {\em generalized expression trees}, and {\em gentrees} for convenience.

Each gentree $T$ (with depth say $d$) corresponds to a symbolic expression: each non-leaf node with height $1$ corresponds to the expression formed by applying the operator at the node to the expressions (i.e., L-monomials) in the children nodes, and non-leaf nodes at greater heights are handled in the same manner in order of height. The only gentree with depth 0 corresponds to the L-monomial $L_1$, whereas the gentrees of depth 1 correspond to the expressions $\sqrt{L_1}$, $L_1+L_2$,\, $L_1 \times L_2$,\, $L_1/L_2$ and $L_1-L_2$, respectively, where $L_1$ and $L_2$ are L-monomials.

\subsection*{Pruning the list of gentrees}

We try to reduce the number of relevant partial expression trees/gentrees by removing a number of "redundant" trees using the fact that
the set of nonzero (the constant $h$ in \ref{lmonomial} is nonzero) L-monomials is closed under multiplication and division. 
Notice that both $L_1\times L_2$ and $L_1/L_2$ are L-monomials and thus can be represented by a single expression $L_1$. In other words, if there is a symbolic expression $f$ of the form $f = L_1\times L_2$ that fits our data, then there is one of the form $f = L_1$.
Accordingly our first few pruning rules are: remove a tree $T$ from the list $\mathcal{T}$ of all gentrees with depth up to $d$ if $T$ has the subexpression
\begin{eqnarray*}
    & \mbox{[R1]}  ~~~& L_1 \times L_2 \mbox{ or } L_1/L_2.\\ 
    & \mbox{[R2a]} ~~~& L_1\times (L_3 \pm L_4).\\
    & \mbox{[R2b]} ~~~& (L_1 \pm L_2)*(L_3 \pm L_4). \\
    & \mbox{[R3]}  ~~~& (L_1 \pm L_2)/L_3.
\end{eqnarray*}
The first rule was explained above. The second follows by associativity: $L_1*(L_3 + L4) = L_1\times L_3 + L1\times L_4 = L_1' + L_2'$, for some L-monomials $L_1'$ and $L_2'$. If a $T$ with a subexpression $L_1*(L_3 + L4)$ appears in $\mathcal{T}$, then replacing this subexpression by $L_1' + L_2'$ results in another gentree $T'$ of the same depth, which must therefore be in $\mathcal{T}'$.
The same idea can be applied to the subexpression $L_1*(L_3 - L_4)$.
We can apply associativity twice to justify rule R2b, as $(L_1 + L_2)*(L_3+ L_4) = L_1' + L_2' + L_3' + L_4'$, for some $L_i'$ (the same argument holds when either $+$ is replied by a $-$ in R2b).
Once again, the second expression has the same depth as the first, and therefore there must be a tree in $\mathcal{T}$ containing it. R3 can be explained similarly.

\subsection*{MINLP formulation for a gentree}

Let the data points be $\mathbf{X}^{(i)}$ for $i \in I$, where $I$ is an index set, and let the features/independent variables be $x_1, \ldots, x_n$.
Let $\mathbf{Y}$ stand for the observations of the dependent variable $y$, with $Y^{(i)}$ standing for the $i$-th observation (of the dependent variable).
Let $T$ be a given gentree with $m$ leaf nodes, and let $\mathbf{p}$ be the vector of all integer variables corresponding to the powers of the independent variables in the different leaf nodes. Then $\mathbf{p} \in \mathbb{Z}^{mn}$. Let $h_1, \ldots h_m$ be the variables corresponding to the constants in leaf nodes $1, \ldots, m$.
Finally, assume we have a 0-1 variable $z_i$ that determines whether or not the $i$th leaf node has a constant $h_i$ that is different from one.
Thus the (vector) variables in our model are $\mathbf{p}, \mathbf{z}$ and $\mathbf{h}$. Let $f_{\mathbf{h},\mathbf{p}, \mathbf{z}, T}$ stand for the symbolic expression defined by fixing the values of these variables.
Then the MINLP we solve can be framed as:
\begin{align}
\min~~      &~~~~ \sum_{i \in I}(Y^{(i)} - f_{\mathbf{h},\mathbf{p}, \mathbf{z}, T}(\mathbf{X}^{(i}))^2  \label{minlp1}\\ 
\mbox{s.t.}~~&~~~~  -\delta \leq p_i \le \delta ~~~ \mbox{for } i=1, \ldots, mn \label{power1}\\ 
&~~~~ -\Omega z_i + (1-z_i) \leq h_i \leq \Omega z_i + (1-z_i) ~~~\mbox{for } i =1,\ldots, m   \label{const1} \\
&~~~~ \sum_{i=1}^m z_i \leq k  \label{maxconst1}\\
&~~~~ \mathbf{z} \in \{0,1\}^m, ~~\mathbf{p} \in \Z^{mn}  \label{integrality}
\end{align}

The constraints (\ref{integrality}) and (\ref{power1}) force $z_i$ to take on 0-1 values, and $p_i$ to take on values in the range $\{-\delta, -\delta+1, \ldots, \delta\}$. The constraint (\ref{const1}) restricts $h_i$ to lie in the range $[-\Omega, \Omega]$ if $z_i$ has value 1, and forces $h_i$ to take on value 1, when $z_i$ has value 0.
The constraints (\ref{maxconst1}) allow at most $k$ of the $z_i$ variables to have value 1, and therefore at most $k$ of the $h_i$ values to be different from 1.
The function $f_{\mathbf{h},\mathbf{p},\mathbf{z},T}$ is composed of the operators in the non-leaf nodes of $T$ from the L-monomials in the leaf nodes.
The objective function is the sum of squares of differences between the symbolic expression values for each input data point and the corresponding dependent variable value (call it the {\em least-square error}).

We use BARON to solve the MINLPs we generate, and thus $T$ and $f_{\mathbf{h},\mathbf{p},\mathbf{z},T}$ are limited by the operators that BARON can handle ($+,-,\times,/,\exp(),\log()$).
We will illustrate $f$ for a few examples, rather than specify it formally.
Suppose we are trying to derive the formula $F=G\, \frac{m_1 m_2}{r^2}$, where
$m_1, m_2$ and $r$ are independent variables, and $G$ is an unknown constant, and we have multiple data points (for $i \in I$) with values for the independent variables, and for associated $F$.
If $T$ is a depth 0 gentree, then $T$ is just a single L-monomial,
and $f_{\mathbf{h},\mathbf{p},\mathbf{z},T} = hm_{1}^am_{2}^br_i^c$ where $a,b,c$ are undetermined integers in the range $[-\delta,\delta]$, and $h$ is an undetermined real number in the range $[-\Omega, \Omega]$. The objective function is then
$$\sum_{i\in I} (F_i-hm_{1i}^a m_{2i}^b r_i^c )^2~, $$
where 
$F_i, m_{1i}, m_{2i}$ and $r_i$ are the values of $F, m_1, m_2, r$ in the $i$th data point.
If the gentree can be written as $L_1 + L_2$, then the objective function becomes
$$\sum_{i\in I} (F_i -(h_1m_{1i}^{a_1} m_{2i}^{b_1} r_i^{c_1}+ h_2m_{1i}^{a_2} m_{2i}^{b_2} r_i^{c_2}))^2; $$
here $a_i, b_i,c_i$ and $h_i$ are undetermined, and we solve for these values.

As depicted in Figure~\ref{fig:expression_tree}, an expression tree of a certain depth can sometimes be represented by a gentree of smaller depth. Therefore, by enumerating gentrees of a certain depth, one obtains a much richer class of functions than can be obtained by expression trees of the same depth.

\subsection*{Enumeration and parallel processing}

By enumerating the generalized expression trees and solving them separately, the problem is divided into multiple, easier-to-solve sub-problems (operator placement in non-leaf nodes does not have to be determined any more) that can be solved much more quickly. This "divide and conquer" formulation can obtain solutions to problems that were intractable for the formulation in \cite{Austel2017}.

When available, we exploit parallelism, and run multiple threads, with one MINLP corresponding to a single gentree in a single thread.
If we obtain a solution that fits the data within a prescribed tolerance from a gentree with depth $d'$, then we stop all processes/threads containing gentrees with depth $> d'$.  We terminate a gentree if the lower bound on the least-square error exceeds the least-square error found from other gentrees of the same or lower depth. Thus we execute a branch-and-bound type search, and implicitly search for the least depth gentree that fits the data.

If we have more gentrees than available cores, we process them in a round robin fashion; if $t$ is the number of cores, we start solving for $t$ gentrees in parallel, and after a fixed amount of time (10 seconds), we pause the first $t$ gentrees, and start solving $t$ more, till we either find a solution or run out of gentrees in which we case we start from the first gentree. The gentrees are sorted by a measure of complexity (roughly equal to the number of nodes).

\section*{Scaling}
The gentree enumeration running time (and thus that of our overall algorithm) grows exponentially with gentree depth $d'$.
As described earlier, we enumerate a number of gentrees, and solve an MINLP per gentree. The number of distinct gentrees grows exponentially with the depth $d'$ of the gentree and also grows rapidly with the number of operators used; see Table~\ref{tab:numgentrees}.

\begin{table*}[ht]
\centering

\begin{tabular}{crr}
\toprule
 & Number of trees with & Number of trees with \\
 Max tree depth  & operators: $+$, $-$, $\times$, $/$ & operators: $+$, $-$, $\times$, $/$, $\sqrt{}$ \\
\midrule
1 & 2 & 2 \\
2 & 6 & 7 \\
3 & 31 & 60 \\
4 & 1153 & 4485 \\
5 & 1,506,165 & $\ge$ 10,000,000 \\
\bottomrule
\end{tabular}
\caption{Number of gentrees as a function of depth}
\label{tab:numgentrees}
\end{table*}

The number of gentrees for $d'=2, 3, 4, 5$ are, respectively, 7, 60, 4485 and over ten million when we use the operators $+$, $-$, $\times$, $/$, $\sqrt{}$. Our gentree enumeration approach is unlikely to be tractable for $d' \ge 6$ and is already hard for $d'=4$ if we include more operators than listed earlier.
But each L-monomial can represent large expression trees. The L-monomial  $hm_1^am_2^br_i^c$, which corresponds to a depth 0 gentree, needs a depth 3 binary expression tree if the powers $a, b,c$ are integers between $-2$ and $2$. Thus our depth-3 L-monomial trees can easily need expression trees with depth 6 or more to represent them when we have three independent variables. If the number of independent variables in the L-monomial is more than 3, or the magnitude of the powers is increased, then even deeper expression trees would be needed.

\begin{figure}[h]
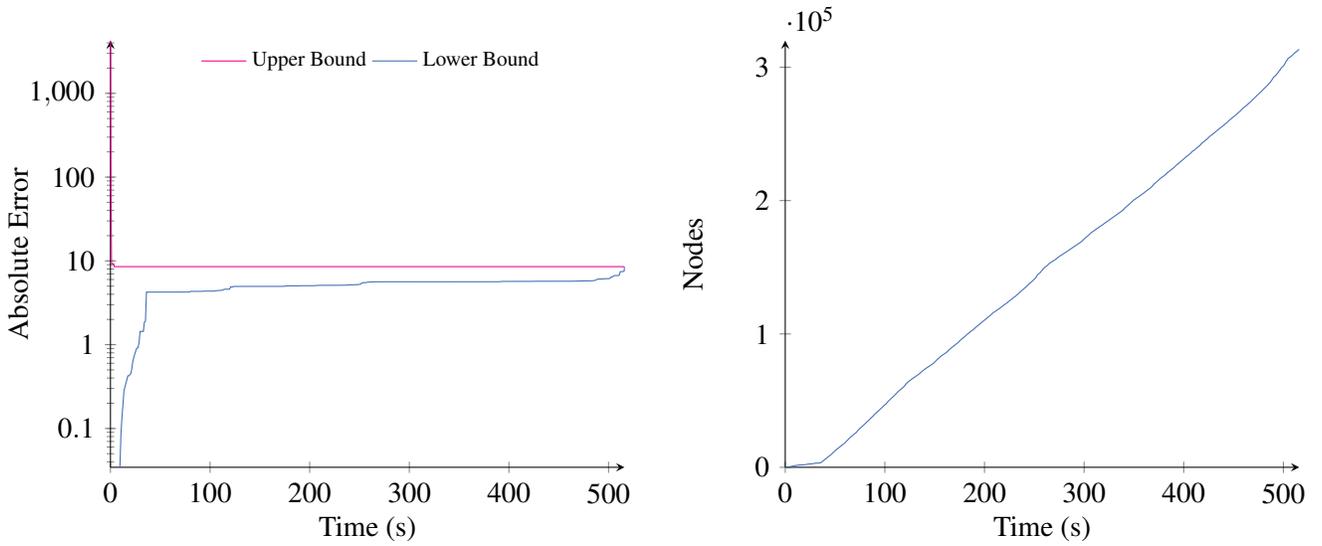

     \centering
     \begin{subfigure}[t]{0.49\textwidth}
        \centering
        \resizebox{\linewidth}{!}{
}
     \end{subfigure}
\caption{(a) Change with time in upper bound and lower bound values for the MINLP associated with the gentree for $L_1 + L_2 + L_3$; (b) Growth in number of branch-and-bound nodes with time.}
\label{fig:b2_growtha}
\end{figure}

\begin{figure}[h]
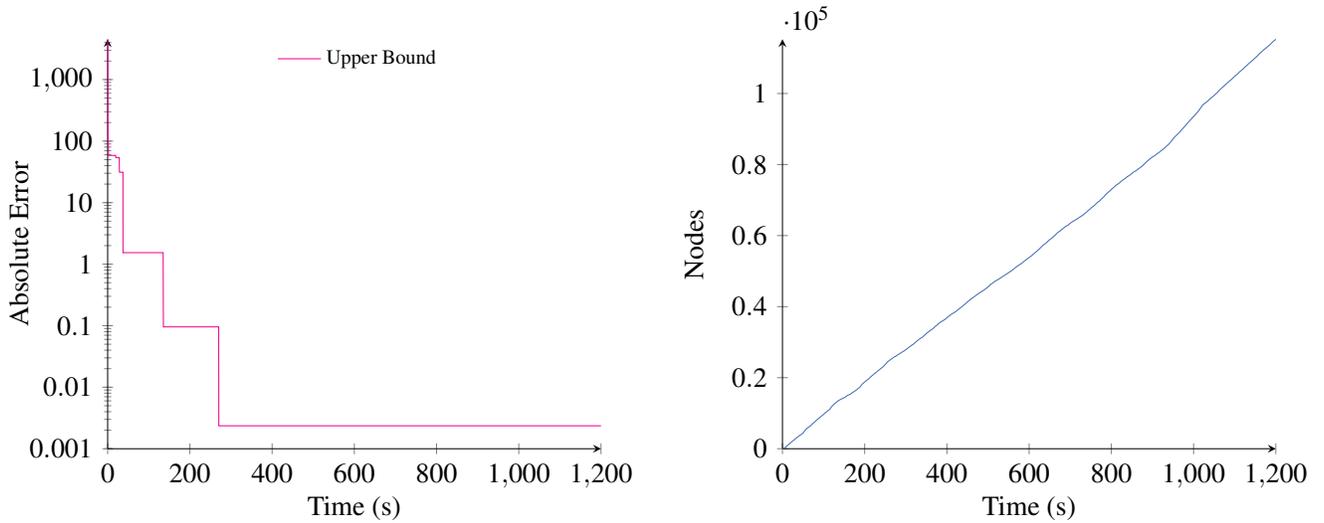

     \centering
     \begin{subfigure}[t]{0.49\textwidth}
        \centering
        \resizebox{\linewidth}{!}{
        %
}
     \end{subfigure}
\caption{(a) Change with time in upper bound for the MINLP associated with the gentree $\big(\sqrt{(L_1 + L_2)}\cdot(L_3 + L_4)\big)/L_5$ (the lower bound remains zero throughout); (b) Growth in number of branch-and-bound nodes with time.}
\label{fig:b2_growthb}
\end{figure}

Solving the MINLP problem associated with each gentree can itself be a hard problem. MINLP solvers such as BARON typically use a branch-and-bound algorithm to solve such a problem to a prescribed tolerance.  
Upon termination (assuming a minimization problem) BARON returns both a solution and a lower bound on all possible solution objective values such that the returned solution objective value and the lower bound differ by less than the prescribed tolerance. However, the number of branch-and-bound nodes -- and the computing time -- can grow exponentially. To avoid exponential growth, we terminate the process within a prescribed time limit and end up with a large gap between the objective value and the lower bound. In Figures~\ref{fig:b2_growtha} and \ref{fig:b2_growthb}, we show how the upper bound (on the absolute error between the derived model and data) and lower bound change with time during the MINLP solution process along with the growth in number of branch-and-bound nodes. In the first figure we observe that shortly after 500 seconds, BARON certifies that the optimal error is roughly 8.5. In the second figure we observe that the error drops to roughly 0.002 shortly before 300 seconds, but the process does not terminate as BARON is unable to improve the lower bound from 0 (and termination conditions are either the absolute gap between bounds should be $10^{-4}$ or the relative gap between bounds should be $10^{-3}$). Consider the single MINLP associated with the gentree in Figure~\ref{fig:b2_growthb}, and assume we have 3 independent variables and the powers of each variable are constrained to be integers between 2 and -2 (see the later section on our experimental settings). The number of possible discrete choices to be made (i.e., the power of each variable in each L-monomial) is at least $5^{15} \approx 30$ billion. BARON is able to obtain a solution of reasonable quality by examining about 120,000 subproblems.

\subsection*{Dimensional analysis}
Consider Newton’s law of universal gravitation which says that the gravitational force between two bodies is proportional to the product of their masses and inversely proportional to the square of the distance between their centers (of gravity): 
$$F \propto \frac{m_1 m_2}{r^2} $$
and the constant of proportionality is G, the gravitational constant. Therefore, we have
$$F=G\, \frac{m_1 m_2}{r^2}.$$
The units of $G$ are chosen so that units of the right-hand-side expression equal the units of force (mass $\times$ distance $/$  time-squared).
Suppose one is given a data set for this example, where each data item has the masses of two bodies and the distance and gravitational force between them.
Dimensional analysis would rule out $F = m_1m_2/r^2$ or $F = m_1/m_2 + m_2r$, for example, as possible solutions.%

If constants can have units and every L-monomial can have a such a constant, then dimensional analysis conveys essentially no information.
For example, we can choose constants $h_1, h_2, h_3$ with appropriate dimensions so that both the expressions $F = h_1m_1m_2/r^2$ and $F = h_2m_1/m_2 + h_3m_2r$ satisfy all dimensional requirements.
We allow constants to have units or not based on an input flag.

We next explain the constraints we add to our formulation to enforce dimensional consistency.
Assume that all constants have no units, and that $k=2$, and $\delta = 2$, and $d=1$.
For the gravitation example (without the gravitational constant $G$ as an input), each L-monomial $L$ has the form
\begin{equation}L= h m_1^a m_1^b r^c,\label{lmon2}\end{equation}
where $a$, $b$ and $c$ are bounded integer variables with an input range of $[-2,2]$, and $h$ is a variable representing a constant.
We add to the system of constraints (\ref{power1}) - (\ref{integrality})
linear constraints that equate the units of the symbolic expression to those of the dependent variable. For the case our gentree consists of a single node (as in the first tree in Figure 1), our symbolic expression has the form (\ref{lmon2}), and we add the linear constraints
$$
 a 
 \left[
\begin{matrix}
 1 \\ 0\\ 0
\end{matrix} 
\right]
+ b 
 \left[
\begin{matrix}
 1 \\ 0\\ 0
\end{matrix} 
\right]
+ c 
 \left[
\begin{matrix}
 0 \\ 1\\ 0
\end{matrix} 
\right]
=
 \left[
\begin{matrix}
 ~~~1 \\ ~~~1\\ -2
\end{matrix} 
\right]~.
 $$
Here, each component of a column vector corresponds to a unit (mass, distance, time, respectively).
The column on the right-hand side represents the units of force, i.e., mass times distance divided by squared time. The columns on the left-hand side represent the dimensionalities of mass and distance, respectively.
The third of the above linear equations does not have a solution. 

Next, assume the symbolic expression is $L_1+L_2$ where $L_1=h_1m_1^{a_1} m_2^{b_1} r^{c_1}$ and $L_2=m_1^{a_2} m_2^{b_2 } r^{c_2}$, and the unkowns are $a_i, b_i, c_i$ and $h_i$. The units of $L_1$ and $L_2$ have to match, and must also equal the units of $F$. The linear constraints we add are
\begin{align}
a_1 \label{eq:constraints1}
\left[
\begin{matrix}
 1 \\ 0\\ 0
\end{matrix} 
\right]
+ b_1 
\left[
\begin{matrix}
 1\\ 0\\ 0
\end{matrix} 
\right]
+ c_1 
\left[
\begin{matrix}
 0 \\ 1\\ 0
\end{matrix} 
\right]
&= 
\left[
\begin{matrix}
 ~~~1 \\ ~~~1\\ -2
\end{matrix} 
\right]
\\ 
\ \nonumber \\
a_2 \label{eq:constraints2}
\left[\begin{matrix}
 1 \\ 0\\ 0
\end{matrix} 
\right]
+ b_2 
\left[\begin{matrix}
 1 \\ 0\\ 0
\end{matrix} 
\right]
+ c_2 
\left[\begin{matrix}
 0 \\ 1\\ 0
\end{matrix} 
\right]
&=
\left[
\begin{matrix}
 ~~~1 \\ ~~~1\\ -2
\end{matrix}
\right]
. 
\end{align}

Finally, for the symbolic expression $L_1 \times L_2$, to compute the units of this expression we need to add up the units of $L_1$ and of $L_2$. Thus we sum the left hand sides of (\ref{eq:constraints1}) and (\ref{eq:constraints2}) and equate this sum to the right hand side of (\ref{eq:constraints1}).
We can similarly deal with dimension matching in the remaining gentrees of depth 1 via linear constraints.
For greater depth gentrees, we apply the ideas above to depth 1 (non-leaf nodes), and then to depth 2 nodes and so on.

\subsection*{Settings for experiments}
Because of the exponential worst-case scaling behavior of our method, we choose specific parameters that allow our method to terminate in a reasonable amount of time.

We set the depth limit $d$ to 3. We experiment with different values of $k$, the number of constants, for each dataset. We terminate when we find an expression with objective value (squared error) less than $10^{-4}$ (error tolerance).
The operators we use are described in Section {\it System Description} of {\it Symbolic Regression}. We set $\Omega$ to 100, so all constants are in the range $[-100,100]$.
We set $\delta = 2$ for computational efficiency. This means that each power in an L-monomial lies in the range $[-2,2]$.
Under this setting, the pruning rules -- R1, R2,R3 -- remove potentially non-redundant gentrees,
and our algorithm does not explore all possible symbolic expressions that are representable as a gentree of depth up to 3. Though L-monomials are closed under multiplication, L-monomials with bounded powers are not.
For example, if there is a solution of the form $x_1^3x_2^2$, we may not find it.

We constrain the search in three more ways.
We bound the sum of the variable powers in an L-monomial by an input number $\tau = 6$.
Thus, for each term of the type (\ref{lmonomial}), we add the constraint $\sum_{i=1}^n|a_i| \leq \tau$, which is representable by a linear constraint using $n$ auxiliary variables $a_i': \sum_{i=1}^n a_i' \leq \tau$ and $a_i' \geq 0, -a_i \leq a_i'$ and $a_i \leq a_i'$ for $i=1, \ldots, n$. 
Secondly, we add another pruning rule. We remove any gentree which contains a subexpression of the form $\sqrt{L}$ where $L$ is an L-monomial (we allow $\sqrt{L_1+L_2}$, for example).
\new{We use dimensional analysis when feasible, other than when we perform a comparison with other SR codes (for example in Tables \ref{tab:ai-feynman-results} - \ref{tab:sota_reasoning}) which do not use dimensional analysis.}

\section*{Reasoning and Derivability}

\subsection*{Logic and Reasoning Background}

We assume the reader has knowledge of basic first-order logic and automated theorem proving terminology and thus will only briefly describe the terms commonly seen throughout this paper. For readers interested in learning more about logical formalisms and techniques see \cite{thelogicbook,enderton2001mathematical}.

In this work, we focus on first-order logic (FOL) with equality and basic arithmetic operators.
In the standard FOL problem-solving setting, an Automated Theorem Prover (ATP), also called a reasoner, is given a \emph{conjecture}, that is, a formula to be proved true or false, and \emph{axioms}, that is, formulae known to be true.
The set of axioms is also called the \emph{background theory}.
By application of an inference rule, a new true formula can be derived from the axioms.
This operation can be repeated, including the use of prior derived formulae; this yields 
a sequence of new true formulae, called a \emph{derivation}.
This is done until the given conjecture appears among the derived formulae; the sequence of applied rules and formulae along the way comprise a \emph{proof} of the given conjecture. 
The set of all \emph{derivable formulae} from a background theory is the set of all logic formulae that can be derived from the set of axioms defining the background theory.

All formulae considered in this work
are FOL formulae with equality and basic arithmetic operators, and so are defined based on the FOL grammar with the addition of the function symbols for equality $=$, inequality $>, ~<$, sum $+$, subtraction $-$, product $*$, division $/$, power \lstinline{^} (the square root is interpreted as a power of $\frac{1}{2}$) and absolute value $|\cdot|$.

\emph{First-order logic formulae} are formal expressions based on an alphabet of predicates, functions, and variable symbols which are combined by logical connectives.
A term is either a %
variable, a constant (function with no arguments), or, inductively, a function applied to a tuple of terms. A formula is either a predicate applied to a tuple of terms or, inductively, a connective (e.g., $\wedge$ read as ``and'', $\vee$ as ``or'', $\neg$ as negation, etc.) 
applied to some number of formulae. %
In addition, variables in formulae can be universally or existentially quantified (i.e., by the quantifiers $\forall$ and $\exists$ which read as ``for all'' and  ``exists'' respectively), where a quantifier introduces a semantic restriction for the interpretation of the variables it quantifies.

\subsection*{\keymaeraX}
Reasoning software tools for arithmetic and calculus that, in principle, have the required logic capabilities of checking consistency, validation, and deduction include:
SymPy~\cite{sympy}, Prolog~\cite{prolog}, Mathematica~\cite{mathematica}, Beagle~\cite{Beagle2015}, and KeYmaera~\cite{keymaera,keymaera_code}.

We integrated in our system several reasoning tools for arithmetic and calculus, analyzing their expressiveness and capabilities (including, but not only, the one mentioned above).
Our findings made us decide in favour of the \keymaeraX reasoner, an automated theorem prover for hybrid systems that combines different types of reasoning: deductive, real algebraic, and computer algebraic reasoning.
The underlying logic supported by \keymaera is differential dynamic logic~\cite{differentialdynamiclogic}, which is a real-valued first-order dynamic logic for hybrid programs.
\keymaera is based on a generalized free-variable sequent calculus inference rule for deductive reasoning and has an underlying CAD system (e.g. Mathematica).

\keymaeraX provides fast computation for first-order logical formulae combined with arithmetic and differential equations, even though it does not provide full explainability for the derivations, meaning that only the final state (provable/not-provable) is available, while the proof-steps are not.

\subsection*{\keymaeraX formulation for Kepler's third law}

We first consider the \keymaera formulation\footnote{We omit Variables and Definitions from the \keymaeraX formulation for simplicity.} for the $\beta^r_{\infty}$ error.
As an example, 
we consider $f(\mathbf{x})$ corresponding to $p = \sqrt{0.1319 \cdot d^3}$ extracted via SR from the solar system dataset.
We compute the error between this function
and $f_{\mathcal{B}}$, represented by $\mathtt{p}$, which is the  variable corresponding to the orbital period\footnote{Note that in the \keymaeraX formulation we  add the suffix $N$ to the variables after they have been normalized (e.g., $\mathtt{p}$ become $\mathtt{pN}$, $\mathtt{m_1}$ become $\mathtt{m1N}$, $\mathtt{d}$ become $\mathtt{dN}$, etc.). In the text description we use the original variable names for simplicity.}.

\nolinenumbers
\begin{lstlisting}
Problem (( m1>0 & m2>0 & p>0 & d2>0 & d1>0 & m1N>0 & m2N>0 & pN>0 
    & G = (6.674 * 10^(-11)) & pi = (3.14) & err =  10^-2 
    & ( m1 * d1 = m2 * d2 ) 
    & ( d = d1 + d2 )  
    & ( Fg = (G * m1 * m2) / d^2 )  
    & ( Fc =  m2 * d2 * w^2 )  
    & ( Fg =  Fc )  
    & ( p = (2 * pi )/w )
    & convP = (1000 * 24 * 60 * 60 )
    & convm1 = (1.9885 * 10^30  )
    & convm2 = (5.972 * 10^24)
    & convD = (1.496 * 10^11)
    & m1N = m1 / convm1
    & m2N = m2 / convm2
    & dN = d / convD
    & pN = p / convP
   )->(((m1N=1 & m2N=0.055 & dN=0.3871) -> 
                abs((0.1319 * dN^3 )^(1/2) - (pN))/pN < err)
     & ((m1N=1 & m2N=0.815 & dN=0.7233) -> 
                abs((0.1319 * dN^3 )^(1/2) - (pN))/pN < err)
     & ((m1N=1 & m2N=1 & dN=1) -> 
                abs((0.1319 * dN^3 )^(1/2) - (pN))/pN < err)
     & ((m1N=1 & m2N=0.107 & dN=1.5237) -> 
                abs((0.1319 * dN^3 )^(1/2) - (pN))/pN < err)
     & ((m1N=1 & m2N=317.8 & dN=5.2044) -> 
                abs((0.1319 * dN^3 )^(1/2) - (pN))/pN < err)
     & ((m1N=1 & m2N=95.159 & dN=9.5826) -> 
                abs((0.1319 * dN^3 )^(1/2) - (pN))/pN < err)
     & ((m1N=1 & m2N=14.536 & dN=19.2184) -> 
                abs((0.1319 * dN^3 )^(1/2) - (pN))/pN < err)
     & ((m1N=1 & m2N=17.147 & dN=30.07) -> 
                abs((0.1319 * dN^3 )^(1/2) - (pN))/pN < err)
) ) End.
\end{lstlisting}
\linenumbers

In particular we see that:
\begin{itemize}
    \item {\bf line 1} describes the feasibility constraints, e.g., the masses only admit positive values;
    \item {\bf line 2} defines some constants such as the gravitational constant, the value of $\pi$ ($\mathtt{pi}$), and the error bound ($\mathtt{err} = 10^{-2}$ );
    \item {\bf lines 3-8} are the axioms of the background theory;
    \item {\bf lines 9-16} are the normalization specifications (e.g. $\mathtt{m1N}$ is the normalized variable corresponding to the mass $\mathtt{m_1}$);
    \item {\bf lines 17-24} contain the specification of the $\beta^r_{\infty}$ error: given each data point, which specifies the value of the two masses and their relative distance, we want to prove that the distance between the formula induced from the data (in this case $p = \sqrt{0.1319 \cdot d^3}$) and the actual formula derivable from the axioms for $p$ is smaller than the value of $\mathtt{err}$. In this way, if all the data points respect the error bound, the $\max$ value will as well.
\end{itemize}

\keymaera is able to produce a successful proof of the formulation above for an error bound of $\mathtt{err} = 10^{-2}$.

The value for the reasoning error $\beta^r_{\infty}$ is computed by using binary search over the values of the relative error  $|(\sqrt{0.1319 * d^3 } - p)|/p$.
In these experiments we used a time limit of $1200$ seconds (20 min) and stopped the binary search process when a precision of $10^{-4}$ is attained.

The formulation for $\beta^r_{2}$ error is very similar, with the difference that the binary search is performed over the single data points as described in Algorithm~\ref{alg:l2norm}, where $$\min_e{\{\keymaera\ell_2(i,e)\mid \texttt{precision}\}}$$ computes an upperbound on the minimum value for $e$ for a give precision level (which is $10^{-4}$ in the experiments) such that $\keymaera \ell_2 (i, e)$ returns $true$.
This is necessary because \keymaera is fundamentally a boolean function, mapping formulations to a success/failure state and is thus not able to perform optimization over continuous values. 
The minimum is therefore calculated approximately (via binary search) in a given interval with an input precision level and fixing a time limit (20 min in the experiments).

\begin{algorithm}[h]
\caption{$\beta^r_{2}$ computed with \keymaera}\label{alg:l2norm}
\begin{algorithmic}[1]
\Procedure{$\beta^r_{2}$}{\texttt{Dataset}}
\For{$i$ \text{ in }\texttt{Dataset}} 
    \State $e_i \gets \min_e{\{\keymaera \ell_2(i,e) \mid \texttt{precision}\}}$
\EndFor
\State $\beta^r_{2} \gets \sqrt{e_i^2 + \cdots + e_n^2}$
\\
\Return $\beta^r_{2}$
\EndProcedure
\end{algorithmic}
\end{algorithm}

The \keymaera formulation for $\keymaera \ell_2(i, e)$ on a specific data point $i$ and an error bound $e$ is a Boolean function that returns $true$ if the following program is provable and $false$ otherwise. $\keymaera \ell_2(i, e)$ returns $true$ when, for a given set of axioms, the absolute value of the relative distance between a given function and a derivable one is smaller than $e$ for a given data point $i$. For example, given the data point $(m_{1},m_{2},d) = (1,0.055,0.3871)$ and error $e = 10^{-2}$ we have the following formulation:

\nolinenumbers
\begin{lstlisting}
Problem (( m1>0 & m2>0 & p>0 & d2>0 & d1>0 & m1N>0 & m2N>0 & pN>0
    & G = (6.674 * 10^(-11)) & pi = (3.14) & e =  10^-2
    & ( m1 * d1 = m2 * d2 ) 
    & ( d = d1 + d2 )  
    & ( Fg = (G * m1 * m2) / d^2 )  
    & ( Fc =  m2 * d2 * w^2 )  
    & ( Fg =  Fc )  
    & ( p = (2 * pi )/w )
    & convP = (1000 * 24 * 60 * 60 )
    & convm1 = (1.9885 * 10^30  )
    & convm2 = (5.972 * 10^24)
    & convD = (1.496 * 10^11)
    & m1N = m1 / convm1
    & m2N = m2 / convm2
    & dN = d / convD
    & pN = p / convP
   )->(((m1N=1 & m2N=0.055 & dN=0.3871) -> 
                abs((0.1319 * dN^3 )^(1/2) - (pN))/pN < e)
) ) End.
\end{lstlisting}
\linenumbers

We observed that the point-wise reasoning errors are not very informative if SR yields a low-error candidate expression (measured with respect to the data), 
and the data itself satisfies the background theory up to a small error, 
which indeed is the case with the data we use; the reasoning errors and numerical errors are very similar.
This is true because if we can evaluate the numerical error, i.e., we have data for the real values of $p$, 
then we may assume that the error of the candidate formula at the data points is substantially equal to the error of the correct formula. 
In particular, this is true when the data is generated synthetically like in \cite{AI-Feynman}.

However, this analysis is still relevant if we wish to evaluate the error in the data:
$$ \textit{Error in the data } = 
   \textit{ Numerical error } 
   - \textit{ Reasoning error}~.$$
The latter is useful for avoiding overfitting entailed in formulae with numerical error smaller than error in the data.

Let's consider again the formula $ \sqrt{0.1319 \cdot d^3}$  (extracted via SR from the solar system dataset) for the variable of interest $p$. 
The \keymaera formulation for the generalization relative reasoning error $\beta^r_{\infty,S}$ is the following:

\nolinenumbers
\begin{lstlisting}
Problem (( p>0 & pN>0
    & G = (6.674 * 10^(-11)) & pi = (3.14) & err =  10^-6
    & m1N>=0.08 & m1N<=1
    & m2N>=0.0002 & m2N<=1
    & dN>=0.0111 & dN<=30.1104
    & ( m1 * d1 = m2 * d2 ) 
    & ( d = d1 + d2 )  
    & ( Fg = (G * m1 * m2) / d^2 )  
    & ( Fc =  m2 * d2 * w^2 )  
    & ( Fg =  Fc )  
    & ( p = (2 * pi )/w )
    & convP = (1000 * 24 * 60 * 60 )
    & convm1 = (1.9885 * 10^30  )
    & convm2 = (5.972 * 10^24)
    & convD = (1.496 * 10^11)
    & m1N = m1 / convm1
    & m2N = m2 / convm2
    & dN = d / convD
    & pN = p / convP
   )->((abs(0.1319 * dN^3 )^(1/2)  - pN ) / (pN) < err )) 
End.
\end{lstlisting}
\linenumbers

In particular we have that:
\begin{itemize}
    \item {\bf line 1} describes the feasibility constraints;
    \item {\bf line 2} defines some constants such as the gravitational constant, $\pi$ value and the error bound;
    \item {\bf lines 3-5} define the intervals $S$ for the variables, extracted from the dataset normalized data points (e.g. the planets in the solar system have a distance more than $0.0111$ astronomical units and less than $30.1104$ astronomical units from the sun);
    \item {\bf lines 6-11} are the axioms of the background theory;
    \item {\bf lines 12-19} are the normalization specifications;
    \item {\bf lines 17-24} contain the specification of the error $\beta^r_{\infty,S}$: given each data point (instantiation of the value of the two masses and their relative distance) we want to prove that the distance between the formula induced from the data (in this case $\sqrt{0.1319 \cdot d^3}$) and the actual formula derivable from the axioms for $p$ is smaller than the value of $\mathtt{err}$. 
\end{itemize}

\subsection*{\keymaera formulation for relativistic time dilation}

We give a formulation to check for the quality of generalization for a given formula, in particular, the function $y = -0.00563 v^2$ computed from data.
The velocity of light $c$ is given as a fixed constant 3 $\times 10^8$ (meters per second). The generalized absolute reasoning error $\beta^a_{\infty,S}$ is bounded above by $1$. The following formulation checks if all the derivable formulae differs (after scaling by $10^{15}$) from the function $-0.00563 v^2$ by at most $1$ in the domain $S$ defined by $37 \leq v \leq 115$. 

\nolinenumbers
\begin{lstlisting}
Problem((d > 0 & L > 0 & c = 3*10^8 & err = 1 
        & v >= 37 & v <= 115
        & dt0 = 2*d/c 
        & L^2 = d^2 + (v*dt/2)^2 
        & dt = 2*L/c 
        & f0 = 1/dt0 
        & f = 1/dt 
        & df = f - f0 
        & y = 10^15*df/f0 )
        -> ( ( abs(  (-0.00563)*(v^2)) - y  < err ) ) ) 
End.
\end{lstlisting}
\linenumbers

\subsection*{\keymaeraX formulation for proving Langmuir's adsorption equation}
The \keymaeraX formulation to prove the Langmuir equation:
\begin{equation}
    q = \frac{S_0 \cdot (k_{\mathrm{ads}}/k_{\mathrm{des}}) \cdot p}{1 + (k_{\mathrm{ads}}/k_{\mathrm{des}}) \cdot p}.
    \label{eq:LangmuirFull}
\end{equation}
from the background theory described in the manuscript (L1--L5)
is the following:

\nolinenumbers
\begin{lstlisting}
Definitions
  Real kads;
  Real kdes;
  Real S0;
End.
Problem
( \forall Q \forall S \forall Sa \forall P \forall rdes \forall rads(
        (kads>0 & kdes>0 & S0>0 & Q>0 & S>0 & Sa>0 & P>0 &  rdes>0 & rads>0
        & S0 = S + Sa
        & rads = kads * P * S
        & rdes = kdes * Sa
        & rads = rdes
        & Q = Sa)
        -> Q = ((S0 * (kads / kdes)) * P) / (1 + ((kads / kdes)*P) ) ))
End.
\end{lstlisting}
\linenumbers

This formulation is successfully proved by KeYmaera.

As expected, when removing one or more axioms that are strictly necessary to prove the theorem (e.g., the axiom $r_{\mathrm{des}} =  k_{\mathrm{des}}\cdot S_{\mathrm{a}}$), it is not possible to prove the formula anymore.
Similarly, if $f$ does not correspond to the correct formula (while the axiom set is complete and correct) \keymaera is not able to provide a provability certification. In this case it is possible to create a numerical counterexample
(a numerical assignment to all the variables involved in the problem formalization, that falsify the implication). 
Moreover, it is possible to add additional constraints and assumptions to generate further counterexamples. 
Moreover, with the addition of redundant or unnecessary axioms we are still able to prove the conjecture.

An example of a counterexample, obtained by removing an essential (to prove the theorem) axiom from the background theory (the axiom $r_{\mathrm{des}} =  k_{\mathrm{des}}\cdot S_{\mathrm{a}}$), is the following:
$$[Q,S,P,S_0,r_{des}, S_a, k_{des}, k_{ads} r_{ads}] = [1,1,1/2,2,1/2,1,1,1,1/2]$$
It is  possible to add additional constraints and assumptions to generate further counterexamples: e.g., if we exclude the value $1$ for the variable $Q$ (the assignment of $Q$ in the previous counterexample)  adding the constraint $Q \mathrel{\mathtt{!=}} 1$, the Counter-Example Search tool provides a new solution: 
$$[Q,S,P,S_0,r_{des}, S_a, k_{des}, k_{ads} r_{ads}] = [1/4,1/4,1/2,1/2,1/8,1/4,1,1,1/8]$$

Langmuir defined expressions to model adsorption onto a material that contains different kinds of sites, with different interaction strengths \cite{Langmuir1918}.
Consequently, we generalize the background theory %
to model this scenario.
Some quantities ($S_{\mathrm{a}}$, $S_0$, $S$, $k_{\mathrm{ads}}$, $k_{\mathrm{des}}$, $r_{\mathrm{ads}}$, and $r_{\mathrm{des}}$) now depend on the site.
The axioms below, %
which are universally quantified over the sites,
hold for each site as before, 
while the definition of the adsorbed amount changes to allow for the presence of multiple sites.
The new set of axioms is:\\
\parbox{3in}{
\begin{tabular}{lll}
& GL1. & $\forall X. ~S_0(X) = S(X) + S_a(X) $            \\[4pt]
& GL2. & $\forall X. ~r_{ads}(X) = k_{ads}(X) \cdot P \cdot S(X) $    \\[4pt]     
& GL3. & $\forall X. ~r_{des}(X) = k_{des}(X) \cdot S_a(X)$ \\[4pt]
& GL4. & $\forall X. ~r_{ads}(X) = r_{des}(X) $ \\[4pt]
& GL5. & $Q = \sum_{i=1}^{\infty} S_a(X) $   \\[4pt]
& GL6. & $\bigwedge_{i=1}^{\infty}  \left( St=i \rightarrow (\bigwedge_{j>i} S_0(j)=0) \right)$   \\[4pt]
& GL7. & $\bigvee_{i=1}^{\infty} St=i$   \\
\end{tabular}
}
~\\
where GL1--GL4 are as before; GL5 ensures that the total amount adsorbed is the sum of the amounts adsorbed in each site; GL6 ensures that if there are $i$ site types ($St=i$) the types with higher index have no sites; and GL7 ensures that there is only one value for the number of site types.

We implemented an alternative \keymaera formulation for the two-sites model:

\nolinenumbers
\begin{lstlisting}
Definitions
  Real kads_1;
  Real kdes_1;
  Real S0_1;
  Real kads_2;
  Real kdes_2;
  Real S0_2;
End.
ProgramVariables
  Real Q;
  Real S_1;
  Real S_2;
  Real Sa_1;
  Real Sa_2;
  Real P;
  Real rdes_1;
  Real rads_1;
  Real rdes_2;
  Real rads_2;
End.
Problem
( ( kads_1>0 & kdes_1>0 & S0_1>0 &  S_1>0 & Sa_1>0 & rads_1>0 & rdes_1>0
      & kads_2>0 & kdes_2>0 & S0_2>0 &  S_2>0 & Sa_2>0 & rads_2>0 & rdes_2>0 & 
      & Q>=0 & P>0 
      & S0_1 = S_1 + Sa_1
      & rads_1 = kads_1 * P * S_1
      & rdes_1 = kdes_1 * Sa_1
      & rads_1 = rdes_1
      & S0_2  = S_2  + Sa_2 
      & rads_2  = kads_2  * P * S_2 
      & rdes_2  = kdes_2 * Sa_2 
      & rads_2  = rdes_2 
      & Q = Sa_1 + Sa_2 )
    -> Q = (S0_1 * (kads_1 / kdes_1) * P)/(1+(kads_1/kdes_1) *P) 
    +(S0_2 * (kads_2/ kdes_2) * P)/ (1 + ((kads_2 / kdes_2) * P)))
End.
\end{lstlisting}
\linenumbers

This formulation is successfully proved by KeYmaera. This formulation is equivalent to the one described for a generic number of sites (in the case of at most two sites), except that it duplicates variables used in the axiom system for the one-site model instead of creating dependencies on the site type. This grounding of the general formulation helps the reasoner converge faster to a solution if the number of sites is small. However, this approach applied to $n$ sites would result in an exponential growth (with $n$)
of the number of formulae and would lead to increasing computing times. In general, a more compact representation (with fewer logical/non-logical symbols)
is preferred.

When the input formula (to be derived) has numerical constants, we need to introduce existentially quantified variables. For example, given an input formula $P/(0.00927 ~ P+0.0759)$, we first transform it to  
$P /(c_1 ~ P + c_2)$ (where $c_1$ and $c_2$ are existentially quantified). The following \keymaera formulation is used:

\nolinenumbers
\begin{lstlisting}
Definitions
  Real kads;
  Real kdes;
  Real S0;
End.
Problem
( \exists c1 \exists c2 \forall Q \forall S 
   \forall Sa \forall P \forall rdes \forall rads(
    (kads>0 & kdes>0 & S0>0 & Q>0 & S>0 & Sa>0 & P>0 &  rdes>0 & rads>0 
    & S0 = S + Sa
    & rads = kads * P * S
    & rdes = kdes * Sa
    & rads = rdes
    & Q = Sa)
    -> ( c1>0 & c2>0 & Q =  P/(c1 * P +c2) )))
End.
\end{lstlisting}
\linenumbers

This formulation is successfully proved by \keymaera, however it does not provide any (symbolic or numerical) instantiation  for the existentially quantified variables that satisfy the logic program.
In this case, manual inspection can show that instantiating $c_1 = 1/S_0$ and $c_2 = \frac{k_{\mathrm{des}}}{k_{\mathrm{ads}} S_0}$ provides a valid solution. To obtain this result automatically would require an extension of \keymaeraX (or more broadly a theorem prover) to allow for explicit variable assignments. %

Likewise, the formulation for proving expressions generated by symbolic regression with the two-site Langmuir model is the following:

\nolinenumbers
\begin{lstlisting}
Definitions
  Real kads_1;
  Real kdes_1;
  Real S0_1;
  Real kads_2;
  Real kdes_2;
  Real S0_2;
End.
Problem
( \exists c1 \exists c2 \exists c3 \exists c4 \forall Q
  \forall S_1 \forall S_2 \forall Sa_1 \forall Sa_2 \forall P(
    ( S_1>0 & Sa_1>0 & S0_1>0 & kads_1>0 & kdes_1>0
    &  S_2>=0 & Sa_2>=0 & S0_2>0 & kads_2>0 & kdes_2>0
    & Q>=0 & P>0 
    & S0_1 = S_1 + Sa_1
    & kdes_1 * Sa_1 = kads_1 * P * S_1
    & S0_2  = S_2  + Sa_2 
    & kdes_2 * Sa_2   = kads_2  * P * S_2 
    & Q = Sa_1 + Sa_2)
    -> (Q = (c1*P^2+c2*P)/(P^2+c3*P+c4) & c1>0 & c2>0 & c3>0 & c4>0 )))
End.
\end{lstlisting}
\linenumbers

Using this formulation, \keymaera timed out without proving or disproving any of the conjectures provided by the SR module. However, we noticed that $g_5$ and $g_7$, of the form $q = \frac{c_1 p^2 + c_2 p}{p^2 + c_3 p + c_4}$, satisfy all the thermodynamic constraints $\mathcal{K}$ (in contrast to the other 4-parameter formulae), so we tried to prove them manually.
We were able to do so, obtaining the following variable instantiations: $c_1 = S_{0,1} + S_{0,2}$, $c_2 = \frac{S_{0,1} K_1 + S_{0,2} K_2}{K_1 K_2}$, $c_3 = \frac{K_1 + K_2}{K_1 K_2}$, and $c_4 = 1/(K_1 K_2)$, where $K_i = k_{\mathrm{ads},i}/k_{\mathrm{des},i}$.
\if
Interestingly, $g_7$ has a combination of constants that make it nearly equivalent to the 3-constant quadratic isotherm \cite{MOREAU1991127}:
\begin{equation*}
    q = \frac{q_{\mathrm{max}} (K_{a} p + h \sigma K_a^2 p^2)}{1 + 2 \sigma K_{a} p + h \sigma^2 K_a^2 p^2},
    \label{eq:quadratic}
\end{equation*}
with $K_a = 15.2$, $h = 1.75$, and $\sigma = 0.0118$.
\fi

\section*{Comparison with other systems}

\new{We compared our system with some state-of-the-art methods on the three real-life problems (and associated datasets) that we considered in this work. In addition, we also performed a comparison on 81 problems out of 100 from the Feynman Symbolic Regression Database \cite{AI-Feynman}; the associated formulas are taken from the Feynman lectures on physics, and we ignored the 19 that contain trigonometric functions. To create the data for each of the 81 problems, we chose 10 datapoints from each (synthetic) dataset and added 1\% error (in the manner described in \cite{AI-Feynman}), thereby creating small, noisy datasets to resemble the real-life data that we use. We note that the original dataset for each problem consists of 100,000 data points with zero error.}

The list of systems we compared against is not meant to be comprehensive, but is a representative sample of state-of-the-art methods. we excluded some systems from the list because they were either not applicable (e.g., LGGA\footnote{The comparison with LGGA was only possible for Langmuir’s problem, since the system requires constraints on the functional form of $f$. We defined such constraints -- the constraints $K$ --  only for Langmuir’s problem. However, the only constraint that is supported by LGGA is $f(0)=0$. The other constraints we use (e.g., monotonicity and limiting properties) are not currently supported by LGGA. We were thus unable to obtain good results and therefore decided not to report them.}~\cite{ashok2020logic}), not freely licensed (e.g., Eureqa~\cite{schmidt14}) or because the code was not available (e.g., LGML~\cite{scott2021lgml}).

The systems we considered are the following:
\begin{itemize}
    \item {\bf AI-Feynman}~\cite{AI-Feynman, AI-Feynman2.0} is a symbolic regression algorithm that combines deep learning techniques with the exploitation of some characteristics typically present in functions studied in physics such as the presence of units, the use of low-order polynomials, symmetry, etc. The algorithm combines multiple modules that perform dimensional analysis, polynomial fit, brute-force enumeration, and neural network fit (and a few other tasks) to produce a list of candidate formulas.
    \item {\bf TuringBot}~\cite{schmidt2009distil} is a simulated annealing method to find expressions that fit the input data. The free version of the software allowed us to consider a maximum of 2 independent variables per problem. For Langmuir’s adsorption equation and for the relativistic time dilation problem two independent variables are enough. For Kepler’s third law, two independent variables are enough except for the Binary Stars dataset; we do not report results with this dataset.
    \item {\bf PySR}~\cite{pysr,cranmer2020discovering}, is a tool that uses regularized evolution, simulated annealing, and gradient-free optimization to search for equations that fit the input data.
    \item {\bf Bayesian Machine Scientist (BMS)}~\cite{bayesian_machine_scientist} is a Markov chain Monte Carlo based method that explores the space of possible models, based on prior expectations learned from a large empirical corpus of mathematical expressions.
\end{itemize}

The above systems return a score (in some cases this corresponds to the error) and the complexity for each output formula.
We applied standard methods to select the best candidates, such as computing the Pareto front and identifying knee points on it, a common technique to choose from the Pareto front candidates~\cite{knee_points,schmidt2009distil}. We used an existing python library, called \emph{kneed}~\cite{knee_points_python_lib}, that uses the kneedle algorithm~\cite{kneedle_algorithm} to compute the knee/elbow points of a function defined by a set of points. The knee point is the point of maximum curvature on such function. The systems PySR and Bayesian Machine Scientist return a best candidate and, therefore, for those methods we didn't identify the knee points.

\begin{figure}[h]
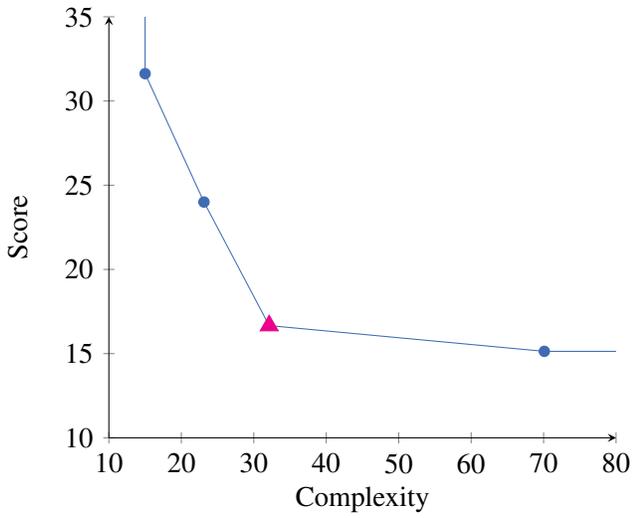
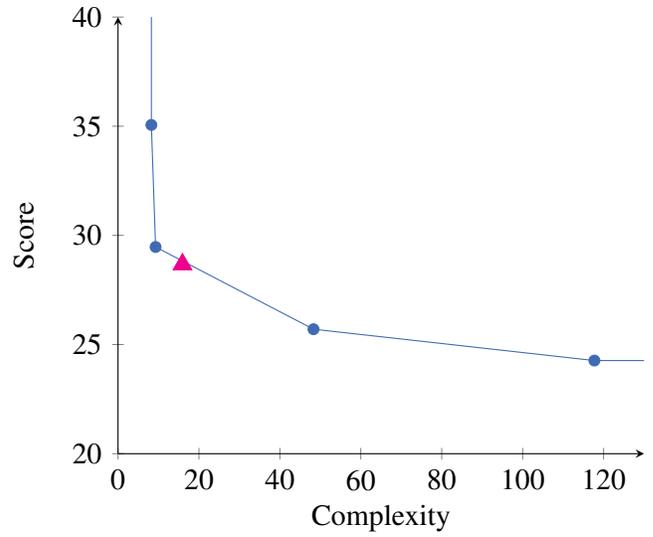

     \centering
     \begin{subfigure}[t]{0.49\textwidth}
        \centering
        \resizebox{\linewidth}{!}{
}
        \caption{Kepler Solar}
     \end{subfigure}
          \hfill
     \begin{subfigure}[t]{0.49\textwidth}
        \centering
        \resizebox{\linewidth}{!}{
        %
}
        \caption{Langmuir \cite[Table IX]{Langmuir1918}}
     \end{subfigure}
\caption{Pareto curves obtained with AI-Feynman software. The triangle indicates the knee point.}
\label{fig:b2_growtha1}
 \end{figure}

\begin{figure}[h]
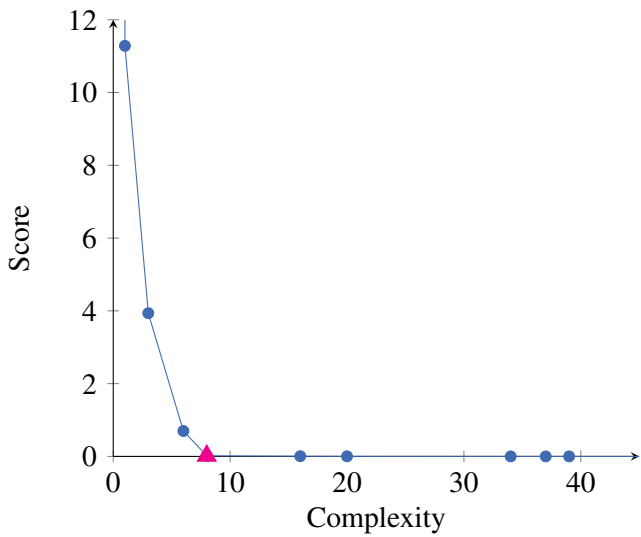
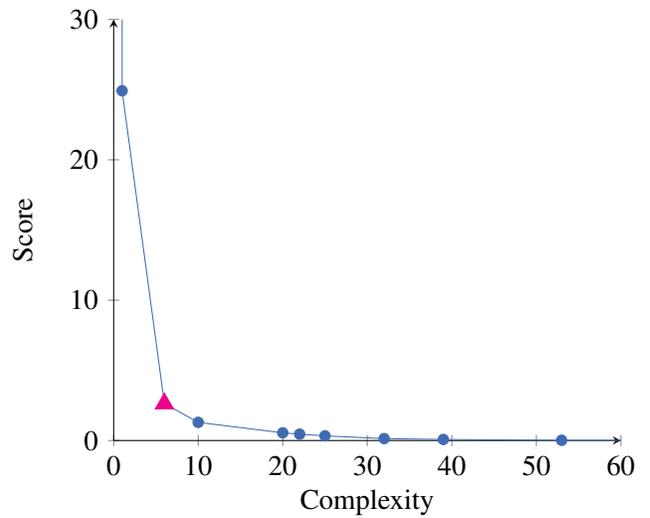

     \centering
     \begin{subfigure}[t]{0.49\textwidth}
        \centering
        \resizebox{\linewidth}{!}{
        %
}
        \caption{Kepler Solar}
     \end{subfigure}
          \hfill
     \begin{subfigure}[t]{0.49\textwidth}
        \centering
        \resizebox{\linewidth}{!}{
        %
}
        \caption{Langmuir \cite[Table IX]{Langmuir1918}}
     \end{subfigure}
\caption{Pareto curves obtained with TuringBot software. The triangle indicates the knee point.}
\label{fig:b2_growtha2}
 \end{figure}

The results for the different datasets are reported in Tables~\ref{tab:pareto_kepler_solar} --
~\ref{tab:pareto_langmuir_sunetal}. We show the Pareto front solutions for each method as well as the best candidate formula in the Pareto front (either the knee point or the best candidate formula returned by the corresponding system). We can see in 
Figures~\ref{fig:b2_growtha1} --
~\ref{fig:b2_growtha4} the comprehensive set of results for TuringBot, AI-Feynman, PySR and Bayesian Machine Scientist respectively, for two of the datasets (Kepler solar and Langmuir \cite[Table IX]{Langmuir1918}).

As we can see from the results the systems very rarely produce the correct formula as the best candidate. However, some systems output the correct formula in the top candidates, but not all of them. Even in the case in which the correct formula appears in the top candidates, the systems do not have a method to identify the right formula in this list. The application of the reasoning component of our system would definitively help to identify the right one.
The output of the systems is produced by using the default parameters configuration provided with the code (more details in Table~\ref{tab:config_param}). A fine-tuning of these parameters for each dataset could lead to better results. However, a general case, in which the correct formula is unknown, would not allow the selection of the best parameters, so we think that using default parameters is more realistic. 

\clearpage

\clearpage

\begin{figure}[h]
     \centering
     \begin{subfigure}[t]{0.49\textwidth}
        \centering
        \resizebox{\linewidth}{!}{
        %
}
        \caption{Kepler Solar}
     \end{subfigure}
     \hfill
     \begin{subfigure}[t]{0.49\textwidth}
        \centering
        \resizebox{\linewidth}{!}{
        %
}
        \caption{Langmuir \cite[Table IX]{Langmuir1918}}
     \end{subfigure}
\caption{Pareto curves obtained with PySR software. The triangle indicates the software best candidate.}
\label{fig:b2_growtha3}
 \end{figure}
\begin{figure}[h]
     \centering
     \begin{subfigure}[t]{0.49\textwidth}
        \centering
        \resizebox{\linewidth}{!}{
        %
}
        \caption{Langmuir \cite[Table IX]{Langmuir1918}}
     \end{subfigure}
\caption{Pareto curves obtained with BMS software. The triangle indicates the software best candidate.}
\label{fig:b2_growtha4}
 \end{figure}

\clearpage

\begin{table*}[h]
\centering
\begin{tabular}{ll}  
\toprule
Software & Candidate formula $p=$ \\ %
\midrule
\multirow{5}{*}{AI Feynman}
    & $ \arccos(-666.0 \sin{\pi}) (0.0 + d \sqrt{d}) $ \\[5pt] %
    & $ 0.3652 d $ \\[5pt] %
    & $\pmb{ 0.3652 d \sqrt{d} }$ \\[5pt] %
    & $ 8.455 d \sin(\sqrt{d} / e^{\pi}) $ \\ %
\midrule
\multirow{15}{*}{TuringBot}
    & $ d $ \\[5pt] %
    & $ 1.801 d $ \\[5pt] %
    & $ 0.8373 \ln \Gamma(d) $ \\[5pt] %
    & $\pmb{ 0.3632 d \sqrt{d} }$ \\[5pt] %
    & $ d \sqrt{0.1318 d - (-0.006766 \cos(d))} $ \\[5pt] %
    & $ d \sqrt{0.1316 d + 0.006946 (\cos(d) + 0.8752)} - 0.009080 $ \\[5pt] %
    & $ (\sqrt{0.01623 + d} / 2.755) \times $ \\ %
    & $(0.02646 \cos(-0.9837(0.3384 + d)) \tanh(-0.3400 + 0.1561 m_2) + d) $ \\[5pt] %
    & $ (\sqrt{0.01554 + d} / 2.755) \times $ \\ %
    & $ (0.02624 \cos(-0.9832 (0.3700 + d)) \tanh(-0.2725 + \text{round}(0.2006 m_2)) + d) $ \\[5pt] %
    & $ (\sqrt{0.01515 + d} / 2.755) \times $ \\ %
    & $ (0.02625 \cos(-0.9835 (0.3613 + d)) \tanh(-0.2965 + \text{round}(0.1732 m_2)) +d ) + $ \\ %
    & $ 0.0002095 $ \\ %
\midrule
\multirow{7}{*}{PySR}
    & $ d $ \\[5pt] %
    & $ d / 0.5553 $ \\[5pt] %
    & $ 0.06912 d^2 $ \\[5pt] %
    &  $\pmb{ \sqrt{|0.1319 d^3|} }$ \\[5pt] %
    & $ d \sqrt{|d / (-7.583 + (\cos(d) - 0.3240) 0.02758)|} $ \\[5pt] %
    & $ d \sqrt{|(d + 0.006866) / (-7.583 + (\cos(d) - 0.3240 / 0.8472) 0.02758)|} $ \\[5pt] %
    & $ d \sqrt{|(d + 0.006866) / (-7.583 + (\cos(d + 0.006866) -0.3240 / 0.8472) 0.02758)|} $ \\ %
\midrule
\multirow{2}{*}{Bayesian}
    & $ 13.35 $ \\[5pt] %
    & $ 1.342^2 d $ \\[5pt] %
    & $ (d / (1.961 m_1))^{1.499} $ \\[5pt] %
    &  $\pmb{ (0.5099 d)^{1.499} }$ \\ %
\bottomrule 
\end{tabular}
\caption{
Pareto front expressions returned by the codes AI Feynman, TuringBot, PySR, and Bayesian Machine Scientist for the Kepler solar dataset. The best candidate solution for each system is marked with bold font. 
}
\label{tab:pareto_kepler_solar}
\end{table*}

\clearpage

\begin{table*}[h]
\centering
\begin{tabular}{ll} 
\toprule
Software & Candidate formula $p=$ \\ %
\midrule
\multirow{4}{*}{AI Feynman}
& $ \sqrt{0} (-666.0 \sin{\pi}) (-666.0 + \sin{\pi}) $ \\[5pt] %
& $ \pmb{\arcsin(0.0003290+d/(m_1/d+1+1)) }$ \\[5pt] %
& $ \arcsin(-0.3644 d (-\sqrt{d/m_1})) $ \\ %
\midrule
\multirow{16}{*}{TuringBot}
& $ d $ \\[5pt] %
& $ 1.801 d $ \\[5pt] %
& $ 0.04182 d (d + 17.79) $ \\[5pt] %
& $\pmb{0.3632 d \sqrt{d}}$ \\[5pt] %
& $ d \sqrt{0.1319 d / m_1} $ \\[5pt] %
& $ 1.724 d \sqrt(0.04435 (d + 0.05064 m_1 \cos(d)) / m_1) $ \\[5pt] %
& $ 1.724 d \sqrt{0.04433 (d + 0.04998 m_1 (\cos(d)+0.2986)) / m_1} $ \\[5pt] %
& $ 1.724 d \times $ \\ %
& $\sqrt{0.04431 (d + 0.05476 (0.9761 \mod (-0.06163+d)) (\cos(d) + 0.4476)) / m_1} $ \\[5pt] %
& $ 1.724 d \sqrt{A} $, where $A = 0.04432 / m_1 \times$ \\ %
& $ (d + 0.05497 (1.002 \mod (-0.2991+d)) m_1 (\cos(d + 0.05343) + 0.4768)) $  \\[5pt] %
& $ 1.724 d \sqrt{A} $, where $A = 0.04431 / m_1 \times$ \\ %
& $ (d + 0.05434 (1.012 \mod (-0.3237+d)) (m_1 + m_1 - 1.003) \times $ \\
& $(\cos((d + 0.04387) / \tanh(1.139 d)) + 0.4895))$ \\ %
\midrule
\multirow{14}{*}{PySR}
& $ 0.03727 $ \\[5pt] %
& $ d^2 $ \\[5pt] %
& $ (\sin(d))^2 $ \\[5pt] %
& $ \sqrt{|0.3989 d^3|} $ \\[5pt] %
& $ \pmb{ \sqrt{|-0.1309 d^3 / m_1|} } $ \\[5pt] %
& $ \sqrt{|-0.1309 d^3 / (m_1 + m_2^2)|} $ \\[5pt] %
& $ \sqrt{|-0.1255 d^3 / m_1|} (1.035 - m_2) $ \\[5pt] %
& $ \sqrt{|\sin(\sin(-0.4373 d^3 / (- m_1 / 0.5873) (m_2 + -0.5244)))|} $ \\[5pt] %
& $ \sqrt{|-0.4373 d^3 / (m_1 / -0.5873) (m_2 d / sin(d) - 0.5244)|} $ \\ %
\midrule
\multirow{7}{*}{Bayesian}
& $ m_2 $ \\[5pt] %
& $ 0.4825 (-0.02983 + \sin(d)) $ \\[5pt] %
& $ (-0.3064^3 + d) 0.4613 $ \\[5pt] %
& $ (d^2 + 0.3411 d) 0.5240 $ \\[5pt] %
& $ d (0.1650 + d / (m_1 / 0.1815)) $  \\[5pt] %
&  $ \pmb{ 0.3641 m_1^{-0.49281} d^{(4.0981 + m_2) 0.3641}
 }$ \\ %
\bottomrule 
\end{tabular}
\caption{
Pareto front expressions returned by the codes AI Feynman, TuringBot, PySR, and Bayesian Machine Scientist for the Kepler exoplanet dataset. The best candidate solution for each system is marked in bold font. 
}
\label{tab:pareto_kepler_exoplanet}
\end{table*}

\clearpage

\begin{table*}[h]
\centering
\begin{tabular}{ll}
\toprule
Software & Candidate formula $p=$ \\ %
\midrule
\multirow{4}{*}{AI Feynman}
& $ 1/(666.0 + \pi / \sin{\pi}) $ \\[5pt] %
& $ \pmb{\arcsin(1.282+(\cos(\cos((\pi-\sqrt{m_2})/d))+m_1) \times }$ \\ %
& $ \pmb{\sin(e^{(\sin(\cos(\cos((\pi-\sqrt{m_2})/d))+m_1+\cos(\cos((\pi-\sqrt{m_2})/d))+m_1))^{-1}})) } $ \\ %
\midrule
\multirow{11}{*}{PySR}
& $ 709.4 $ \\[5pt] %
& $ d / 0.1362 $  \\[5pt] %
&  $ \pmb{ \sqrt{|d^3 / (m_1 + m_2)|} }$ \\[5pt] %
& $ \sqrt{|d^3 / (m_1 + m_2)|} + m_1 $ \\[5pt] %
& $ \sqrt{|-0.9276 d^3 / (0.8620 m_1 + m_2) |} $ \\[5pt] %
& $ \sqrt{|(d + 0.2389)^3 / 1.096 / (m_2 + 0.8451 m_1) |} $  \\[5pt] %
& $ \sqrt{| (d + 0.2070)^3 / 1.093 / |m_2 + 0.8485 m_1| |} $  \\[5pt] %
& $ 0.9852 \sqrt{| d^3 / (m_1 + \sqrt{|m_1 + -0.3363|}) |} - 5.460 + m_1 - 5.491 $ \\ %
\midrule
\multirow{4}{*}{Bayesian}
& $ d $ \\[5pt] %
& $ 4.302 ((-26.43 m_1 + d)^{1.035} + d / (m_1 m_2)) m_2 $ \\[5pt] %
& $\pmb{ (d (m_2 + m_1 - 0.006465 (m_2^3 + m_1^{m_2}) )^{-0.3340})^{1.499} }$ \\ %
\bottomrule 
\end{tabular}
\caption{
Pareto front expressions returned by the codes AI Feynman, PySR, and Bayesian Machine Scientist for the Kepler binary stars dataset. The best candidate solution is marked with bold font. }
\label{tab:pareto_kepler_binarystars}
\end{table*}

\clearpage

\begin{table*}[h]
\centering
\begin{tabular}{ll}
\toprule
Software & Candidate formula $y=$ \\ %
\midrule
\multirow{7}{*}{AI Feynman}
& $ \arccos(2.909+(\cos{\pi}-1)) $ \\[5pt] %
& $ \sqrt{0.04-0.02 v} $ \\[5pt] %
&  $\pmb{ -0.05122 v }$ \\[5pt] %
& $ -1.018 + \sin( e^{ e^{ \sin( \sin(v-1-1)) } } ) $ \\[5pt] %
& $ -0.02009 e^{ (\sqrt{v} + \sin((v-1)^{-1}) ) } $ \\ %
\midrule
\multirow{22}{*}{TuringBot}
& $ -2.471 $ \\[5pt] %
& $ -0.1675 v $ \\
\bottomrule 
\end{tabular}
\caption{
Pareto front expressions returned by the codes AI Feynman, TuringBot, PySR, and Bayesian Machine Scientist for the time dilation dataset. The best candidate solution for each system is marked with bold font.
}
\label{tab:pareto_time_dilation}
\end{table*}

\clearpage

\begin{table*}[h]
\centering
\begin{tabular}{ll}
\toprule
Software & Candidate formula $q=$ \\ %
\midrule
\multirow{6}{*}{AI Feynman}
& $ 1/ 0.03000 $ \\[5pt] %
& $ 38.50 + p $ \\[5pt] %
&  $\pmb{ 30.60 + p + p }$ \\[5pt] %
& $ \arcsin(0.8065 \sin(e^{p+p} - p)) $ \\[5pt] %
& $ 27.10(\log(p+p+1+1+1)-1) $ \\ %
\midrule
\multirow{14}{*}{TuringBot}
& $ 67.01 $ \\[5pt] %
& $\pmb{ 19.45 \arccosh(p) }$  \\[5pt] %
& $ (1.570+p)/(0.1128+0.008879p) $  \\[5pt] %
& $ (\cos(\text{round}(-9.038p))+19.94) \arcsinh(\lfloor p \rfloor) $ \\[5pt] %
& $ (\cos(\text{round}(-9.038p))+19.90)\arcsinh(\lfloor p \rfloor + 0.09014) $ \\[5pt] %
& $ (\cos(\text{round}(-9.038p))+19.86)\arcsinh(\lfloor 1.014 p \rfloor + 0.1024) $ \\[5pt] %
& $ 5.349(0.4604-\cos(0.09345(\tan(-0.1359 p)-1.649- \lceil p \rceil)))+ \arcsinh(p) / 0.05244 $ \\[5pt] %
& $ 5.425(0.5252-\cos(0.09347(0.7537(\text{round}(\tan(-0.1360(p-0.38043)))-$ \\ %
& $2.036)- \lceil p \rceil)))+ \arcsinh(p) / 0.05269 $ \\[5pt] %
& $ 19.74 \arccosh(-0.3987+p) + \sqrt{0.8632(p+1.820)} \times$ \\ %
& $ \cos(-0.7427(4.618/(\tan(0.06620+p)+p-1.738 \cos(p))-(-0.8143 + \text{round}(p)))) $ \\ %
\midrule
\multirow{11}{*}{PySR}
& $ 67.01 $ \\[5pt] %
& $\pmb{ 23.50 \log(|p|) }$ \\[5pt] %
& $ 20.07 (\log(|p|) + 0.5671) $ \\[5pt] %
& $ 65.87 \log(|\log(|p + 1.991|)|) $ \\[5pt] %
& $ 66.09 \log(|\log(|p + \log(|11.97 - p|)|)|) $ \\[5pt] %
& $ 19.54 (\log(|p|) + \cos( \sin( \log(|(p - 8.270) / 0.8487|) ) ) ) $ \\[5pt] %
& $ 19.54 (\log(|p|) + \cos( \sin( \log(|-1.175 (p - 8.270 + 0.1542)|) ) ) )
 $ \\[5pt] %
& $ 19.53 (\log(|p|) + \cos( \sin( \log(|(|p - 8.320 / 1.048| + 0.2113) / 20.07|) ) ) ) $ \\ %
\midrule
\multirow{5}{*}{Bayesian}
& $ 67.01 $ \\[5pt] %
& $ (200865.8 p)^{0.2770} $ \\[5pt] %
& $ ((12.71 p + 19.97) / 0.1128) / (p + 12.71) $ \\[5pt] %
&  $\pmb{ (112.6 p + 177.0) / (p + 12.71) }$ \\ %
\bottomrule 
\end{tabular}
\caption{
Pareto front expressions returned by the codes AI Feynman, TuringBot, PySR, and Bayesian Machine Scientist for the Langmuir \cite[Table IX]{Langmuir1918} dataset. The best candidate solution for each system is marked with bold font.
}
\label{tab:pareto_langmuir_tableIX}
\end{table*}
\newpage

\begin{table*}[h]
\centering
\begin{tabular}{ll}\toprule
Software & Candidate formula $q=$ \\ %
\midrule
\multirow{7}{*}{AI Feynman}
& $ (-666.0 \sin{\pi})^2 $ \\[5pt] %
& $ \arcsin(0.03p+0.72) $ \\[5pt] %
& $\pmb{ 0.06p+0.79 }$ \\[5pt] %
& $ -0.1042+\sqrt{\sqrt{\sqrt{\sqrt{p}}}} $ \\[5pt] %
& $ 1.615 \sin(\log(\log(p+\pi+1+1))) $ \\ %
\midrule
\multirow{17}{*}{TuringBot}
& $ 1.155 $ \\[5pt] %
& $ p^{0.1177} $ \\[5pt] %
&  $\pmb{ p/(p+8.057)+0.7518 }$ \\[5pt] %
& $ 1.615-0.9128^{2.6189+p-0.1221/p} $ \\[5pt] %
& $ \arctan(0.8093+0.1852p)+0.1238-0.006844/(\tan(p)-0.09319p) $ \\[5pt] %
& $ \arctan(0.7984+0.1826p)+0.1258-0.004073/(\tan(p)-0.09197p-0.02751) $ \\[5pt] %
& $ 0.001026(-24.31+p)+ $ \\ %
& $ \arctan((18.68(-0.06174+p))^{\log(1.054+0.02370(\cos(-1.124(-0.7298+p))+p))} ) $ \\[5pt] %
& $ 0.001052(-25.27+p)+$ \\ %
& $ \arctan((17.83(-0.05962+p))^{\log(1.056+0.02719(\cos(1.125(-0.6507+p))+0.8783p))} ) $ \\[5pt] %
& $ 0.001067((0.4667-0.01580p) \tan(p) -24.73+p) + $ \\ %
& $ \arctan((16.71(-0.05784+p))^{\log(1.059+0.02732(\cos(1.087(-0.5273+p))+0.8647p))} ) $ \\[5pt] %
& $ 0.001057(\cos(p)+(0.4217-0.01331p) \tan(p)-24.46+p) + $ \\ %
& $ \arctan((16.09(-0.05778+p))^{\log(1.060+0.02710(\cos((-0.5464+p)/(-0.9153))+0.8754p))} ) $ \\ %
\midrule
\multirow{12}{*}{PySR}
& $ 1.155 $ \\[5pt] %
& $\pmb{ \sqrt{| \sqrt{| \sqrt{|p|} |} |} }$  \\[5pt] %
& $ \sqrt{|\sqrt{|\sqrt{|p|}|}|} -0.05434 $ \\[5pt] %
& $ \sqrt{| \sqrt{| \sqrt{|p|} - \sin(0.3785 p) |} |} $ \\[5pt] %
& $ \sqrt{| \sqrt{| \sqrt{|p|} - ( \sin( \log(|p + 0.9549|) ) )^2 |} |} $ \\[5pt] %
& $ \sqrt{| \sqrt{|  \sqrt{|p|} - ( \sin( \sqrt{|p / 1.495|} ) )^3  - (0.01481 p)^2 |} |} $ \\[5pt] %
& $ \sqrt{| \sqrt{| \sqrt{|p|} - ( \sin( \sqrt{|p / -1.413|} ) )^3 / 1.145 - (-0.01478 p)^2 |} |} $ \\ %
\midrule
\multirow{2}{*}{Bayesian}
& $ 1.155 $ \\[5pt] %
&  $\pmb{ p^{0.1177} }$ \\ %
\bottomrule 
\end{tabular}
\caption{
Pareto front expressions returned by the codes AI Feynman, TuringBot, PySR, and Bayesian Machine Scientist for the Sun et al. \cite[Table 1]{sunetal1998} dataset. The best candidate solution for each system is marked with bold font.
}
\label{tab:pareto_langmuir_sunetal}
\end{table*}

\clearpage

\new{In Table~\ref{tab:ai-feynman-results}, we show the results for AI-Descartes, AI-Feynman, PySR and BMS on 81 problems (problems that do not contain trigonometric functions
) from the FSRD (Feynman Symbolic Regression Database) using our version of the data (fewer data point and with noise) for each problem.}

\new{For each solver, we check if the list of output solutions contains the true formula, which is indicated by a \checkmark in the appropriate column.  
If the functional form is correct but has a single multiplicative constant that is slightly different from the true constant, we indicate this by \checkmark$^1$. As an example, instead of obtaining $q_2E_f$ for I.12.5, we obtain $0.99679 q_2E_f$ as a potential solution. 
If a constant that is added to other terms is slightly incorrect, we indicate this by \checkmark$^2$. For example, PySR obtains $p_FV/(\gamma - 1.0014104)$ instead of $p_FV/(\gamma - 1)$ as a solution for problem I.39.11. 
Finally, if a formula is correct other than for a slightly erroneous power, we indicate this by a \checkmark$^3$; for example, in the solution returned by Bayesian Machine Scientist to II.3.24, the power of $r$ is 1.98285 instead of 2. 
It is clear that all of the solvers struggle with many of the problems, especially those that are not a simple ratio of monomials. Thus our version of FSRD dataset - which is designed to mimic real-life data in that we have both few data points and a nontrivial amount of noise - is quite nontrivial to deal with.
Our tool, AI-Descartes, is the best performing, achiving an accuracy of 60.49\%, obtaining the correct solution for 49 out of 81 problems as a candidate solution modulo a small error in a multiplicative constant or in an addition, compared to 40.74\% for AI Feynman, 49.38\% for PySR, and 48.15\% for Bayesian Machine Scientist (BMS). Therefore our method outperform the next best solver (PySR) with an improvement of  11.11\%. This gap in performance is the reason why we choose to use our symbolic regression engine instead of other methods.
}

\new{In Table~\ref{tab:ai-feynman-results_turingbot}, we provide the  results on a subset of 15 problems from the ones used in  Table~\ref{tab:ai-feynman-results}, to compare with TuringBot. We determined this subset to comply with the requirements of the freely available version of TuringBot which allows a maximum of 2 variables per problem.
Also in this subset of problems, we outperform all the state-of-the-art methods. However, the performance of the tools is generally higher in Table~\ref{tab:ai-feynman-results_turingbot}, compared to  Table~\ref{tab:ai-feynman-results}, since these selected 15 problems are much simpler and easier to solve.}

\new{Finally, we manually extracted the background theory\footnote{For simplicity, and without loss of generality, we considered all the variables greater than 0.} of 5 problems from Feynman's Lectures ({I.27.6}, {I.34.8}, {I.43.16}, {II.10.9} and {II.34.2}). We were able to execute the full AI-Descartes pipeline: The results are provided in Table~\ref{tab:sota_reasoning}. For all of these 5 problems the SR module is able to induce the right formula (as showed in Table~\ref{tab:ai-feynman-results}) and the Reasoning module is able to derive it, either directly or with existential quantification over the numerical constants. This result shows that for some problems from FSRD, the reasoning module is able to successfully identify the correct formula from a set of candidates with comparable errors over the data.}

\new{
In conclusion, the tools we compared against are unable to consistently identify the correct formula on the six datasets (obtained from real experiments) considered in this work, or on the 81 synthetic datasets derived from the Feynman Symbolic Regression Database. Although some of them produce the correct formula among the top candidates, they do not have a principled way to identify it from among the list of candidates.}

\clearpage

{\small
\begin{longtable}{llcccc}
\toprule
\bf Label      & \bf  Formula & \bf  AI Descartes	& \bf 	AI Feynman	&\bf 	PySR	&\bf 	BMS	\\
\midrule
I.6.20a  	&	 $e^{-\theta^2/2}/\sqrt{2 \pi }$               	&	{\sffamily X}\phantom{$^0$}	& {\sffamily X}\phantom{$^0$}	&	{\sffamily X}\phantom{$^0$}	&	{\sffamily X}\phantom{$^0$}	\\
I.6.20  	&	 $e^{-\frac{\theta^2}{2\sigma^2}}/\sqrt{2 \pi \sigma^2}$	&	{\sffamily X}\phantom{$^0$}	&	{\sffamily X}\phantom{$^0$}	&	{\sffamily X}\phantom{$^0$}	&	{\sffamily X}\phantom{$^0$}	\\
I.6.20b 	&	 $e^{-\frac{(\theta-\theta_1)^2}{2\sigma^2}}/\sqrt{2 \pi \sigma^2}$ 	&	{\sffamily X}\phantom{$^0$}	&	{\sffamily X}\phantom{$^0$}	&	{\sffamily X}\phantom{$^0$}	&	{\sffamily X}\phantom{$^0$}	\\[2pt]
I.8.14    	&	 $\sqrt{(x_2-x_1)^2+(y_2-y_1)^2}$          	&		{\sffamily X}\phantom{$^0$}	&	{\sffamily X}\phantom{$^0$}	&	{\sffamily X}\phantom{$^0$} &	{\sffamily X}\phantom{$^0$}	\\
I.9.18    	&	 $\frac{Gm_1m_2}{(x_2-x_1)^2+(y_2-y_1)^2+(z_2-z_1)^2}$  	&	{\sffamily X}\phantom{$^0$}	&	{\sffamily X}\phantom{$^0$}	&	{\sffamily X}\phantom{$^0$}	&	{\sffamily X}\phantom{$^0$}\\
I.10.7   	&	 $\frac{m_0}{\sqrt{1-v^2/c^2}}$             	&	\checkmark\phantom{$^0$}	&		{\sffamily X}\phantom{$^0$}	&	{\sffamily X}\phantom{$^0$}	&	{\sffamily X}\phantom{$^0$}	\\
I.11.19   	&	 $x_1y_1+x_2y_2+x_3y_3$                     	&	{\sffamily X}\phantom{$^0$} & {\sffamily X}\phantom{$^0$}	&	{\sffamily X}\phantom{$^0$}	&	{\sffamily X}\phantom{$^0$}	\\
I.12.1    	&	 $\mu N_n$                                  	&	\checkmark\phantom{$^0$}	&	$\checkmark^2$	&	\checkmark\phantom{$^0$}	&	\checkmark\phantom{$^0$}	\\
I.12.2    	&	 $q_1q_2 / (4\pi \varepsilon r^2)$     	&	$\checkmark^1$	&	X	&	$\checkmark^1$	&	$\checkmark^1$	\\
I.12.4    	&	 $q_1 / (4\pi \varepsilon r^2)$        	&	$\checkmark^1$	&	$\checkmark^1$	&	$\checkmark^1$	&	$\checkmark^1$	\\
I.12.5    	&	 $q_2 E_f$                                	&	$\checkmark^1$	&	$\checkmark^2$	&	 \checkmark\phantom{$^0$} 	&	 \checkmark\phantom{$^0$} 	\\
I.13.4    	&	 $\frac{1}{2}m(v^2+u^2+w^2)$               	&	{\sffamily X}\phantom{$^0$}	&	{\sffamily X}\phantom{$^0$}	&	{\sffamily X}\phantom{$^0$}	&	{\sffamily X}\phantom{$^0$}\\
I.13.12   	&	 $Gm_1m_2(\frac{1}{r_2}-\frac{1}{r_1})$    	&	{\sffamily X}\phantom{$^0$}	&	{\sffamily X}\phantom{$^0$}	&	{\sffamily X}\phantom{$^0$}	&	{\sffamily X}\phantom{$^0$}\\
I.14.3    	&	 $mgz$                                	&	 \checkmark\phantom{$^0$} 	&	\checkmark$^1$	&	 \checkmark\phantom{$^0$} 	&	 \checkmark\phantom{$^0$} 	\\
I.14.4    	&	 $k_{spring}x^2/2$                      	&	\checkmark$^1$	&	\checkmark$^1$	&	\checkmark$^1$	&	\checkmark$^1$	\\
I.15.3x   	&	 $\frac{x-ut}{\sqrt{1-u^2/c^2}}$       	&	{\sffamily X}\phantom{$^0$}	&	{\sffamily X}\phantom{$^0$}	&	{\sffamily X}\phantom{$^0$}	&	{\sffamily X}\phantom{$^0$}\\
I.15.3t  	&	 $\frac{t-ux/c^2}{\sqrt{1-u^2/c^2}}$       &	{\sffamily X}\phantom{$^0$}	&	{\sffamily X}\phantom{$^0$}	&	{\sffamily X}\phantom{$^0$}	&	{\sffamily X}\phantom{$^0$}\\
I.15.10  	&	 $\frac{m_0v}{\sqrt{1-v^2/c^2}}$            	&	\checkmark$^1$	&	{\sffamily X}\phantom{$^0$}	&	{\sffamily X}\phantom{$^0$}	&	{\sffamily X}\phantom{$^0$}	\\
I.16.6    	&	 $\frac{u+v}{1+uv/c^2}$                   	&	\checkmark$^2$	&	{\sffamily X}\phantom{$^0$}	&	{\sffamily X}\phantom{$^0$}	&	{\sffamily X}\phantom{$^0$}	\\
I.18.4   	&	 $\frac{m_1r_1+m_2r_2}{m_1+m_2}$            	&	 \checkmark\phantom{$^0$} 	&	{\sffamily X}\phantom{$^0$}	&	{\sffamily X}\phantom{$^0$}	&	{\sffamily X}\phantom{$^0$}	\\[2pt]
I.24.6    	&	 $\frac{1}{4}m(\omega^2 + \omega_0^2)x^2$  &	{\sffamily X}\phantom{$^0$}	&	{\sffamily X}\phantom{$^0$}	&	{\sffamily X}\phantom{$^0$}	&	{\sffamily X}\phantom{$^0$}\\
I.25.13   	&	 $q/C$                                    	&	 \checkmark\phantom{$^0$} 	&	\checkmark$^1$	&	 \checkmark\phantom{$^0$} 	&	 \checkmark\phantom{$^0$} 	\\
I.27.6    	&	 $1 / (\frac{1}{d_1}+\frac{n}{d_2})$  	&	\checkmark$^1$	&	\checkmark$^2$	&	 \checkmark\phantom{$^0$} 	&	{\sffamily X}\phantom{$^0$}	\\
I.29.4  	&	$\frac{\omega}{C}$	&	 \checkmark\phantom{$^0$} 	&	\checkmark$^2$	&	 \checkmark\phantom{$^0$} 	&	 \checkmark\phantom{$^0$} 	\\
I.32.5   	&	 $q^2a^2/(6\pi \varepsilon c^3)$     &	{\sffamily X}\phantom{$^0$}	&	{\sffamily X}\phantom{$^0$}	&	{\sffamily X}\phantom{$^0$}	&	{\sffamily X}\phantom{$^0$}\\
I.32.17   	&	 $(\frac{1}{2} \epsilon c E^2_f)(\frac{8\pi r^2}{ 3})\big(\frac{\omega^4}{(\omega^2 - \omega^2_0)^2}\big)$                                  &	{\sffamily X}\phantom{$^0$}	&	{\sffamily X}\phantom{$^0$}	&	{\sffamily X}\phantom{$^0$}	&	{\sffamily X}\phantom{$^0$}\\
I.34.8   	&	 $q v B / p$                                   	&	 \checkmark\phantom{$^0$} 	&	\checkmark$^1$	&	 \checkmark\phantom{$^0$} 	&	\checkmark$^1$	\\
I.34.10   	&	 $\omega_0 / (1-v/c)$                 	&	 \checkmark\phantom{$^0$} 	&	\checkmark$^1$	&	 \checkmark\phantom{$^0$} 	&	{\sffamily X}\phantom{$^0$}	\\
I.34.14   	&	 $\frac{1+v/c}{1-v^2/c^2}\omega_0$        &	{\sffamily X}\phantom{$^0$}	&	{\sffamily X}\phantom{$^0$}	&	{\sffamily X}\phantom{$^0$}	&	{\sffamily X}\phantom{$^0$}\\
I.34.27	    &	 $\frac{h \omega}{2 \pi}$	&	\checkmark$^1$	&	\checkmark$^1$	&	\checkmark$^1$	&	\checkmark$^1$	\\
I.38.12   	&	 $4 \pi \epsilon h^2 / (m q^2)$                 	&	\checkmark$^1$	&		&	\checkmark$^1$	&	\checkmark$^2$	\\
I.39.10   	&	 $\frac{3}{2} p_F V$                 	&	\checkmark$^1$	&	\checkmark$^1$	&	\checkmark$^1$	&	\checkmark$^1$	\\
I.39.11   	&	 $\frac{1}{\gamma -1}p_FV$                 	&	\checkmark$^2$	&	\checkmark$^1$	&	\checkmark$^2$	&	\checkmark$^2$	\\
I.39.22   	&	 $n k_b T / V$              	&	 \checkmark\phantom{$^0$} 	&	\checkmark$^1$	&	 \checkmark\phantom{$^0$} 	&	 \checkmark\phantom{$^0$} 	\\
I.40.1   	&	 $n_0 e^{-\frac{m g x}{k_b T}}$                 &	{\sffamily X}\phantom{$^0$}	&	{\sffamily X}\phantom{$^0$}	&	{\sffamily X}\phantom{$^0$}	&	{\sffamily X}\phantom{$^0$}\\
I.41.16   	&	 $\frac{h \omega^3}{\pi^2 c^2 (e^{\frac{h \omega}{k_b T}} - 1)}$                &	{\sffamily X}\phantom{$^0$}	&	{\sffamily X}\phantom{$^0$}	&	{\sffamily X}\phantom{$^0$}	&	{\sffamily X}\phantom{$^0$}\\[2pt]
I.43.16   	&	 $\mu_{drift} q V_e / d$              	&	\checkmark$^1$	&	\checkmark$^2$	&	 \checkmark\phantom{$^0$} 	&	 \checkmark\phantom{$^0$} 	\\
I.43.31   	&	 $\mu_e k_b T$              	&	 \checkmark\phantom{$^0$} 	&	\checkmark$^1$	&	 \checkmark\phantom{$^0$} 	&	\checkmark$^2$	\\
I.43.43   	&	 $\frac{k_bv}{(\gamma -1)A}$               	&		{\sffamily X}\phantom{$^0$} &	\checkmark$^1$	&	{\sffamily X}\phantom{$^0$}	&	{\sffamily X}\phantom{$^0$}	\\
I.44.4	    &	$n k_b T \ln{\frac{V2}{V1}}$&	{\sffamily X}\phantom{$^0$}	&	{\sffamily X}\phantom{$^0$}	&	{\sffamily X}\phantom{$^0$}	&	{\sffamily X}\phantom{$^0$}\\
I.47.23   	&	 $\sqrt{\frac{\gamma p r}{\rho}}$               	&	\checkmark$^2$	&	\checkmark$^1$	&	 \checkmark\phantom{$^0$} 	&	\checkmark$^2$	\\
I.48.20   	&	 $\frac{mc^2}{\sqrt{1-v^2/c^2}}$          	&	 \checkmark\phantom{$^0$} 	&	{\sffamily X}\phantom{$^0$}	&	{\sffamily X}\phantom{$^0$}	&	{\sffamily X}\phantom{$^0$}	\\
II.2.42	&	 $k(T_2-T_1)A/d$                  	&	{\sffamily X}\phantom{$^0$}	&	{\sffamily X}\phantom{$^0$}	&	 \checkmark\phantom{$^0$} 	&	{\sffamily X}\phantom{$^0$}	\\
II.3.24    	&	 $P / (4 \pi r^2)$                                  	&	\checkmark$^1$	&	\checkmark$^1$	&	\checkmark$^1$	&	\checkmark$^3$	\\
II.4.23    	&	 $q / (4 \pi \epsilon r)$                                  	&	\checkmark$^1$	&	\checkmark$^1$	&	\checkmark$^1$	&	\checkmark$^1$	\\
II.6.15a    	&	 $\frac{3}{4 \pi \epsilon}\frac{p_d z}{r^5}\sqrt{x^2 + y^2}$                                 &	{\sffamily X}\phantom{$^0$}	&	{\sffamily X}\phantom{$^0$}	&	{\sffamily X}\phantom{$^0$}	&	{\sffamily X}\phantom{$^0$}\\
II.8.7      	&	 $\frac{3}{5} \frac{q^2}{4 \pi \epsilon d}$                                 	&	\checkmark$^1$	&	\checkmark$^1$	&	\checkmark$^1$	&	\checkmark$^2$	\\
II.8.31     	&	 $\epsilon E^2_f / 2$                                 	&	\checkmark$^1$	&	\checkmark$^1$	&	\checkmark$^1$	&	\checkmark$^1$	\\
II.10.9     	&	 $\frac{\sigma_{den}}{\epsilon}\frac{1}{1 + \chi}$                                 	&	\checkmark$^1$	&	\checkmark$^2$	&	\checkmark$^2$	&	\checkmark$^2$	\\
II.11.3   	&	 $\frac{qE_f}{m(\omega_0^2-\omega^2)}$    &	{\sffamily X}\phantom{$^0$}	&	{\sffamily X}\phantom{$^0$}	&	{\sffamily X}\phantom{$^0$}	&	{\sffamily X}\phantom{$^0$}\\
II.11.20    	&	 $n_{\rho}p^2_dE_f / (3k_bT)$        	&	\checkmark$^1$	&	{\sffamily X}\phantom{$^0$}	&	{\sffamily X}\phantom{$^0$}	&	\checkmark$^2$	\\
II.11.27  	&	 $\frac{n\alpha}{1-n\alpha/3}\epsilon E_f$ 	&	\checkmark$^1$	&	{\sffamily X}\phantom{$^0$}	&	{\sffamily X}\phantom{$^0$}	&	{\sffamily X}\phantom{$^0$}	\\
II.11.28	&	 $1+\frac{n\alpha}{1-n\alpha/3}$          	&	\checkmark$^1$	&	{\sffamily X}\phantom{$^0$}	&	{\sffamily X}\phantom{$^0$}	&	{\sffamily X}\phantom{$^0$}	\\
II.13.17    	&	 $\frac{1}{4 \pi \epsilon c^2}\frac{2 I}{r}$                                 	&	\checkmark$^1$	&	{\sffamily X}\phantom{$^0$}	&	\checkmark$^1$	&	{\sffamily X}\phantom{$^0$}	\\
II.13.23    	&	 $\frac{\rho_{c_0}}{\sqrt{1 - v^2 / c^2}}$                                 	&	\checkmark$^1$	&	{\sffamily X}\phantom{$^0$}	&	{\sffamily X}\phantom{$^0$}	&	{\sffamily X}\phantom{$^0$}	\\
II.13.34    	&	 $\frac{\rho_{c_0} v}{\sqrt{1 - v^2 / c^2}}$                                 &	{\sffamily X}\phantom{$^0$}	&	{\sffamily X}\phantom{$^0$}	&	{\sffamily X}\phantom{$^0$}	&	{\sffamily X}\phantom{$^0$}\\
II.21.32    	&	 $\frac{q}{4\pi \varepsilon r(1-v/c)} $ &	{\sffamily X}\phantom{$^0$}	&	{\sffamily X}\phantom{$^0$}	&	{\sffamily X}\phantom{$^0$}	&	{\sffamily X}\phantom{$^0$}\\
II.24.17  	&	 $\sqrt{\frac{\omega^2}{c^2}-\frac{\pi^2}{d^2}}$ &	{\sffamily X}\phantom{$^0$}	&	{\sffamily X}\phantom{$^0$}	&	{\sffamily X}\phantom{$^0$}	&	{\sffamily X}\phantom{$^0$}\\
II.27.16    	&	 $\epsilon c E^2_f$                                 	&	 \checkmark\phantom{$^0$} 	&	{\sffamily X}\phantom{$^0$}	&	 \checkmark\phantom{$^0$} 	&	 \checkmark\phantom{$^0$} 	\\
II.27.18    	&	 $\epsilon E^2_f$                                 	&	\checkmark$^1$	&	\checkmark$^1$	&	 \checkmark\phantom{$^0$} 	&	 \checkmark\phantom{$^0$} 	\\
II.34.2a    	&	 $q v / (2 \pi r)$                                 	&	\checkmark$^1$	&	\checkmark$^1$	&	\checkmark$^1$	&	\checkmark$^1$	\\
II.34.2      	&	 $q v r / 2$                                	&	\checkmark$^1$	&	\checkmark$^1$	&	\checkmark$^1$	&	\checkmark$^1$	\\
II.34.11    	&	 $gqB/(2m)$                     	&	 \checkmark\phantom{$^0$} 	&	\checkmark$^1$	&	\checkmark$^1$	&	\checkmark$^1$	\\
II.34.29a    	&	 $q h / (4 \pi m)$                                	&	\checkmark$^1$	&	\checkmark$^1$	&	\checkmark$^1$	&	\checkmark$^1$	\\
II.34.29b    	&	 $g \mu_M B J_z / h$                                	&	\checkmark$^2$	&	{\sffamily X}\phantom{$^0$}	&	\checkmark$^1$	&	\checkmark$^1$	\\
II.35.18	&	$\frac{n_0}{e^{\frac{mom \times B}{kb \times T}}+e^{\frac{-mom \times B}{(kb \times T)}}}$&	{\sffamily X}\phantom{$^0$}	&	{\sffamily X}\phantom{$^0$}	&	{\sffamily X}\phantom{$^0$}	&	{\sffamily X}\phantom{$^0$}\\
II.36.38    	&	 $\frac{\mu_m B}{k_bT} + \frac{\mu_m\alpha M}{\varepsilon c^2k_bT}$  &	{\sffamily X}\phantom{$^0$}	&	{\sffamily X}\phantom{$^0$}	&	{\sffamily X}\phantom{$^0$}	&	{\sffamily X}\phantom{$^0$}\\
II.37.1    	&	 $\mu_M(1+\chi)B$                  	&	\checkmark$^1$	&	{\sffamily X}\phantom{$^0$}	&	\checkmark$^1$	&	\checkmark$^2$	\\
II.38.3   	&	 $YAx / d$                    	&	\checkmark$^1$	&	\checkmark$^2$	&	 \checkmark\phantom{$^0$} 	&	\checkmark$^2$	\\
II.38.14     	&	 $\frac{Y}{2 (1 + \sigma)}$                                	&	\checkmark$^1$	&	\checkmark$^2$	&	{\sffamily X}\phantom{$^0$}	&	\checkmark$^2$	\\
III.4.32     	&	 $1 / (e^{\frac{h \omega}{k_b T}}-1)$                                &	{\sffamily X}\phantom{$^0$}	&	{\sffamily X}\phantom{$^0$}	&	{\sffamily X}\phantom{$^0$}	&	{\sffamily X}\phantom{$^0$}\\
III.4.33     	&	 $h \omega / (e^{\frac{h \omega}{k_b T}}-1)$                                &	{\sffamily X}\phantom{$^0$}	&	{\sffamily X}\phantom{$^0$}	&	{\sffamily X}\phantom{$^0$}	&	{\sffamily X}\phantom{$^0$}\\
III.7.38     	&	 $2 \mu_M B / h$                                	&	\checkmark$^1$	&		&	\checkmark$^1$	&	\checkmark$^1$	\\
III.10.19 	&	 $\mu_M \sqrt{B_x^2+B_y^2+B_z^2}$    &	{\sffamily X}\phantom{$^0$}	&	{\sffamily X}\phantom{$^0$}	&	{\sffamily X}\phantom{$^0$}	&	{\sffamily X}\phantom{$^0$}\\
III.12.43    	&	 $n h$                                	&	\checkmark$^1$	&	\checkmark$^1$	&	\checkmark$^1$	&	\checkmark$^1$	\\
III.13.18    	&	 $2 E d^2 k / h$                                	&	\checkmark$^1$	&	{\sffamily X}\phantom{$^0$}	&	{\sffamily X}\phantom{$^0$}	&	\checkmark$^1$	\\
III.14.14    	&	 $I_0 (e^{\frac{q V_e}{k_b T}} - 1)$                                &	{\sffamily X}\phantom{$^0$}	&	{\sffamily X}\phantom{$^0$}	&	{\sffamily X}\phantom{$^0$}	&	{\sffamily X}\phantom{$^0$}\\
III.15.14    	&	 $h^2 / (2 E d^2)$                                	&	{\sffamily X}\phantom{$^0$}	&	{\sffamily X}\phantom{$^0$}	&	\checkmark$^1$	&	\checkmark$^2$	\\
III.15.27    	&	 $2 \pi \alpha / (n d)$                                	&	\checkmark$^1$	&	\checkmark$^1$	&	\checkmark$^1$	&	\checkmark$^1$	\\
III.19.51    	&	 $\frac{-m q^4}{2 (4 \pi \epsilon)^2 h^2}\frac{1}{n^2}$                                &	{\sffamily X}\phantom{$^0$}	&	{\sffamily X}\phantom{$^0$}	&	{\sffamily X}\phantom{$^0$}	&	{\sffamily X}\phantom{$^0$}\\
III.21.20    	&	 $- \rho_{c_0} q A_{vec} / m$                                	&	\checkmark$^1$	&	\checkmark$^1$	&	 \checkmark\phantom{$^0$} 	&	 \checkmark\phantom{$^0$} 	\\
\midrule
 & & \bf AI-Descartes	&	\bf AI-Feynman	& \bf	PySR	& \bf	BMS \\
 \cmidrule{3-6}
  \multicolumn{2}{l}{\new{Number of (\checkmark, \checkmark$^1$, \checkmark$^2$, \checkmark$^3$, {\sffamily X})}} & (13, 32, 4, 0, 32) & (0, 25, 8, 0, 48) & (16, 21, 2, 0, 41) & (10, 17, 11, 1, 42) \\
   \multicolumn{2}{l}{\new{Total $\checkmark^*$ }} & 49/81 & 33/81 & 40/81 & 39/81 \\ 

 \multicolumn{6}{@{}l@{}}{\dashrule}  \\
 \multicolumn{2}{l}{\new{ \bf Accuracy}} & \bf 60.49\%  & 40.74\%  & 49.38\%  & 48.15\%  \\ 
\bottomrule
\caption{\normalsize\new{Results on 81/100 problems from the Feynman Database for Symbolic Regression (problems not containing trigonometric functions
).}}
\label{tab:ai-feynman-results}
\end{longtable}
}

\begin{table*}[h] \small
\centering
\begin{tabular}{lccccc}
\toprule
\bf Label  & \bf AI-Descartes	&	\bf AI-Feynman	&	\bf PySR	&	\bf BMS & \bf TuringBot\\\midrule
I.6.20a	& {\sffamily X}\phantom{$^0$}& {\sffamily X}\phantom{$^0$}& {\sffamily X}\phantom{$^0$}& {\sffamily X}\phantom{$^0$}&	{\sffamily X}\phantom{$^0$}	\\
I.6.20  	&{\sffamily X}\phantom{$^0$} & {\sffamily X}\phantom{$^0$}& {\sffamily X}\phantom{$^0$} & {\sffamily X}\phantom{$^0$} &	{\sffamily X}\phantom{$^0$}	\\
I.12.1    	& \checkmark\phantom{$^0$}& \checkmark$^2$& \checkmark\phantom{$^0$}& \checkmark\phantom{$^0$}&	 \checkmark\phantom{$^0$} 	\\
I.12.5    	& \checkmark$^1$ & \checkmark$^2$& \checkmark\phantom{$^0$}& \checkmark\phantom{$^0$}&	 \checkmark\phantom{$^0$} 	\\
I.14.4    	& \checkmark$^1$& \checkmark$^1$& \checkmark$^1$& \checkmark$^1$&	\checkmark$^1$	\\
I.25.13   	& \checkmark\phantom{$^0$}& \checkmark$^1$& \checkmark\phantom{$^0$}& \checkmark\phantom{$^0$}&	 \checkmark\phantom{$^0$} 	\\
I.29.4	& \checkmark\phantom{$^0$}& \checkmark$^2$& \checkmark\phantom{$^0$}& \checkmark\phantom{$^0$}&	 \checkmark\phantom{$^0$}	\\
I.34.27	& \checkmark$^1$&\checkmark$^1$ & \checkmark$^1$& \checkmark$^1$&	\checkmark$^1$	\\
I.39.10   	& \checkmark$^1$& \checkmark$^1$& \checkmark$^1$& \checkmark$^1$&	\checkmark$^1$	\\
II.3.24    	& \checkmark$^1$& \checkmark$^1$& \checkmark$^1$& \checkmark$^3$&	\checkmark$^1$	\\
II.8.31 & \checkmark$^1$& \checkmark$^1$& \checkmark$^1$& \checkmark$^1$&	\checkmark$^1$	\\
II.11.28	& \checkmark$^1$& {\sffamily X}\phantom{$^0$}& {\sffamily X}\phantom{$^0$}& {\sffamily X}\phantom{$^0$}&	{\sffamily X}\phantom{$^0$}	\\
II.27.18    	& \checkmark$^1$ & \checkmark$^1$& \checkmark\phantom{$^0$}& \checkmark\phantom{$^0$}&	 \checkmark\phantom{$^0$} 	\\
II.38.14     & \checkmark$^1$& \checkmark$^2$& {\sffamily X}\phantom{$^0$}& \checkmark$^2$	&	\checkmark$^2$	\\
III.12.43    & \checkmark$^1$& \checkmark$^1$& \checkmark$^1$& 	\checkmark$^1$&	\checkmark$^1$	\\
\midrule
 Number of & 
 \multirow{2}{*}{(3, 10, 0, 0, 2)} & 
 \multirow{2}{*}{(0, 8, 4, 0, 3)} & 
 \multirow{2}{*}{(5, 6, 0, 0, 4)} & 
 \multirow{2}{*}{(5, 5, 1, 1, 3)} & 
 \multirow{2}{*}{(5, 6, 1, 0, 3)}\\
 (\checkmark, \checkmark$^1$, \checkmark$^2$, \checkmark$^3$, {\sffamily X}) & & & & & \\
  Total $\checkmark^*$ &  13/15 & 12/15 & 11/15 & 12/15 & 12/15 \\ 
 \multicolumn{6}{@{}l@{}}{\dashrule}  \\
 Accuracy & \bf 86.67\% & 80\%  & 73.33\% & 80\%  & 80\% \\ 
\bottomrule
\end{tabular}
\caption{\new{Results from running TuringBot on 15 problems from the Feynman Database for Symbolic Regression with up to two variables. The performance is higher compared to Table~\ref{tab:ai-feynman-results} since problems with at most 2 variables are easier to solve.} }\label{tab:ai-feynman-results_turingbot}
\end{table*}

\begin{table*}[h]
\centering
\begin{tabular}{lllcc}
\toprule
 Label &  Ground truth Formula &  AI-Descartes formula (SR) & Derivability &  $\exists$-Derivability\\
\midrule
I.27.6 &
$1 / (\frac{1}{d_1}+\frac{n}{d_2})$  & 
$\frac{ 0.9979 d_1^{-1} n^{-2} }{ d_1^{-2} n^{-2} + d_1^{-1}  d_2^{-1} n^{-1} }$ &
{\sffamily X} & \checkmark\\[1.5ex]
I.34.8 &  
$q v B / p$   & 
$q v B p^{-1}$ &
\checkmark & NA \\[1.5ex]
I.43.16 & 
$\mu_{d} q V_e / d$ & 
$0.9993 q \mu_{d}  {V_e} d^{-1}$ &
{\sffamily X} & \checkmark\\[1.5ex]
II.10.9 & 
$\frac{\sigma_{den}}{\epsilon(1 + \chi)}$ &
$\frac{ 0.9960 \sigma_{den}}{ \epsilon + \epsilon  \chi }$ &
{\sffamily X} & \checkmark\\[1.5ex]
II.34.2 & 
$q v r / 2$ & 
$0.5022 q  v  r$ &
{\sffamily X} & \checkmark \\
\bottomrule 
\end{tabular}
\caption{
\new{Results from running the AI-Descartes' Reasoning module on 5 selected problems (with available background theory) from the Feynman Database for Symbolic Regression. We are able to prove (either directly or existentially quantifying over the numerical constants) the correct formula, which is a part of the output of the SR module, for all the problems.}}
\label{tab:sota_reasoning}
\end{table*}

\begin{table*}[ht]
\centering
\resizebox{\linewidth}{!}{
\begin{tabular}{ll}
\toprule
Method & Configurations \\
\midrule
\multirow{2}{*}{AI-Feynman}
    & {\lstinline[]$NN_epochs=500$} \\
    & {\lstinline[]$All other parameters=default$} \\
\midrule
\multirow{7}{*}{TuringBot}
    & {\lstinline[]$Time limit (s)=120$} \\
    & {\lstinline[]$binary_operators=["*", "/", "+", "pow", "fmod"]$} \\
    & {\lstinline[]$unary_operators=["sin", "cos", "tan", "arcsin", "arccos", "arctan", "exp", "log", "log2", "sqrt",$} \\
    & ~~~~~ {\lstinline[]$"sinh", "cosh", "tanh", "arcsinh", "arccosh", "arctanh", "abs", "floor", "ceil", "round", "sign",$} \\
    & ~~~~~ {\lstinline[]$"tgamma", "lgamma", "er"]$}\\
    & {\lstinline[]$Search metric="RMS error"$}\\
\midrule
\multirow{4}{*}{PySR}
    & {\lstinline[]$niterations=300$} \\
    & {\lstinline[]$binary_operators=["*", "/", "+", "-", "div", "mult", "plus", "sub"]$} \\
    & {\lstinline[]$unary_operators=["sin", "cos", "exp", "sqrt", "square", "cube", "log", "neg", "abs"] $}\\
    & {\lstinline[]$model_selection="best"$}\\
\midrule
\multirow{6}{*}{Bayesian M.S.}
& kepler solar: {\lstinline[]$./Prior/final_prior_param_sq.named_equations.nv3.np3.2017-06-13 08:55:24.082204.dat$} \\
&  kepler exo: {\lstinline[]$./Prior/final_prior_param_sq.named_equations.nv3.np3.2017-06-13 08:55:24.082204.dat$}
\\
& kepler binary: {\lstinline[]$./Prior/final_prior_param_sq.named_equations.nv3.np3.2017-06-13 08:55:24.082204.dat$}
 \\
& langmuir Sun: {\lstinline[]$./Prior/final_prior_param_sq.named_equations.nv1.np3.2017-10-18 18:07:35.262530.dat$}
 \\
& langmuir tab: {\lstinline[]$./Prior/final_prior_param_sq.named_equations.nv1.np3.2017-10-18 18:07:35.262530.dat$}
 \\
& relativity: {\lstinline[]$./Prior/final_prior_param_sq.named_equations.nv1.np3.2017-10-18 18:07:35.262530.dat$}
 \\
\bottomrule 
\end{tabular}
}
\caption{
Configuration parameters used for comparing with AI-Feynman, TuringBot, PySR and Bayesian Machine Scientist.}
\label{tab:config_param}
\end{table*}

\clearpage

\section*{Acknowledgements}
We thank J. Ilja Siepmann for initially suggesting adsorption as a problem for symbolic regression. We thank James Chin-wen Chou for providing the atomic clock data.
{\it Funding:} This work was supported in part by the Defense Advanced Research Projects Agency (DARPA) (PA-18-02-02). The U.S. Government is authorized to reproduce and distribute reprints for Governmental purposes notwithstanding any copyright notation thereon. TRJ was supported by the U.S. Department of Energy (DOE), Office of Basic Energy Sciences, Division of Chemical Sciences, Geosciences and Biosciences (DE-FG02-17ER16362), as well as startup funding from the University of Maryland, Baltimore County. TRJ also gratefully acknowledges the University of Minnesota Institute for Mathematics and its Applications (IMA).

\end{document}